\makeatletter
\providecommand\@saveprimitive[2]{\@ifundefined{\expandafter\@gobble\string#2}{\let#2#1}{}}
\makeatother
\documentclass[11pt,reqno]{amsart}
\usepackage{microtype}
\usepackage{amsmath,amssymb,amsthm}
\usepackage{graphicx}
\usepackage{xcolor}
\usepackage{enumitem}
\usepackage[most]{tcolorbox}
\usepackage{float}
\usepackage{booktabs}
\usepackage{array}
\usepackage{algorithm}
\usepackage{algpseudocode}
\tcbset{resultbox/.style={breakable,enhanced,colback=white,boxrule=0.8pt,arc=0pt,
  left=6pt,right=6pt,top=2pt,bottom=2pt,before skip=8pt,after skip=8pt}}
\usepackage[margin=1.15in]{geometry}
\usepackage[colorlinks=true,linkcolor=blue!50!black,citecolor=blue!50!black,urlcolor=blue!50!black]{hyperref}

\theoremstyle{plain}
\newtheorem{theorem}{Theorem}[section]
\newtheorem{proposition}[theorem]{Proposition}
\newtheorem{lemma}[theorem]{Lemma}
\newtheorem{corollary}[theorem]{Corollary}
\theoremstyle{definition}
\newtheorem{definition}[theorem]{Definition}
\newtheorem{assumption}[theorem]{Assumption}

\theoremstyle{remark}
\newtheorem{remark}[theorem]{Remark}
\tcolorboxenvironment{theorem}{resultbox,colframe=blue!55!black}
\tcolorboxenvironment{proposition}{resultbox,colframe=blue!55!black}
\tcolorboxenvironment{corollary}{resultbox,colframe=blue!55!black}
\tcolorboxenvironment{lemma}{resultbox,colframe=teal!70!black}
\tcolorboxenvironment{assumption}{resultbox,colframe=orange!80!black}
\tcolorboxenvironment{definition}{resultbox,colframe=violet!65!black}
\tcolorboxenvironment{remark}{resultbox,colframe=black!45}
\tcolorboxenvironment{example}{resultbox,colframe=black!45}

\makeatletter
\renewcommand\subsection{\@startsection{subsection}{2}{\z@}{.7\linespacing\@plus\linespacing}%
  {.5\linespacing}{\normalfont\bfseries}}
\makeatother

\newcommand{\runin}[1]{\par\smallskip\noindent\textbf{#1.}\ }

\title[Market making and rough volatility]{Algorithmic Market Making Can Lead to Rough Volatility Through the Order Book}
\author{Sai Niranjan Ramachandran}
\thanks{Technical University of Munich and Munich Center for Machine Learning, Munich, Germany.
Email: \href{mailto:sainiranjan.ramachandran@tum.de}{sainiranjan.ramachandran@tum.de}}
\date{\today}


\begin{document}
\begin{abstract}
Existing work suggests that markets can show volatility with paths rougher than those of Brownian
motion. In theory this can happen when the order flow is close to instability with memory that
decays as a power law. However, it is not fully clear how
such order flow could arise from the predominantly algorithmic market making that characterizes
order book dynamics. We show one such consistent mechanism. We treat each market maker as a desk
that uses a first-order gradient rule to move the distances of its orders from the mid price
toward target distances provided by an unknown oracle. We assume that distinct desks do not act
synchronously in time but act whenever new information relevant to their respective oracles
arrives, which each desk samples over its own characteristic lookback horizon. Across
all desks the rule therefore takes the form of stochastic gradient descent (SGD) with
information-driven (mini-batch) noise. In the continuum limit this update rule makes the volume
of the order book follow a conservative stochastic partial differential equation (the
Dean--Kawasaki equation) whose fluctuations drive the order flow. The distinct horizons of the
desks further induce a persistent memory in the order flow that decays as a power law. Standard
scaling arguments then lead to rough volatility. We confirm the key predictions of the model on
simulated agents and on six months of Binance data with about 790 million orders. In liquid markets governed mainly by market makers, such as the perpetual futures of Binance, the order books are consistent with how our desks change the book.
\end{abstract}
\maketitle

\section{Introduction}

Empirical studies of many markets report volatility whose paths are rougher than those of
Brownian motion \cite{GJR18}. Such roughness can arise in theory when each order sets off very
long chains of further orders whose influence fades as a power of the elapsed time. In a scaling limit the
exponent of this power law sets the roughness of the order flow, which opposing buy and sell
flows then carry into the price \cite{EFR18,JR16}. Such order flow has been argued to follow from
the absence of arbitrage together with the splitting of large orders during execution
\cite{EFR18,JuR20}. These arguments rest on the behavior of traders. However, it is
not a priori clear whether market makers that act purely on new information can produce such
order flow through the order book on their own.

In this work we provide one mechanism that allows for this. To reflect algorithm-driven
strategies, we treat each market maker as a desk that uses a first-order gradient rule to move
the distances of its orders from the mid price toward target distances provided by an unknown
oracle. As our model has no takers, we let the mid price move toward the side of the book with
less volume near it. Each desk thus keeps correcting its orders against a moving reference. We
next assume that each desk samples information over its own characteristic horizon
\cite{LdP18}. Since at each moment only the desks that have received new information take a
step, order placements are asynchronous. Across all desks the rule therefore takes the form of
stochastic gradient descent (SGD) with information-driven (mini-batch) noise.

Adapting prior constructions of the continuum limit of SGD \cite{GGK25}, we show that the volume
of the order book then follows a conservative stochastic partial differential equation
(the Dean--Kawasaki equation \cite{Dea96,Kaw94}) driven by noise that is white in time and colored across the book, with a spatial correlation set by how desks sample information. As desks only redistribute volume, the volume of the order book is conserved irrespective of the nature of this noise. Further, every update of a desk is in turn a source of new information for the other desks; thus the aggregation of this information raises the chance of further updates across the market with a strength set by the shape of the book. Lastly, even though the horizon of each desk leaves a memory that fades exponentially, the distinct horizons cause the market to retain a more persistent memory that appears in the order flow as a power-law decay. Standard scaling arguments then lead to rough volatility when the order flow approaches instability in the asymptotic limit \cite{EFR18,JR16}.

We then proceed to test each step of this mechanism, first on simulated desks and then on six
months of Binance trades and order books covering about 790 million orders. In simulation the
average changes of the order book agree with the theory, with fluctuations that match
the information-driven noise. We further confirm that the market shows the predicted memory and
that roughness emerges only when desks sample information over distinct horizons. On Binance the order flow is also rough, and its Hurst parameter lies close to that of simulated desks with the same fitted memory. The order books of the perpetual futures there are consistent with how our desks change the book, as orders in these books partly return to their distances after a move of the mid price and move to neighboring distances (Section~\ref{sec:binbook}).

\subsection{Related literature}
Our work broadly straddles the following lines of existing research.

\runin{Rough volatility}
Rough volatility was first documented in estimates from realized volatility \cite{GJR18}.
However, other works argue that such estimates could partly reflect estimation error \cite{CD}.
Alternative justifications of roughness that do not rest on these estimates have also been
proposed, such as order flow that is close to instability with power-law memory
\cite{EFR18,JR16} and the absence of arbitrage \cite{Fuk21,JuR20}. In our work we provide a
different route that relies purely on algorithmic market making, with roughness measured from the
order count distributions that emerge from it.

\runin{Limit order dynamics}
Models of the order book itself usually prescribe how orders arrive and leave, either at random
\cite{CST10,SFGK03} or through a latent supply and demand that evolves by reaction and diffusion
\cite{BBDG18,DBMB15}. We instead choose to rely on optimization dynamics more consistent with
modern liquidity provision \cite{AS08,GLFT13,Men13}. Unlike other agent-based models
\cite{CI02,CB00,LM99}, we use a gradient-based rule consistent with modern learning approaches
and information-based sampling \cite{LdP18} that guides the actions of fast-reacting desks.

\runin{Continuum dynamics}
Multiple models of order book evolution have been proposed, but most are consistent with the
Brownian-like nature of prices \cite{BMP02,MTB14,TLDLKB11,YTST14}. We instead rely on existing
continuum approaches to analyzing optimization dynamics \cite{GGK25,MMN18,RV22,SS20} and modify
them to be more reflective of the self-exciting and persistent nature of financial order flow.

\subsection{Outline}
We provide a map for the work that follows. First we introduce the desks and their gradient rule
in Section~\ref{sec:setting}, after which we derive the conservative equation for the volume of
the order book in Section~\ref{sec:continuum}. We then proceed to show how fluctuations of the
book drive self-exciting order flow with power-law memory (Section~\ref{sec:flow}) and connect
this order flow to rough volatility (Section~\ref{sec:rough}). Sections~\ref{sec:simulation}
and~\ref{sec:binance} test the mechanism on simulated desks and on Binance data respectively.
Additional details are provided in the Appendix.

\section{Setting: SGD market makers on the order book}\label{sec:setting}

Our model has five parts, each of which is needed to turn the gradient rule of a single desk into
the dynamics of a whole book. We first describe the order book (Section~\ref{sec:book}) and how its
mid price moves (Section~\ref{sec:mid}), then the information that each desk tracks
(Section~\ref{sec:info}), the gradient rule by which a desk moves its orders
(Section~\ref{sec:rule}) and finally when a desk acts (Section~\ref{sec:batch}). Section~\ref{sec:protocol} then puts these parts together in the order
in which they occur.

\subsection{The order book}\label{sec:book}

We consider an idealized order book that consists solely of market makers and evolves in discrete
time with steps $n=0,1,\dots$. Each order is characterized by its side $\sigma\in\{-1,+1\}$,
corresponding to the bid and the ask respectively, and by its distance $x\ge0$ from a mid price
$M_n$, measured in units of a price increment $h>0$. Conventionally the mid price is defined
as $p_n=\tfrac12(A_n+B_n)$, the average of the best ask $A_n$ and the best bid $B_n$. However, in
our model $M_n$ is a reference price around which all market makers quote. In this setup, if $a_n$ and
$b_n$ denote the distances of the closest ask and the closest bid from $M_n$, then
$A_n=M_n+h a_n$ and $B_n=M_n-h b_n$, so that $p_n=M_n+\tfrac{h}{2}(a_n-b_n)$, with
equality holding if and only if $a_n=b_n$.

Our choice follows standard models of market making in which market makers quote around a reference price
\cite{AS08,GLFT13} that in practice is made to account for the imbalance of volume between the two
sides around the conventional mid price \cite{Sto18}. Under this setup the mid price moves toward
the side with less volume \cite{Sto18}, as made precise in \eqref{eq:mid}. Further, despite being an idealized quantity it is close to
the conventional mid price as it differs from the latter by at most half the spread
\begin{equation*}
 |p_n-M_n|=\tfrac{h}{2}\,|a_n-b_n|\le\tfrac12\,(A_n-B_n).
\end{equation*}
The two are thus nearly identical in highly liquid markets with a narrow spread.

In a real order book each market-making participant, which we refer to as a desk $d_i$, holds at
step $n$ a number $O_i(n)$ of orders on either side of the book. Each order is a point mass of
volume at its distance from the mid price, and desks change the number, sizes, sides and distances
of their orders over time.

In our model we represent each participant that quotes on both sides of the book by two desks, one
for each side, such that each desk $d_i$ quotes on a single side $\sigma_i$. As volume that a
participant shifts between its two desks leaves the total volume of the book unchanged, our
framework still applies to the book as a whole. We further let each desk hold a fixed number $L$ of
orders of equal volume $m_i$, which discretizes the volume of the desk into equal parts such that a
real order corresponds to one or more of these parts at the same distance. Desks in our model only
move their orders, which on an exchange corresponds to cancelling an order and immediately placing
the same volume at a new distance. Our idealized market thus has no other arrivals, cancellations or
annihilations of orders.\footnote{This idealization is closest to liquid markets where market makers cancel and requote
their orders within seconds \cite{HS09} to limit adverse selection \cite{CJR14}.} As the volume of a liquid book spreads over many ticks we treat distances
as continuous, so that desk $d_i$ holds its orders $o_{i,1},\dots,o_{i,L}$ at the distances
$x_i(n)=(x_{i,1}(n),\dots,x_{i,L}(n))\in[0,\infty)^L$ and the volume on side $\sigma$ at step $n$
is the measure
\begin{equation*}
 V^{\sigma}_n=\sum_{i:\,\sigma_i=\sigma} m_i\sum_{j=1}^{L}\delta_{x_{i,j}(n)},
\end{equation*}
which places the volume $m_i$ of each order of desk $d_i$ at its distance.
The total volume $V^{+}_n([0,\infty))+V^{-}_n([0,\infty))$ of the book is constant in $n$. In our
simulations participants do not shift volume between their two desks, such that the volume on each
side is constant as well, which we normalize to one.

\subsection{How the mid price moves}\label{sec:mid}
We first measure how much volume sits near the mid price on each side. We weigh each order by its
volume and by its distance from the mid price as
\begin{equation}\label{eq:near}
 Z^{\sigma}_n=\sum_{i:\,\sigma_i=\sigma} m_i\sum_{j=1}^{L}e^{-x_{i,j}(n)/\ell},
\end{equation}
such that an order at the mid price itself ($x=0$), which we call the touch, counts with its full
volume while the weight of an order decays with its distance over a length $\ell>0$. We chose this soft weighting rather than counting only the
volume within a fixed distance, as a fixed cutoff would make $Z^{\sigma}_n$ jump whenever an order
crosses it. Instead, the chosen soft weighting allows the mid price to respond smoothly to every
move. As each side is individually normalized to have total volume one, $0\le Z^{\sigma}_n\le1$,
and thus $Z^{\sigma}_n$ naturally represents the share of the volume of side $\sigma$ near the mid
price. We relate this share of the volume to the movement of the mid price through
\begin{equation}\label{eq:mid}
 \theta_n=-c\,\bigl(Z^{+}_n-Z^{-}_n\bigr),\qquad M_{n+1}=M_n+h\,\theta_n,
\end{equation}
where $c>0$ sets how far the mid price moves for a given imbalance of volume near it. The above
linear response is consistent to first order with empirical observations of how moves of the mid
price depend on the imbalance of volume \cite{CDJ18,CKS14,GB16}. Our weighting additionally accounts
for the contribution of volume at larger distances from the mid price \cite{XGH20}.

Since all distances are measured from the mid price, every order shifts with each move of the mid
price. Each order of desk $d_i$ thus moves as
\begin{equation}\label{eq:shift}
 x_{i,j}\leftarrow\max\{x_{i,j}-\sigma_i\theta_n,\,0\},
\end{equation}
such that orders on the side toward which the mid price moves come closer to it while orders on the
other side move away. An order that the mid price reaches stays at the touch, as in our idealized
market it can neither be executed nor cross to the other side of the book.

\subsection{Desks and their information}\label{sec:info}
Next, we group the set of all desks into $K$ clusters, where cluster $k$ contains the desks that
sample information over a characteristic horizon $T_k=T_1\Omega^{k-1}$ with a ratio $\Omega>1$
\cite{Cor09,LdP18,Mue97}. As a large part of the liquidity in modern markets is provided by fast
desks \cite{Men13}, we let the volume of a cluster decrease as a power of its horizon. The desks of
cluster $k$ thus together hold the volume
\begin{equation}\label{eq:weights}
 w_k=\frac{T_k^{-\alpha}}{\sum_{l=1}^{K}T_l^{-\alpha}},\qquad\alpha>0,
\end{equation}
on each side, such that the volumes of all clusters sum to one. For simplicity each of the $N$ desks
of cluster $k$ on a given side holds the same volume, such that $m_i=w_k/(NL)$. In a real market
this amounts to fixing the resolution at which we describe the volume of each cluster.

The desks of cluster $k$ then track the moving averages of two quantities, which reflect two
economic realities.
\begin{enumerate}[label=(\roman*)]
 \item Market makers adjust their quotes to the recent trend of the mid price, as a predictable
  short-term drift exposes them to adverse selection \cite{CJR14}. The first moving average $P_k$
  therefore follows this trend over the horizon $T_k$.
 \item Activity in order books is to a large extent self-exciting, such that recent updates of
  other participants are themselves a source of information for further updates
  \cite{FS12,HBB13}. The second moving average $U_k$ therefore follows the recent size of the
  updates of all desks, which measures market activity.
\end{enumerate}
At every step both are updated as
\begin{equation}\label{eq:memory}
 P_k\leftarrow P_k+\frac{\theta_n-P_k}{T_k},\qquad U_k\leftarrow U_k+\frac{I_{n-1}-U_k}{T_k},
\end{equation}
where $I_{n-1}$ measures how strongly the updates of the previous step moved the mid price, as
defined in \eqref{eq:push} below.

\subsection{The gradient rule}\label{sec:rule}
We now state the exact optimization rule. Each desk $d_i$ of cluster $k$ receives target distances
$x^{*}_i\in[0,\infty)^L$ for its orders from an unknown oracle, which stands for the desk's own view
of where its orders should sit. As stated earlier, following the
recent trend of the mid price, it further leans its orders such that in an upward trend its asks
move away from the mid price while its bids move closer. We express both through the loss
\begin{equation}\label{eq:loss}
 \mathcal L_i(x)=\frac12\sum_{j=1}^{L}\bigl(x_j-x^{*}_{i,j}\bigr)^2-\sigma_i\lambda_k\sum_{j=1}^{L}x_j,
 \qquad\lambda_k=\frac{\Pi P_k}{\gamma},
\end{equation}
where $\Pi\ge0$ sets how strongly the desk leans with the trend and dividing by $\gamma$ makes the
lean of a single step equal to $\Pi P_k$. When the desk acts, it takes a
single gradient step of size $\gamma\in(0,1)$ as
\begin{equation*}
 x_i\leftarrow\max\bigl\{x_i-\gamma\nabla\mathcal L_i(x_i),\,0\bigr\}
 =\max\bigl\{x_i-\gamma\,(x_i-x^{*}_i)+\sigma_i\Pi P_k,\,0\bigr\}.
\end{equation*}
Notice that as this step affects the volume near the mid price, the next move of the mid price in
turn changes by $c\,m_i|\Delta Z_i|$, where $m_i\Delta Z_i$ is the change of $Z^{\sigma_i}_n$ caused by
desk $d_i$ with
$\Delta Z_i=\sum_{j}\bigl(e^{-x^{\mathrm{new}}_{i,j}/\ell}-e^{-x^{\mathrm{old}}_{i,j}/\ell}\bigr)$. The updates of all desks
acting in step $n$ thus change the next move of the mid price by at most
\begin{equation}\label{eq:push}
 I_n=\sum_{d_i\ \text{acts in step}\ n} c\,m_i\,|\Delta Z_i| ,
\end{equation}
which we call the size of the updates in step $n$.

\subsection{The mini-batch}\label{sec:batch}
Lastly we note that each desk can act only when new information relevant to its oracle arrives.
This is modelled by letting each desk act independently with probability
\begin{equation}\label{eq:batch}
 p_k=\min\{p_0+\beta U_k,\,1\},
\end{equation}
where $k$ denotes its cluster membership, $p_0>0$ accounts for information that does not come from the desks and $\beta\ge0$ sets how
strongly desks respond to the recent updates of all desks, including their own.
Further, writing $\xi_i(n)\sim\mathrm{Bernoulli}(p_k)$ for the indicator that desk $d_i$ acts in
step $n$, its orders move as
\begin{equation}\label{eq:sgd}
 x_i\leftarrow\max\bigl\{x_i-\xi_i(n)\,\gamma\nabla\mathcal L_i(x_i),\,0\bigr\}.
\end{equation}
These random updates introduce stochasticity into gradient descent akin to mini-batch optimization
in machine learning \cite{BCN18}. Unlike standard schemes that sample a fixed number of data points
per update, our batch consists of the desks that are active and thus grows or shrinks with the size
of recent updates.

\subsection{The market protocol}\label{sec:protocol}
We initialize $P_k=U_k=I_{-1}=0$ and place every order at its target. Though
$\nabla\mathcal L_i=0$ for every desk, differing targets for distinct desks create a movement in
the mid price and thus kickstart our market dynamics. We now walk through the sequence of events in a single step $n$ of the market. First, the mid price
moves by $\theta_n$ and shifts all orders following \eqref{eq:near}--\eqref{eq:shift}. Each desk
then updates its internal trend $P_k$ of the mid price and its estimate $U_k$ of market activity
through \eqref{eq:memory}. Following this update each desk acts independently with probability $p_k$ in
\eqref{eq:batch} and, if it acts, takes a gradient step toward its targets by \eqref{eq:sgd}. Consequently,
all such steps together change the volume near the mid price. This change in turn sets the next
move of the mid
price as the market passes to step $n+1$, and its size is at most the size of the updates $I_n$ in \eqref{eq:push}.
In the following section we study the continuum dynamics of the order book induced by our setting.

\section{The order book in the continuum}\label{sec:continuum}

We now pass from the discrete protocol (Section~\ref{sec:setting}) to a continuum description of
the order book. We first write one step of the protocol as an exact sum of a mean and a fluctuation
(Section~\ref{sec:particles}), which leads to a conservative limit as the step size $\gamma$ tends to
zero (Section~\ref{sec:spde}). In this limit the order book preserves its total volume, such that
any redistribution occurs only through continuous flows between depths
(Section~\ref{sec:conservation}).

\subsection{From the protocol to particles}\label{sec:particles}
We first place both sides of the book on a single line. For each order $o_{i,j}$ let its signed
distance be defined as $y_{i,j}=\sigma_i x_{i,j}$. Clearly bids satisfy $y\le0$ and asks $y\ge0$,
while the mid price lies exactly at the origin. As long as no order reaches the touch, the shift
\eqref{eq:shift} moves each order as $y_{i,j}\leftarrow y_{i,j}-\theta_n$. The movement of the mid
price is therefore just a uniform translation of the whole book.

As the drift of each order $o_{i,j}$ depends on its index $(i,j)$ through its target and cluster,
we keep this index in the state of the book. The state at step $n$ is then the measure
\begin{equation}\label{eq:lifted}
 \mu_n=\sum_{i}m_i\sum_{j=1}^{L}\delta_{(y_{i,j}(n),\,(i,j))}
\end{equation}
together with the moving averages $P_k$ and $U_k$, from which the volume $V^{\pm}_n$ on each side
follows by ignoring the indices. Following Section~\ref{sec:protocol}, the initial state places every order at its
target, such that $\mu_0=\sum_i m_i\sum_j\delta_{(\sigma_i x^{*}_{i,j},\,(i,j))}$ and
$P_k=U_k=I_{-1}=0$.

We next split the step \eqref{eq:sgd} of a desk into its mean and a fluctuation. Recall that desk
$d_i$ acts in step $n$ when the Bernoulli variable $\xi_i(n)\in\{0,1\}$ equals one, which happens with
probability $p_k$. Writing
\begin{equation*}
 g_{i,j}=-\gamma\,\sigma_i\,\partial_j\mathcal L_i(x_i)
\end{equation*}
for the signed move of order $o_{i,j}$ when its desk acts, the expected move of this order given the
state $\mathcal F_n$ of the market at step $n$ is
\begin{equation*}
 \mathbb E\bigl[\xi_i(n)\,g_{i,j}\mid\mathcal F_n\bigr]=p_k\,g_{i,j}+(1-p_k)\cdot0=p_k\,g_{i,j}.
\end{equation*}
The move itself thus splits into this expected value and a fluctuation with mean zero,
\begin{equation*}
 \xi_i(n)\,g_{i,j}=p_k\,g_{i,j}+\bigl(\xi_i(n)-p_k\bigr)\,g_{i,j}.
\end{equation*}
Notice that the fluctuations of different orders are correlated through the actions of individual
desks. We quantify this covariance below.

\begin{lemma}[Noise of the mini-batch]\label{lem:noise}
The fluctuations $\zeta_{i,j}=(\xi_i(n)-p_k)\,g_{i,j}$ satisfy
\begin{equation}\label{eq:cov}
 \mathbb E\bigl[\zeta_{i,j}\,\zeta_{i',j'}\mid\mathcal F_n\bigr]
 =\mathbf 1\{i=i'\}\;p_k(1-p_k)\,\gamma^2\,\partial_j\mathcal L_i(x_i)\,\partial_{j'}\mathcal L_i(x_i).
\end{equation}
Orders of the same desk thus move together, while orders of different desks move independently.
\end{lemma}

\begin{proof}
Given $\mathcal F_n$, the moves $g_{i,j}$ are fixed, as they depend only on the positions, the targets
and the trend $P_k$ at step $n$. Hence
\[
 \mathbb E\bigl[\zeta_{i,j}\,\zeta_{i',j'}\mid\mathcal F_n\bigr]
 =g_{i,j}\,g_{i',j'}\;\mathbb E\bigl[(\xi_i(n)-p_k)(\xi_{i'}(n)-p_{k'})\mid\mathcal F_n\bigr].
\]
For $i\neq i'$ the two indicators are independent given $\mathcal F_n$, hence the expectation
factorizes into $\mathbb E[\xi_i(n)-p_k\mid\mathcal F_n]\,\mathbb E[\xi_{i'}(n)-p_{k'}\mid\mathcal F_n]=0$.
For $i=i'$ it equals the variance $\mathbb E[(\xi_i(n)-p_k)^2\mid\mathcal F_n]=p_k(1-p_k)$ of a Bernoulli
variable. Lastly, noting that $\sigma_i^2=1$, we obtain $g_{i,j}\,g_{i,j'}=\gamma^2\,\partial_j\mathcal L_i(x_i)\,
\partial_{j'}\mathcal L_i(x_i)$.
\end{proof}

We now proceed to study the asymptotic limit of this decomposition into mean and fluctuation terms as
the step size $\gamma$ tends to zero.

\subsection{The conservative limit}\label{sec:spde}
We pass to the continuum in two steps. We first let the step size $\gamma$ tend to zero and show that the orders, the trends $P_k$ and the estimates $U_k$ follow a stochastic differential equation (SDE) with one Brownian motion per desk. Summing this SDE over all orders, we then obtain a conservative stochastic partial differential equation for the volume of the book, of the Dean--Kawasaki type \cite{Dea96,Kaw94} studied in \cite{GGK25}.

\runin{The time scale and the regime}We consider the time scale on which step $n=0,1,\dots$ occurs at the time $t=n\gamma$, such that fixing $t>0$ and letting $\gamma\to0^{+}$ allows us to study the continuum limit. We then choose how the parameters of the protocol scale with $\gamma$, such that every part of the protocol remains consistent in this limit. If a quantity changes by $\delta$ in each step, then up to the time $t$ it changes by
\begin{equation*}
 \sum_{n<t/\gamma}\delta=\frac{t}{\gamma}\,\delta,
\end{equation*}
which stays bounded as $\gamma\to0^{+}$ only if $\delta=O(\gamma)$. Following the gradient step (Section~\ref{sec:rule}), an acting desk moves its orders by $-\gamma(x_i-x^*_i)+\sigma_i\Pi P_k$, where only the first term is $O(\gamma)$. Further, $\theta_n=-c\,(Z^+_n-Z^-_n)$ and its average $P_k$ are $O(c)$, while each moving average moves by its gap divided by $T_k$. We therefore assume that our regime satisfies the following.

\begin{assumption}[Regime]\label{ass:regime}
There exist constants $\bar c>0$, $\tau_1<\dots<\tau_K$ and $\bar\beta\ge0$, independent of $\gamma$, such that
\begin{equation*}
 c=\gamma\,\bar c,\qquad T_k=\frac{\tau_k}{\gamma},\qquad\beta=\frac{\bar\beta}{\gamma^{2}}.
\end{equation*}
\end{assumption}

Notice that the assumption follows from the fact that each scaling is the only one under which its part of the protocol remains consistent in the limit. If $c=o(\gamma)$ the mid price does not move up to the time $t$, whereas if $\gamma=o(c)$ it moves by an unbounded amount. Similarly, $T_k=\tau_k/\gamma$ fixes the horizon of cluster $k$ at the time $\tau_k$. Further, since $|e^{-a/\ell}-e^{-b/\ell}|\le|a-b|/\ell$ for $a,b\ge0$, an acting desk changes its volume near the mid price by
\begin{equation*}
 |\Delta Z_i|\le\frac1\ell\sum_{j=1}^{L}\bigl|x^{\mathrm{new}}_{i,j}-x^{\mathrm{old}}_{i,j}\bigr|=O(\gamma),
\end{equation*}
such that $I_n=O(\gamma c)=O(\gamma^2)$. The term $\beta U_k$ thus stays $O(1)$ without vanishing only if $\beta$ is proportional to $\gamma^{-2}$.

Now, let
\begin{equation*}
 \theta_n=\gamma\,v_n,\qquad P_k=\gamma\,\pi_k,\qquad U_k=\gamma^{2}u_k,\qquad I_n=\gamma^{2}\iota_n,
\end{equation*}
where $v_n=-\bar c\,(Z^+_n-Z^-_n)$ is the velocity of the mid price. Then $\lambda_k=\Pi\pi_k$ and $p_k=\min\{p_0+\bar\beta u_k,1\}$ do not depend on $\gamma$.

Next, we keep the protocol away from the points at which its functions are not smooth, as the limit requires Lipschitz coefficients. The protocol contains three such functions,
\begin{equation*}
 u\mapsto\min\{p_0+\bar\beta u,1\},\qquad x\mapsto\max\{x,0\},\qquad p\mapsto\sqrt{p(1-p)} .
\end{equation*}
The first two have a kink, while the standard deviation $p\mapsto\sqrt{p(1-p)}$ of $\xi_i(n)\sim\mathrm{Bernoulli}(p)$ is not Lipschitz continuous at $p=1$.

\begin{assumption}[Interior]\label{ass:interior}
At every step,
\begin{enumerate}[label=(\alph*),leftmargin=*]
\item $p_0+\bar\beta u_k\le\bar p$ for every cluster $k$ and some $\bar p<1$;
\item $x_{i,j}-\sigma_i\theta_n>0$ and $x_{i,j}-\xi_i(n)\,\gamma\,\partial_j\mathcal L_i(x_i)>0$ for every order $o_{i,j}$.
\end{enumerate}
\end{assumption}

Under (a) the standard deviation is Lipschitz continuous. Under (b) a move of the mid price shifts every order by $-\theta_n$ in signed distances. Both conditions concern the path of the protocol rather than its parameters, and we check them along simulated paths in Section~\ref{sec:simulation}.\footnote{Both conditions are close to real markets. Condition (a) reflects that desks do not requote at every opportunity, and trading venues in the European Union further limit the ratio of unexecuted orders to transactions \cite{EU17}. Condition (b) reflects that the mid price lies strictly between the best bid and the best ask, such that no order sits at it.}

\runin{The protocol in signed distances}We now write one step of the protocol as an explicit map of the state of the market, such that we can compute its mean and fluctuation. Let $y^*_{i,j}=\sigma_ix^*_{i,j}$, and let $k(i)$ denote the cluster of desk $d_i$. Since $\partial_j\mathcal L_i(x_i)=x_{i,j}-x^*_{i,j}-\sigma_i\Pi\pi_{k(i)}$ and $\sigma_i^2=1$, the move $g_{i,j}$ (Section~\ref{sec:particles}) reads
\begin{equation*}
 g_{i,j}=-\gamma\sigma_i\bigl(x_{i,j}-x^*_{i,j}-\sigma_i\Pi\pi_{k(i)}\bigr)=\gamma\,b_{i,j},\qquad b_{i,j}=-\bigl(y_{i,j}-y^*_{i,j}\bigr)+\Pi\pi_{k(i)} .
\end{equation*}
Writing $Z^+-Z^-$ in signed distances and expanding the exponentials, we similarly obtain
\begin{equation*}
 \begin{aligned}
v&=-\bar c\sum_i\sigma_im_i\sum_{j=1}^{L}e^{-\sigma_iy_{i,j}/\ell},\\
\Delta Z_i&=\sum_{j=1}^{L}\Bigl(e^{-(x_{i,j}+\gamma\sigma_ib_{i,j})/\ell}-e^{-x_{i,j}/\ell}\Bigr)=\gamma a_i+O(\gamma^2),\qquad a_i=-\frac{\sigma_i}{\ell}\sum_{j=1}^{L}e^{-\sigma_iy_{i,j}/\ell}\,b_{i,j}.
\end{aligned}
\end{equation*}
The protocol of Section~\ref{sec:protocol} updates $U_k$ with $I_{n-1}$. We therefore let step $n$ start just before the desks act, which keeps the order of events and updates $U_k$ with $I_n$. The state $X_n=(y_{i,j},\pi_k,u_k)_{i,j,k}$ of the market then evolves as
\begin{equation*}
 \begin{aligned}
y_{i,j}(n+1)&=y_{i,j}(n)+\gamma\,\xi_i(n)\,b_{i,j}-\gamma v,\\
\pi_k(n+1)&=\pi_k(n)+\frac{\gamma}{\tau_k}\bigl(v-\pi_k(n)\bigr),\\
u_k(n+1)&=u_k(n)+\frac{\gamma}{\tau_k}\bigl(\iota_n-u_k(n)\bigr),\qquad\iota_n=\frac{\bar c}{\gamma}\sum_i\xi_i(n)\,m_i|\Delta Z_i|,
\end{aligned}
\end{equation*}
where $v$ is evaluated after the gradient steps. Let $\mathcal F_n$ be the $\sigma$-algebra generated by $\xi(0),\dots,\xi(n-1)$. As the initial state is fixed, $X_n$ is $\mathcal F_n$-measurable. Given $\mathcal F_n$, the indicators $\xi_i(n)$ are independent with $\mathbb E[\xi_i(n)\mid\mathcal F_n]=p_{k(i)}$.

\runin{Mean and fluctuation}Following Section~\ref{sec:particles}, we next split each step into its conditional mean and a fluctuation with mean zero, and we compute its covariance with Lemma~\ref{lem:noise}. Summed over many steps, the fluctuation becomes Gaussian with independent increments, from which we obtain an SDE. Taking the conditional expectation and using $|\Delta Z_i|=\gamma|a_i|+O(\gamma^2)$, we obtain
\begin{equation*}
 \mathbb E[X_{n+1}-X_n\mid\mathcal F_n]=\gamma F(X_n)+O(\gamma^2)
\end{equation*}
with the drift $F$ given by
\begin{equation*}
 F_{y_{i,j}}=p_{k(i)}b_{i,j}-v,\qquad F_{\pi_k}=\frac{v-\pi_k}{\tau_k},\qquad F_{u_k}=\frac{1}{\tau_k}\Bigl(\bar c\sum_ip_{k(i)}m_i|a_i|-u_k\Bigr).
\end{equation*}
Let $s_k=\sqrt{p_k(1-p_k)}$, and let $S_i$ be the vector with the entry $s_{k(i)}b_{i,j}$ for each order of $d_i$, the entry $s_{k(i)}\bar c\,m_i|a_i|/\tau_k$ for each $u_k$, and zero otherwise. The fluctuation
\begin{equation*}
 \zeta_{n+1}=\frac{1}{\gamma}\Bigl(X_{n+1}-X_n-\mathbb E[X_{n+1}-X_n\mid\mathcal F_n]\Bigr)=\sum_i\frac{\xi_i(n)-p_{k(i)}}{s_{k(i)}}\,S_i(X_n)+O(\gamma)
\end{equation*}
is then linear in the centered indicators and, by construction, has $\mathbb E[\zeta_{n+1}\mid\mathcal F_n]=0$. For the entries of the orders, we obtain the covariance \eqref{eq:cov} divided by $\gamma^2$ from Lemma~\ref{lem:noise}. The same computation applies to the entries of $u_k$, as their fluctuation is driven by the same centered indicators. Together, we obtain
\begin{equation*}
 \mathbb E\bigl[\zeta_{n+1}\zeta_{n+1}^{\top}\mid\mathcal F_n\bigr]=\sum_iS_i(X_n)S_i(X_n)^{\top}+O(\gamma).
\end{equation*}
Since $\zeta_{m+1}$ is $\mathcal F_k$-measurable for $m<k$, we further obtain from the tower property
\begin{equation*}
 \mathbb E\bigl[\zeta_{m+1}^{\top}\zeta_{k+1}\bigr]=\mathbb E\bigl[\zeta_{m+1}^{\top}\,\mathbb E[\zeta_{k+1}\mid\mathcal F_k]\bigr]=0 .
\end{equation*}
Therefore $\mathcal M_n=\sum_{m<n}\zeta_{m+1}$ is a martingale with orthogonal increments.

For $\gamma\ll\Delta\ll1$ the state changes little over the $\Delta/\gamma$ steps from $t$ to $t+\Delta$. By the central limit theorem we then obtain
\begin{equation*}
 \gamma\sum_{n=t/\gamma}^{(t+\Delta)/\gamma-1}\frac{\xi_i(n)-p_{k(i)}}{s_{k(i)}}\,S_i(X_n)\approx\gamma\sqrt{\frac{\Delta}{\gamma}}\,S_i(X_{t/\gamma})\,\mathcal N_i=\sqrt{\gamma}\,S_i(X_{t/\gamma})\sqrt{\Delta}\,\mathcal N_i,
\end{equation*}
with independent standard Gaussian variables $\mathcal N_i$. Since the indicators are independent across steps, these increments are independent over disjoint intervals. From the sampling we therefore obtain that they are the increments of independent standard Brownian motions $W_i$, which leads us to the SDE
\begin{equation*}
 dX=F(X)\,dt+\sqrt{\gamma}\sum_iS_i(X)\,dW_i.
\end{equation*}
Without the noise, the SDE becomes an ordinary differential equation (ODE) for $\bar X$. In both cases the mid price follows from $M_{n+1}=M_n+h\gamma v_n$,
\begin{equation*}
 \frac{d\bar X}{dt}=F(\bar X),\qquad M(t)=M(0)+h\int_0^tv(s)\,ds.
\end{equation*}

\runin{Distance from the ODE}We now bound the distance of the protocol and of the SDE from $\bar X$ on a finite horizon $T$. We use that $F$ is Lipschitz, such that deviations from $\bar X$ grow at most exponentially, while the fluctuations of $T/\gamma$ steps add up to a variance proportional to $\gamma$. We need Lipschitz coefficients for both facts. We therefore first show that the protocol stays in a bounded set, on which $F$ and $S_i$ are Lipschitz.

\begin{lemma}[Bounded state]\label{lem:bounded}
Let Assumptions~\ref{ass:regime} and~\ref{ass:interior} hold, let $\gamma\le\tau_1$ and $T>0$, and let $R_T=(1+\Pi)\,\bar c\,(T+1)$. Then for every step $n$ with $n\gamma\le T$,
\begin{equation*}
 |v|\le\bar c,\qquad|\pi_k|\le\bar c,\qquad|y_{i,j}-y^*_{i,j}|\le R_T,\qquad0\le u_k\le\bar u_T=\frac{2\bar c\,(R_T+\Pi\bar c)}{\ell}.
\end{equation*}
\end{lemma}

\begin{proof}
Since $0\le Z^\sigma\le1$ and $\gamma\le\tau_k$, the velocity and the trends satisfy
\begin{equation*}
 |v|=\bar c\,|Z^+-Z^-|\le\bar c,\qquad|\pi_k(n+1)|\le\Bigl(1-\frac{\gamma}{\tau_k}\Bigr)|\pi_k(n)|+\frac{\gamma}{\tau_k}\,|v|\le\bar c,
\end{equation*}
where the second bound follows by induction from $\pi_k(0)=0$. For the distances, we obtain from the protocol
\begin{equation*}
 \begin{aligned}
|y_{i,j}(n+1)-y^*_{i,j}|&=\bigl|(1-\gamma\xi_i(n))\bigl(y_{i,j}(n)-y^*_{i,j}\bigr)+\gamma\bigl(\xi_i(n)\Pi\pi_{k(i)}-v\bigr)\bigr|\\
&\le|y_{i,j}(n)-y^*_{i,j}|+\gamma(1+\Pi)\bar c .
\end{aligned}
\end{equation*}
But since Section~\ref{sec:protocol} places every order at its target and the first move of the mid price is at most $\gamma\bar c$, every order starts within $\gamma\bar c$ of its target. Hence $|y_{i,j}(n)-y^*_{i,j}|\le(n+1)\gamma(1+\Pi)\bar c\le R_T$. Lastly, using $|b_{i,j}|\le R_T+\Pi\bar c$, Assumption~\ref{ass:interior}(b) and $\sum_iLm_i=2$, we obtain
\begin{equation*}
 0\le\iota_n\le\frac{\bar c}{\gamma}\sum_im_i|\Delta Z_i|\le\frac{\bar c}{\gamma}\sum_im_i\sum_{j=1}^{L}\frac{\gamma|b_{i,j}|}{\ell}\le\frac{2\bar c\,(R_T+\Pi\bar c)}{\ell}.
\end{equation*}
The bound on $u_k$ follows by the same induction as for $\pi_k$. 
\end{proof}

Lemma~\ref{lem:bounded} bounds the state of the protocol, while the noise can carry the SDE beyond these bounds. We therefore change $F$ and $S_i$ only outside the bounds (Lemma~\ref{lem:bounded}), which leaves the protocol unaffected. With the changed coefficients, we compare all three processes using Lipschitz estimates. Let $K_T$ be the box of states with $|\pi_k|\le\bar c$, $|y_{i,j}-y^*_{i,j}|\le R_T$ and $0\le u_k\le\min\{\bar u_T,(\bar p-p_0)/\bar\beta\}$. By Lemma~\ref{lem:bounded} and Assumption~\ref{ass:interior}(a) the protocol stays in $K_T$. On $K_T$ we have $p_k\in[p_0,\bar p]$, such that $F$ and every $S_i$ are bounded and Lipschitz continuous with a common constant $\Lambda$. Outside $K_T$ we evaluate them at the nearest point of the box,
\begin{equation*}
 \tilde F(x)=F\bigl(\mathrm{pr}_{K_T}(x)\bigr),\qquad\tilde S_i(x)=S_i\bigl(\mathrm{pr}_{K_T}(x)\bigr),
\end{equation*}
where $\mathrm{pr}_{K_T}$ clips every coordinate to its interval. Clipping is $1$-Lipschitz and equals the identity on $K_T$. The extensions therefore agree with $F$ and $S_i$ on $K_T$ and are bounded and $\Lambda$-Lipschitz everywhere. We write $\Gamma$ for a common bound of $|\tilde F|$, of $\sum_i|\tilde S_i|^2$ and of $|X_{n+1}-X_n|/\gamma$.

\begin{theorem}[Continuum limit]\label{thm:limit}
Let Assumptions~\ref{ass:regime} and~\ref{ass:interior} hold, let $\gamma\le\min\{\tau_1,1\}$ and $T>0$, and let $\bar X$ and $X$ solve $d\bar X=\tilde F(\bar X)\,dt$ and $dX=\tilde F(X)\,dt+\sqrt\gamma\sum_i\tilde S_i(X)\,dW_i$, both started from $X_0$. Then there is a constant $C$, which depends on $T$, $\bar c$, $\Pi$, $\bar\beta$, $\ell$, $p_0$, $\bar p$, $\tau_1,\dots,\tau_K$ and the targets but not on $\gamma$, such that
\begin{enumerate}[label=(\roman*),leftmargin=*]
\item $\mathbb E\max_{n\gamma\le T}|X_n-\bar X(n\gamma)|^2\le C\gamma$,
\item $\mathbb E\sup_{t\le T}|X(t)-\bar X(t)|^2\le C\gamma$.
\end{enumerate}
\end{theorem}

On every finite horizon the protocol and the SDE thus stay within $O(\sqrt\gamma)$ of $\bar X$.

\begin{proof}
We begin with (i). By Lemma~\ref{lem:bounded} the protocol stays in $K_T$, where $\tilde F=F$. With the fluctuation $\zeta_{n+1}$ from above, each step reads
\begin{equation*}
 X_{n+1}=X_n+\gamma\tilde F(X_n)+\gamma r_n+\gamma\zeta_{n+1},\qquad|r_n|\le \Gamma_1\gamma,\qquad|\zeta_{n+1}|\le2\Gamma,
\end{equation*}
where $r_n$ is the $O(\gamma^2)$ term of the conditional mean divided by $\gamma$. As $|\bar X(s)-\bar X(n\gamma)|\le \Gamma(s-n\gamma)$, the solution $\bar X$ satisfies
\begin{equation*}
 \bar X\bigl((n+1)\gamma\bigr)=\bar X(n\gamma)+\gamma\tilde F\bigl(\bar X(n\gamma)\bigr)+e_n,\qquad|e_n|\le\int_{n\gamma}^{(n+1)\gamma}\Lambda \Gamma(s-n\gamma)\,ds=\frac{\Lambda \Gamma \gamma^2}{2}.
\end{equation*}
Subtracting and summing from $D_0=0$, we obtain for the deviation $D_n=X_n-\bar X(n\gamma)$ with $n\gamma\le T$
\begin{equation*}
 \begin{aligned}
|D_n|&\le\gamma\Lambda\sum_{m<n}|D_m|+n\Bigl(\Gamma_1+\frac{\Lambda \Gamma}{2}\Bigr)\gamma^2+\gamma|\mathcal M_n|\\
&\le\gamma\Lambda\sum_{m<n}|D_m|+\Bigl(\Gamma_1+\frac{\Lambda \Gamma}{2}\Bigr)T\gamma+\gamma\max_{l\le T/\gamma}|\mathcal M_l| .
\end{aligned}
\end{equation*}
By the discrete Gronwall inequality and $(1+\gamma\Lambda)^n\le e^{\Lambda T}$, we obtain
\begin{equation*}
 \max_{n\gamma\le T}|D_n|\le e^{\Lambda T}\Bigl(\bigl(\Gamma_1+\tfrac{\Lambda \Gamma}{2}\bigr)T\gamma+\gamma\max_{l\le T/\gamma}|\mathcal M_l|\Bigr).
\end{equation*}
As $\mathcal M_n$ is a martingale with orthogonal increments, we obtain from Doob's maximal inequality
\begin{equation*}
 \mathbb E\max_{l\le T/\gamma}|\mathcal M_l|^2\le4\,\mathbb E\bigl|\mathcal M_{\lfloor T/\gamma\rfloor}\bigr|^2=4\sum_{m<T/\gamma}\mathbb E|\zeta_{m+1}|^2\le\frac{16\Gamma^2T}{\gamma}.
\end{equation*}
Squaring the previous bound with $(a+b)^2\le2a^2+2b^2$, we obtain
\begin{equation*}
 \mathbb E\max_{n\gamma\le T}|D_n|^2\le2e^{2\Lambda T}\Bigl(\bigl(\Gamma_1+\tfrac{\Lambda \Gamma}{2}\bigr)^2T^2\gamma^2+16\Gamma^2T\gamma\Bigr)\le C\gamma .
\end{equation*}
For (ii), the deviation $D(t)=X(t)-\bar X(t)$ satisfies
\begin{equation*}
 \sup_{s\le t}|D(s)|\le\Lambda\int_0^t\sup_{r\le s}|D(r)|\,ds+\sqrt\gamma\sup_{s\le T}|N_s|,\qquad N_t=\sum_i\int_0^t\tilde S_i(X)\,dW_i .
\end{equation*}
By Gronwall's inequality, Doob's maximal inequality and the It\^o isometry, we then obtain
\begin{equation*}
 \mathbb E\sup_{t\le T}|D(t)|^2\le\gamma e^{2\Lambda T}\,\mathbb E\sup_{t\le T}|N_t|^2\le4\gamma e^{2\Lambda T}\,\mathbb E\int_0^T\sum_i|\tilde S_i(X)|^2\,dt\le4\Gamma Te^{2\Lambda T}\gamma .\qedhere
\end{equation*}
\end{proof}

\begin{remark}[The role of the noise]\label{rem:noise}
In Theorem~\ref{thm:limit} we bounded the fluctuations around $\bar X$ from above, while \cite[Theorem~2]{KH79} describes their size and shape near equilibrium. Let $X^*$ be a globally asymptotically stable rest point of the ODE, at which the book is stationary relative to the mid price. Under the conditions of \cite{KH79}, the rescaled deviation $(X_n-X^*)/\sqrt\gamma$ of the protocol converges to the stationary solution of
\begin{equation*}
 dU=\nabla F(X^*)\,U\,dt+\Bigl(\sum_iS_i(X^*)S_i(X^*)^{\top}\Bigr)^{1/2}dB .
\end{equation*}
Since $F(X^*)=0$, linearizing the SDE around $X^*$, we obtain the same equation for $(X(t)-X^*)/\sqrt\gamma$. Further, for stochastic gradient descent on many particles, \cite[Theorem~1.6]{GGK25} shows that the equation with noise approximates the dynamics more closely than the one without.
\end{remark}

\runin{From the orders to the volume}We now sum the SDE over all orders to obtain an equation for the volume of the book. We want this equation to have a unique solution, as the equation then describes the same book as the orders. Intuitively, each order moves by an SDE whose coefficients depend on the whole book. With It\^o's formula we turn the sum of these moves into an equation for the measure of all orders. Conversely, \cite{GGK25} shows that every solution of such an equation started from finitely many point masses arises in this way. Our book enters the framework of \cite{GGK25} once the indices become a coordinate and the memory is frozen, such that uniqueness reduces to that of our SDE. We first state this framework in full.

Let $D\ge1$, and let $\Theta$ be a measure space with a finite measure $\vartheta$. Let $W$ be a cylindrical Wiener process on $L^2(\Theta,\vartheta)$, that is, a family of independent standard Brownian motions $W^e$ indexed by an orthonormal basis $(e)$ of $L^2(\Theta,\vartheta)$. Further, let $(\mathcal F_t)$ be the filtration generated by $W$, completed by all subsets of events of probability zero and made right-continuous, $\mathcal F_t=\bigcap_{s>t}\mathcal F_s$. Here, completion ensures that changing a process on an event of probability zero keeps it adapted. Right-continuity in turn ensures that the information at time $t$ includes what is revealed immediately after. Then, the stochastic integral of an adapted process $g_t\in L^2(\Theta,\vartheta)$ is
\begin{equation*}
 \int_0^t\!\!\int_\Theta g_s(\varsigma)\,W(d\varsigma,ds)=\sum_e\int_0^t\Bigl(\int_\Theta g_s\,e\,d\vartheta\Bigr)dW^e_s .
\end{equation*}
We compare measures by the Wasserstein distance on the set $\mathcal P_2(\mathbb R^D)$ of probability measures with finite second moment,
\begin{equation*}
 W_2(\nu,\nu')=\inf\Bigl\{\bigl(\mathbb E|Y-Y'|^2\bigr)^{1/2}:\ Y\sim\nu,\ Y'\sim\nu'\Bigr\}.
\end{equation*}
Lastly, the coefficients are a velocity $\mathcal V(t,x,\nu)\in\mathbb R^D$ and a noise coefficient $G(t,x,\nu,\varsigma)$ with $G(t,x,\nu,\cdot)\in L^2(\Theta,\vartheta)^D$, both possibly random, from which the diffusion matrix follows as
\begin{equation*}
 A(t,x,\nu)=\Bigl(\int_\Theta G_aG_b\,d\vartheta\Bigr)_{a,b\le D}.
\end{equation*}
The noise of the equation below is thus white in time and has the spatial covariance
\begin{equation*}
 \int_\Theta G(x,\varsigma)\,G(x',\varsigma)^{\top}\vartheta(d\varsigma)
\end{equation*}
between two points $x$ and $x'$.

\begin{definition}[{Solutions \cite[Definitions~2.2 and~2.6]{GGK25}}]\label{def:solutions}

\begin{enumerate}[label=(\roman*),leftmargin=*]
\item A continuous $(\mathcal F_t)$-\hspace{0pt}adapted process $\nu_t$ in $\mathcal P_2(\mathbb R^D)$ is a \emph{solution} if, for every $\varphi\in C^2_c(\mathbb R^D)$ and $t\ge0$, almost surely
\begin{equation*}
 \begin{aligned}
\int\varphi\,d\nu_t=\int\varphi\,d\nu_0&+\int_0^t\!\!\int\nabla\varphi\cdot \mathcal V(s,\cdot,\nu_s)\,d\nu_s\,ds\\
&+\frac12\int_0^t\!\!\int D^2\varphi:A(s,\cdot,\nu_s)\,d\nu_s\,ds\\
&+\int_0^t\!\!\int_\Theta\!\Bigl(\int\nabla\varphi\cdot G(s,\cdot,\nu_s,\varsigma)\,d\nu_s\Bigr)W(d\varsigma,ds).
\end{aligned}
\end{equation*}
\item A solution started from $\nu_0=\sum_{l=1}^J\alpha_l\delta_{x^l}$ is a \emph{superposition solution} if $\nu_t=\sum_{l=1}^J\alpha_l\delta_{X^l(t)}$, where the points $X^l$ solve the coupled SDEs
\begin{equation*}
 dX^l(t)=\mathcal V\bigl(t,X^l(t),\nu_t\bigr)\,dt+\int_\Theta G\bigl(t,X^l(t),\nu_t,\varsigma\bigr)\,W(d\varsigma,dt),\qquad X^l(0)=x^l .
\end{equation*}
\end{enumerate}
\end{definition}

In a superposition solution each point mass moves by an SDE whose coefficients see the whole measure. Applying It\^o's formula to $\varphi(X^l)$ and summing over $l$ with the weights $\alpha_l$, we obtain the identity in (i) \cite[Lemma~2.13]{GGK25}. The identity holds for any measure, while the coupled SDEs require point masses. Uniqueness therefore asks whether every solution started from point masses is of this form, which \cite[Theorem~2.16]{GGK25} proves under two conditions.

\begin{theorem}[{\cite[Theorem~2.16]{GGK25}}]\label{thm:ggk}
Assume that
\begin{enumerate}[label=(G\arabic*),leftmargin=*]
\item for every $t$, the restrictions of $\mathcal V$ and $G$ to $[0,t]$ are $\mathcal B([0,t])\otimes\mathcal B(\mathbb R^D)\otimes\mathcal B(\mathcal P_2)\otimes\mathcal F_t$-measurable and bounded on compact sets;
\item for every $T>0$ there is $L_G$ such that almost surely, for all $t\le T$, $x,x'\in\mathbb R^D$ and $\nu,\nu'\in\mathcal P_2(\mathbb R^D)$,
\begin{equation*}
 |\mathcal V(t,x,\nu)-\mathcal V(t,x',\nu')|+\|G(t,x,\nu,\cdot)-G(t,x',\nu',\cdot)\|_{L^2(\vartheta)}\le L_G\bigl(|x-x'|+W_2(\nu,\nu')\bigr),
\end{equation*}
\begin{equation*}
 |\mathcal V(t,0,\delta_0)|+\|G(t,0,\delta_0,\cdot)\|_{L^2(\vartheta)}\le L_G .
\end{equation*}
Let $\nu_0$ consist of $J$ point masses. Then there is a unique solution started from $\nu_0$ with $\sup_{t\le T}\int|x|^{J^2+J}\,d\nu_t<\infty$ almost surely for every $T$, and it is a superposition solution.
\end{enumerate}
\end{theorem}

Condition (G1) makes the integrands adapted, such that the stochastic integrals are defined. Condition (G2) in turn makes the coupled SDEs well posed. In \cite{GGK25}, these SDEs are solved by freezing the measure, solving, updating the measure and repeating. Condition (G2) makes the errors of this iteration shrink like $(CT)^m/m!$. For uniqueness, \cite{GGK25} uses the functional
\begin{equation*}
 \mathcal D(\nu)=\int\prod_{l\ne l'}|z^l-z^{l'}|\;\nu(dz^1)\cdots\nu(dz^{J+1}),
\end{equation*}
which is the expected product of all distances between $J+1$ points drawn independently from $\nu$. If $\nu$ has at most $J$ point masses, two of the $J+1$ points coincide and $\mathcal D(\nu)=0$, while otherwise $\mathcal D(\nu)>0$. With an It\^o formula for functions of finitely many integrals $\int\varphi_l\,d\nu$, (G2) and Gronwall's inequality, \cite{GGK25} obtains
\begin{equation*}
 \mathbb E\,\mathcal D(\nu_t)\le C\int_0^t\mathbb E\,\mathcal D(\nu_s)\,ds,\qquad\mathcal D(\nu_0)=0,
\end{equation*}
such that $\mathcal D(\nu_t)=0$ and every solution keeps at most $J$ point masses \cite[Proposition~2.18 and Lemma~2.21]{GGK25}. Here, the moment bound of order $J^2+J$ makes $\mathcal D(\nu_t)$ integrable. Lastly, the proof of \cite[Theorem~2.16]{GGK25} shows that each of these point masses moves by the coupled SDEs, whose solution is unique by (G2). The solution is therefore the superposition solution.

\runin{The conditions for our book}We now place our book into this framework and check its two conditions. The indices become a second coordinate, $\Theta$ becomes the set of desks, and the memory enters as a frozen input. Let $J$ be the number of orders. We place each index $(i,j)$ at its own point $q_{i,j}\in\mathbb R$, and we write $\sigma(q)$, $y^*(q)$, $k(q)$ and $\mathbf 1_i(q)$ for the side, signed target, cluster and desk indicator of the order at $q$. Between the points $q_{i,j}$ we interpolate these functions linearly, which are extended outside this range as constants. The book then becomes the probability measure
\begin{equation*}
 \nu_t=\frac12\sum_im_i\sum_{j=1}^{L}\delta_{(y_{i,j}(t),\,q_{i,j})}\qquad\text{on }\mathbb R^2,
\end{equation*}
whose second coordinate never moves. Further, let $\Theta$ be the set of desks with counting measure, such that $W(\{d_i\},t)=W_i(t)$. Let $\bar y=\mathrm{pr}(y,q)$ clip $y$ to $[y^*(q)-R_T,\,y^*(q)+R_T]$, and let $\bar\chi=(\bar\pi,\bar u)$ clip a memory value $\chi=(\pi,u)$ to the box $K_T$. In terms of $\nu$, the quantities $b$, $v$ and $a_i$ then read
\begin{equation*}
 \begin{aligned}
b(y,q,\chi)&=-\bigl(\bar y-y^*(q)\bigr)+\Pi\,\bar\pi_{k(q)},\\
v(\nu)&=-2\bar c\int\sigma(q)\,e^{-\sigma(q)\bar y/\ell}\,\nu(dy,dq),\\
a_i(\nu,\chi)&=-\frac{2\sigma_i}{\ell\,m_i}\int\mathbf 1_i(q)\,e^{-\sigma(q)\bar y/\ell}\,b(y,q,\chi)\,\nu(dy,dq).
\end{aligned}
\end{equation*}
For a memory path $\chi(t)=(\pi(t),u(t))$ we define the coefficients with frozen memory
\begin{equation*}
 \begin{aligned}
 \mathcal V^\chi(t,(y,q),\nu)&=\begin{pmatrix}p_{k(q)}\bigl(\bar u(t)\bigr)\,b\bigl(y,q,\chi(t)\bigr)-v(\nu)\\0\end{pmatrix},\\
 G^\chi(t,(y,q),\nu,d_i)&=\begin{pmatrix}\sqrt\gamma\,\mathbf 1_i(q)\,s_{k(q)}\bigl(\bar u(t)\bigr)\,b\bigl(y,q,\chi(t)\bigr)\\0\end{pmatrix},
 \end{aligned}
\end{equation*}
where $p_k(u)=p_0+\bar\beta u_k$ and $s_k(u)=\sqrt{p_k(u)(1-p_k(u))}$. At the point masses of $\nu$, these coefficients are the order components of $\tilde F$ and $\sqrt\gamma\,\tilde S_i$.

\begin{lemma}[Frozen memory]\label{lem:frozen}
Let $\chi$ be a continuous $(\mathcal F_t)$-adapted process in $\mathbb R^{2K}$. Then $\mathcal V^\chi$ and $G^\chi$ satisfy (G1) and (G2) with a constant $L_G$ that does not depend on $\chi$.
\end{lemma}

\begin{proof}
For (G1), the coefficients have the form $\mathcal V^\chi(t,z,\nu,\omega)=\Psi\bigl(z,\nu,\chi(t,\omega)\bigr)$ with a continuous function $\Psi$, and similarly for $G^\chi$. A continuous adapted process is progressively measurable. Hence for every $t$ the map
\begin{equation*}
 (s,z,\nu,\omega)\mapsto\bigl(z,\nu,\chi(s,\omega)\bigr)\mapsto\Psi\bigl(z,\nu,\chi(s,\omega)\bigr),\qquad s\le t,
\end{equation*}
is $\mathcal B([0,t])\otimes\mathcal B(\mathbb R^2)\otimes\mathcal B(\mathcal P_2)\otimes\mathcal F_t$-measurable. Moreover, the coefficients are bounded, as all their factors are evaluated at clipped arguments.

For (G2) we use that a product of bounded Lipschitz functions is Lipschitz,
\begin{equation*}
 |fg(z)-fg(z')|\le\bigl(\|f\|_\infty\,\mathrm{Lip}(g)+\|g\|_\infty\,\mathrm{Lip}(f)\bigr)|z-z'| .
\end{equation*}
The factors $\sigma(q)$, $y^*(q)$, $\mathbf 1_i(q)$ and $p_{k(q)}(\bar u)$ are bounded and Lipschitz in $q$ by the linear interpolation, with constants set by the smallest gap between the points $q_{i,j}$. Similarly, the factor $\bar y$ is $1$-Lipschitz in $y$ and bounded, such that $b$ is bounded and Lipschitz in $(y,q)$. Further, $s_{k(q)}(\bar u)$ is bounded and Lipschitz as $p_{k(q)}(\bar u)\in[p_0,\bar p]$. Hence
\begin{equation*}
 |\mathcal V^\chi(t,z,\nu)-\mathcal V^\chi(t,z',\nu)|+\|G^\chi(t,z,\nu,\cdot)-G^\chi(t,z',\nu,\cdot)\|_{L^2(\vartheta)}\le L_1|z-z'| .
\end{equation*}
Lastly, the coefficients depend on the measure only through $v(\nu)$, whose integrand $f(y,q)=\sigma(q)e^{-\sigma(q)\bar y/\ell}$ is bounded and Lipschitz by the same argument. For any $Y\sim\nu$ and $Y'\sim\nu'$, we obtain from the Cauchy--Schwarz inequality
\begin{equation*}
 |v(\nu)-v(\nu')|=2\bar c\,\bigl|\mathbb E f(Y)-\mathbb E f(Y')\bigr|\le2\bar c\,\mathrm{Lip}(f)\,\mathbb E|Y-Y'|\le2\bar c\,\mathrm{Lip}(f)\,\bigl(\mathbb E|Y-Y'|^2\bigr)^{1/2}.
\end{equation*}
Taking the infimum over $(Y,Y')$, we obtain $|v(\nu)-v(\nu')|\le2\bar c\,\mathrm{Lip}(f)\,W_2(\nu,\nu')$. With $L_G=L_1+2\bar c\,\mathrm{Lip}(f)+\sup|\mathcal V^\chi|+\sup\|G^\chi\|$, both inequalities of (G2) hold. None of these constants involves $\chi$, since $\chi$ enters only through clipped values. 
\end{proof}

The book itself generates its memory. We now build this property into the notion of solution and call a pair $(\nu,\chi)$ a \emph{coupled solution} if $\nu$ solves Definition~\ref{def:solutions}(i) with the coefficients $(\mathcal V^\chi,G^\chi)$ and the moment bound (Theorem~\ref{thm:ggk}), and if $\chi=(\pi,u)$ solves
\begin{equation*}
 \begin{aligned}
d\pi_k&=\frac{v(\nu_t)-\pi_k}{\tau_k}\,dt,\\
du_k&=\frac{1}{\tau_k}\Bigl(\bar c\sum_ip_{k(i)}(\bar u)\,m_i\,\bigl|a_i(\nu_t,\chi)\bigr|-u_k\Bigr)dt+\frac{\sqrt\gamma}{\tau_k}\sum_is_{k(i)}(\bar u)\,\bar c\,m_i\,\bigl|a_i(\nu_t,\chi)\bigr|\,dW_i,
\end{aligned}
\end{equation*}
from $\pi_k(0)$ and $u_k(0)$ of the initial state.

\begin{proposition}[A variant of the Dean--Kawasaki equation]\label{prop:dk}
Let $X$ solve the SDE with the extended coefficients, and let $\nu_t$ be the measure of its orders. Then $(\nu,(\pi,u))$ is a coupled solution, and for every $\varphi\in C^2_b(\mathbb R)$
\begin{equation}\label{eq:weakform}
 \begin{split}
 d\sum_im_i\sum_{j=1}^{L}\varphi(y_{i,j})&=\sum_im_i\sum_{j=1}^{L}\Bigl(\varphi'(y_{i,j})\mathcal V_{i,j}+\frac{\gamma}{2}\varphi''(y_{i,j})G_{i,j}^2\Bigr)dt\\
 &\quad+\sqrt\gamma\sum_i\Bigl(m_i\sum_{j=1}^{L}\varphi'(y_{i,j})G_{i,j}\Bigr)dW_i,
 \end{split}
\end{equation}
with $\mathcal V_{i,j}=p_{k(i)}b_{i,j}-v$ and $G_{i,j}=s_{k(i)}b_{i,j}$ at the clipped state. In particular, the total volume $\sum_iLm_i=2$ is constant. Every coupled solution equals $(\nu,(\pi,u))$.
\end{proposition}

Integrating by parts in $y$, we obtain for the volume $\mu_t=2\nu_t$ a variant of the Dean--Kawasaki equation \cite{Dea96,Kaw94}
\begin{equation*}
 d\mu_t=-\partial_y\bigl(\mathcal V\mu_t\bigr)\,dt+\frac{\gamma}{2}\,\partial_y^{2}\bigl(G^2\mu_t\bigr)\,dt-\sqrt{\gamma}\sum_i\partial_y\bigl(G\,\mathbf 1_i\,\mu_t\bigr)\,dW_i .
\end{equation*}
Every term on its right side is a derivative in $y$, such that the equation is conservative.

\begin{proof}
By the SDE, each order moves as $dy_{i,j}=\mathcal V_{i,j}\,dt+\sqrt\gamma\,G_{i,j}\,dW_i$, and we obtain from It\^o's formula
\begin{equation*}
 d\varphi(y_{i,j})=\Bigl(\varphi'(y_{i,j})\mathcal V_{i,j}+\frac{\gamma}{2}\varphi''(y_{i,j})G_{i,j}^2\Bigr)dt+\sqrt\gamma\,\varphi'(y_{i,j})G_{i,j}\,dW_i .
\end{equation*}
Multiplying by $m_i$ and summing over all orders, we obtain \eqref{eq:weakform}. Choosing $\varphi\equiv1$, we see that every term on the right vanishes, so the total volume is constant. Since $\mathcal V_{i,j}$ and $\sqrt\gamma G_{i,j}$ equal $\mathcal V^\chi$ and $G^\chi(\cdot,d_i)$ at the point $(y_{i,j},q_{i,j})$, the same computation for $\varphi\in C^2_c(\mathbb R^2)$ gives the identity (Definition~\ref{def:solutions}(i)). Further, the moment bound holds because the coefficients are bounded, while the memory equations are the $\pi$ and $u$ rows of the SDE. Hence $(\nu,(\pi,u))$ is a coupled solution.

For uniqueness, let $(\nu',\chi')$ be a coupled solution. Freezing $\chi'$, we obtain from Lemma~\ref{lem:frozen} and Theorem~\ref{thm:ggk} that $\nu'$ is the superposition solution for the coefficients $(\mathcal V^{\chi'},G^{\chi'})$,
\begin{equation*}
 \nu'_t=\frac12\sum_im_i\sum_{j=1}^{L}\delta_{(y'_{i,j}(t),\,q_{i,j})},\qquad dy'_{i,j}=\mathcal V^{\chi'}\bigl(t,(y'_{i,j},q_{i,j}),\nu'_t\bigr)dt+G^{\chi'}\bigl(t,(y'_{i,j},q_{i,j}),\nu'_t,d_i\bigr)dW_i,
\end{equation*}
where the second coordinate stays at $q_{i,j}$ as its coefficients vanish. The state $X'=(y'_{i,j},\pi',u')$ therefore solves the order rows of the SDE, while $\chi'$ solves its memory rows. Hence $X'$ solves the SDE with the extended coefficients. These coefficients are Lipschitz, such that the SDE has a unique strong solution. Consequently $X'=X$ almost surely and $(\nu',\chi')=(\nu,(\pi,u))$. 
\end{proof}

\begin{remark}[Latent information]\label{rem:latent}
In our derivation we only relied on the observation that the variables driving the fluctuation are independent across steps. Suppose that at every step a latent signal $\Xi_n\sim\vartheta$, drawn independently of the past, also enters the protocol. It could enter through the targets $x^*_{i,j}+\upsilon(x_{i,j},\Xi_n)$ or through the probabilities of acting $p_{k(i)}\varrho_i(\Xi_n)$ with $\int\varrho_i\,d\vartheta=1$. The fluctuation then has the covariance $\int_\Theta S(\varsigma)S(\varsigma)^{\top}\vartheta(d\varsigma)$. If the resulting coefficients satisfy (G1) and (G2), the proofs of Theorem~\ref{thm:limit} and Proposition~\ref{prop:dk} carry over with $W(d\varsigma,dt)$ in place of the $W_i$. The noise of the book is then white in time with any spatial covariance
\begin{equation*}
 C(y,y')=\int_\Theta e(y,\varsigma)\,e(y',\varsigma)\,\vartheta(d\varsigma),
\end{equation*}
where $e(y,\varsigma)$ is the fluctuation of an order at signed distance $y$ under the signal $\varsigma$. The protocol of Section 2 is the case in which $\Theta$ is the set of desks.
\end{remark}

We next follow how the conservative equation moves volume between distances.

\subsection{Conservation}\label{sec:conservation}
In the queueing models of the order book \cite{AJ13,BBDG18,CdL13,CST10,GPWMFH13,HLR15,SFGK03}, the mid price moves when the queue of orders at the best bid or the best ask is depleted, or when an order arrives inside the spread. In the latent order book of \cite{DBMB15} and in the continuum limits of \cite{HKN20}, the volume spreads over a continuum of prices, while the price remains the boundary between the bids and the asks, where volume is depleted. In our book the volume of each side also spreads over a continuum of distances, while no order reaches the mid price under Assumption~\ref{ass:interior}(b), such that no volume is depleted at any price. As each desk keeps its side, the volume of each side moreover stays one, as normalized in Section~\ref{sec:book}. Each side is therefore a probability distribution over distances, and the mid price can move only through the shape of these two distributions near it. We now show how the two distributions evolve, how their evolution changes $\theta_n$, and how the motion of the mid price in turn reshapes the two distributions.

\runin{The mid price and the shape of the book}We first write the move of the mid price through the two distributions, such that its dependence on their shape becomes explicit. With the measure $V^\sigma_n$ (Section~\ref{sec:book}), the volume of side $\sigma$ between the distances $a$ and $b$ at step $n$ is
\begin{equation*}
 V^\sigma_n\bigl((a,b]\bigr)=\sum_{i:\,\sigma_i=\sigma}m_i\sum_{j=1}^{L}\mathbf 1\bigl\{x_{i,j}(n)\in(a,b]\bigr\},\qquad0\le a<b.
\end{equation*}
Then \eqref{eq:near} and \eqref{eq:mid} read
\begin{equation*}
 Z^\sigma_n=\int_0^\infty e^{-x/\ell}\,V^\sigma_n(dx),\qquad\theta_n=-c\,\bigl(Z^+_n-Z^-_n\bigr),
\end{equation*}
such that $Z^\sigma_n$ is the Laplace transform of the distribution of side $\sigma$ at $1/\ell$. One may picture the two sides as two fluids that press on the mid price from either side, each with the volume it holds near the mid price. The mid price then moves away from the side that presses harder. Let the asks and their targets follow, for instance, a Gamma distribution with shape $2$ and scale $1.5$, and let the bids and their targets be uniform on $[1,11]$. We then obtain
\begin{equation*}
 Z^+=\Bigl(1+\frac{1.5}{\ell}\Bigr)^{-2},\qquad Z^-=e^{-1/\ell}\,\frac{\ell}{10}\bigl(1-e^{-10/\ell}\bigr),
\end{equation*}
such that $Z^+=1/4$ and $Z^-\approx0.077$ for $\ell=1.5$. The asks thus press harder, and by \eqref{eq:mid} the mid price moves toward the bids with $\theta_n\approx-0.17\,c$.

The move of the mid price in turn reshapes both distributions. By \eqref{eq:shift} every ask moves away from the mid price by $|\theta_n|$, and every bid moves closer by $|\theta_n|$. As a shift of a whole side by $\Delta$ multiplies its Laplace transform by $e^{-\Delta/\ell}$, the shift lowers $Z^+$ and raises $Z^-$. Without the desks, the two sides would press equally after a total shift $\Delta^*$ with
\begin{equation*}
 e^{-\Delta^*/\ell}Z^+=e^{\Delta^*/\ell}Z^-,\qquad\Delta^*=\frac{\ell}{2}\ln\frac{Z^+}{Z^-}\approx0.88,
\end{equation*}
before any bid reaches the mid price. Yet each acting desk moves its orders back toward their targets, such that the asks again press harder. By the gradient rule (Section~\ref{sec:rule}), the desks also move their orders with the trends $P_k$, which turn negative as the mid price falls. The asks then move closer and the bids away by $\Pi|P_k|$, and the mid price falls faster. The shape of the two distributions thus sets the move of the mid price, which reshapes the two distributions in turn. Throughout, $\theta_n$ changes only because volume moves between distances, as the volume of each side stays one.

\runin{Flows in the protocol}In the protocol each step moves finitely many point masses, such that we can count the volume that moves between distances exactly. For a distance $r\ge0$, let
\begin{equation*}
 \Phi^\sigma_n(r)=\sum_{i:\,\sigma_i=\sigma}m_i\sum_{j=1}^{L}\Bigl(\mathbf 1\bigl\{x_{i,j}(n)\le r<x_{i,j}(n+1)\bigr\}-\mathbf 1\bigl\{x_{i,j}(n+1)\le r<x_{i,j}(n)\bigr\}\Bigr)
\end{equation*}
denote the flow of side $\sigma$ through $r$. Its first term is the volume of the orders that move across $r$ away from the mid price in step $n$, while its second term is the volume of those that move across $r$ toward it.

\begin{lemma}[Flow identity of the protocol]\label{lem:flow}
For every side $\sigma$, every step $n$ and all $0\le a<b$,
\begin{equation}\label{eq:flow}
 V^\sigma_{n+1}\bigl((a,b]\bigr)=V^\sigma_n\bigl((a,b]\bigr)+\Phi^\sigma_n(a)-\Phi^\sigma_n(b).
\end{equation}
Under Assumption~\ref{ass:interior}(b), moreover, $\Phi^\sigma_n(0)=0$. Under the assumptions of Lemma~\ref{lem:bounded}, $\Phi^\sigma_n(r)=0$ for $n\gamma\le T$ and $r>r_T=\max_{i,j}x^*_{i,j}+R_T$.
\end{lemma}

The volume between $a$ and $b$ at the next step is thus its present volume, plus the flow into the interval across $a$, minus the flow out of it across $b$. In particular, no volume flows through the mid price or far from it.

\begin{proof}
For a single order with $x=x_{i,j}(n)$ and $x'=x_{i,j}(n+1)$, we split the indicator of $(a,b]$ at $a$ and $b$,
\begin{equation*}
 \begin{aligned}
\mathbf 1\{x'\in(a,b]\}-\mathbf 1\{x\in(a,b]\}&=\bigl(\mathbf 1\{x'>a\}-\mathbf 1\{x>a\}\bigr)-\bigl(\mathbf 1\{x'>b\}-\mathbf 1\{x>b\}\bigr),\\
\mathbf 1\{x'>r\}-\mathbf 1\{x>r\}&=\mathbf 1\{x\le r<x'\}-\mathbf 1\{x'\le r<x\}.
\end{aligned}
\end{equation*}
Multiplying by $m_i$ and summing over the orders of side $\sigma$, we obtain \eqref{eq:flow}. Under Assumption~\ref{ass:interior}(b) every order keeps $x,x'>0$, such that both indicators in $\Phi^\sigma_n(0)$ vanish. Lastly, from Lemma~\ref{lem:bounded} we obtain $x,x'\le r_T$ for $n\gamma\le T$. For $r>r_T$ every indicator in $\Phi^\sigma_n(r)$ therefore vanishes. 
\end{proof}

Notice that the identity telescopes over adjacent intervals. For distances $0\le r_0<r_1<\dots<r_m$ we obtain
\begin{equation*}
 \sum_{l=1}^{m}\Bigl(V^\sigma_{n+1}\bigl((r_{l-1},r_l]\bigr)-V^\sigma_n\bigl((r_{l-1},r_l]\bigr)\Bigr)=\Phi^\sigma_n(r_0)-\Phi^\sigma_n(r_m),
\end{equation*}
such that volume leaving one interval through $r_l$ enters the adjacent one. The identity \eqref{eq:flow} thus has the conservation form of finite-volume schemes \cite[Section~4.1]{LeV02}, in which the quantity held in each interval changes only through the flows across its two endpoints. In deriving \eqref{eq:flow} we only relied on the observation that each step moves point masses. The identity therefore holds for any sampling, including the latent signals (Remark~\ref{rem:latent}).

We now turn to the mid price. Since $Z^\sigma_n$ weights every order by its distance from the mid price, we write this weight as an integral over the distances between the order and the mid price.

\begin{proposition}[Motion of the mid price]\label{prop:mid}
Under Assumption~\ref{ass:interior}(b), for every step $n$,
\begin{equation*}
 \theta_{n+1}-\theta_n=c\int_0^\infty\frac{e^{-r/\ell}}{\ell}\,\bigl(\Phi^+_n(r)-\Phi^-_n(r)\bigr)\,dr .
\end{equation*}
\end{proposition}

The move of the mid price thus responds only to flows, each weighted by $e^{-r/\ell}$ according to its distance from the mid price. In the picture above, a side whose volume withdraws from the mid price presses less, such that the mid price moves toward it.

\begin{proof}
For $x>0$ we have
\begin{equation*}
 e^{-x/\ell}=\int_0^\infty\frac{e^{-r/\ell}}{\ell}\,\mathbf 1\{x\le r\}\,dr .
\end{equation*}
As every order satisfies $x>0$ under Assumption~\ref{ass:interior}(b), we obtain
\begin{equation*}
 Z^\sigma_n=\int_0^\infty\frac{e^{-r/\ell}}{\ell}\,V^\sigma_n\bigl((0,r]\bigr)\,dr .
\end{equation*}
By Lemma~\ref{lem:flow} with $\Phi^\sigma_n(0)=0$, the volume within $r$ evolves as $V^\sigma_{n+1}((0,r])=V^\sigma_n((0,r])-\Phi^\sigma_n(r)$. Hence
\begin{equation*}
 Z^\sigma_{n+1}-Z^\sigma_n=-\int_0^\infty\frac{e^{-r/\ell}}{\ell}\,\Phi^\sigma_n(r)\,dr,
\end{equation*}
and substituting into $\theta_{n+1}-\theta_n=-c\bigl((Z^+_{n+1}-Z^+_n)-(Z^-_{n+1}-Z^-_n)\bigr)$ proves the claim. 
\end{proof}

\runin{Flows in the continuum}In the protocol each order enters or leaves an interval at most once per step. Lemma~\ref{lem:flow} thus counts every change of the volume of the interval. In the continuum, an order driven by Brownian noise instead flickers back and forth across each distance it crosses. By \cite[Theorem~2.28]{MP10}, it returns to that distance infinitely often near each crossing time, such that the crossings cannot be counted and the volume of an interval has no It\^o differential. As for \eqref{eq:near} in Section~\ref{sec:book}, we therefore count the volume beyond a distance with a smooth step of width $\varepsilon$ rather than a sharp cutoff. An order within $\varepsilon$ of the distance then counts only partly, and its share changes smoothly, such that It\^o's formula applies. To cover the SDE (Theorem~\ref{thm:limit}) and its version with latent signals (Remark~\ref{rem:latent}) at once, we let the distances move as
\begin{equation*}
 dx_{i,j}=\eta_{i,j}\,dt+dN_{i,j},\qquad dN_{i,j}=\sum_{l}H_{(i,j),l}\,dW_l,
\end{equation*}
where the $W_l$ are finitely many independent Brownian motions and the coefficients $\eta_{i,j}$ and $H_{(i,j),l}$ are bounded and adapted. The noise of two orders then has the covariance rate $Q_{(i,j),(i',j')}=\sum_lH_{(i,j),l}H_{(i',j'),l}$, and we write $A_{i,j}=Q_{(i,j),(i,j)}$. For the SDE (Theorem~\ref{thm:limit}), since $x_{i,j}=\sigma_iy_{i,j}$, we have $\eta_{i,j}=\sigma_i\mathcal V_{i,j}$ with $\mathcal V_{i,j}$ (Proposition~\ref{prop:dk}), and the Brownian motions are the $W_i$ of the desks with $H_{(i,j),l}=\sigma_i\sqrt\gamma\,G_{i,j}\mathbf 1\{l=i\}$.

Let $\rho$ be a smooth even function with support in $[-1,1]$, with $\rho\ge0$ and $\int\rho=1$, and let
\begin{equation*}
 \rho_\varepsilon(z)=\frac1\varepsilon\,\rho\Bigl(\frac z\varepsilon\Bigr),\qquad\psi_\varepsilon(z)=\int_{-\infty}^{z}\rho_\varepsilon(u)\,du .
\end{equation*}
The function $\psi_\varepsilon$ rises smoothly from zero to one on $[-\varepsilon,\varepsilon]$. The sum $\sum_{i:\sigma_i=\sigma}m_i\sum_j\psi_\varepsilon(x_{i,j}-r)$ is therefore the volume of side $\sigma$ beyond $r$, smoothed over the width $\varepsilon$. We then define the smoothed flow of side $\sigma$ through $r$ by
\begin{equation}\label{eq:smoothflow}
 d\Phi^{\varepsilon,\sigma}_t(r)=\sum_{i:\,\sigma_i=\sigma}m_i\sum_{j=1}^{L}\Bigl(\rho_\varepsilon(x_{i,j}-r)\bigl(\eta_{i,j}\,dt+dN_{i,j}\bigr)+\frac12\,\rho_\varepsilon'(x_{i,j}-r)\,A_{i,j}\,dt\Bigr).
\end{equation}

\begin{proposition}[Flow identity in the continuum]\label{prop:flowcont}
For every side $\sigma$, every $\varepsilon>0$ and all $0\le a<b$, the smoothed indicator $\varphi^\varepsilon_{a,b}=\psi_\varepsilon(\cdot-a)-\psi_\varepsilon(\cdot-b)$ of $(a,b]$ satisfies
\begin{equation*}
 d\sum_{i:\,\sigma_i=\sigma}m_i\sum_{j=1}^{L}\varphi^\varepsilon_{a,b}(x_{i,j})=d\Phi^{\varepsilon,\sigma}_t(a)-d\Phi^{\varepsilon,\sigma}_t(b),
\end{equation*}
and the smoothed flow \eqref{eq:smoothflow} reads
\begin{equation*}
 \begin{aligned}
 d\Phi^{\varepsilon,\sigma}_t(r)&=\Bigl(\sum_{i:\,\sigma_i=\sigma}m_i\sum_{j=1}^{L}\rho_\varepsilon(x_{i,j}-r)\,\eta_{i,j}-\frac12\,\partial_r\sum_{i:\,\sigma_i=\sigma}m_i\sum_{j=1}^{L}\rho_\varepsilon(x_{i,j}-r)\,A_{i,j}\Bigr)dt\\
 &\quad+\sum_{i:\,\sigma_i=\sigma}m_i\sum_{j=1}^{L}\rho_\varepsilon(x_{i,j}-r)\,dN_{i,j}.
 \end{aligned}
\end{equation*}
\end{proposition}

The smoothed flow through $r$ thus depends only on the orders within $\varepsilon$ of $r$.

\begin{proof}
Since $\psi_\varepsilon'=\rho_\varepsilon$ and $\psi_\varepsilon''=\rho_\varepsilon'$, we obtain from It\^o's formula \cite[Theorem~IV.3.3]{RY99} for each order
\begin{equation*}
 d\psi_\varepsilon(x_{i,j}-r)=\rho_\varepsilon(x_{i,j}-r)\bigl(\eta_{i,j}\,dt+dN_{i,j}\bigr)+\frac12\,\rho_\varepsilon'(x_{i,j}-r)\,A_{i,j}\,dt .
\end{equation*}
Multiplying by $m_i$ and summing over the orders of side $\sigma$, we obtain $d\Phi^{\varepsilon,\sigma}_t(r)$. As $\varphi^\varepsilon_{a,b}$ is the difference of two such steps, the identity for $(a,b]$ follows. Lastly, since $\partial_r\rho_\varepsilon(x-r)=-\rho_\varepsilon'(x-r)$, we obtain
\begin{equation*}
 \frac12\sum_{i:\,\sigma_i=\sigma}m_i\sum_{j=1}^{L}\rho_\varepsilon'(x_{i,j}-r)\,A_{i,j}=-\frac12\,\partial_r\sum_{i:\,\sigma_i=\sigma}m_i\sum_{j=1}^{L}\rho_\varepsilon(x_{i,j}-r)\,A_{i,j},
\end{equation*}
which is the second form of the smoothed flow. 
\end{proof}

As in the protocol, the identity telescopes over adjacent intervals, such that volume leaving one interval enters the adjacent one for every width $\varepsilon$.

\runin{No volume crosses the mid price}In the protocol no volume flows through the mid price or far from it, while the noise of the SDE can carry an order across the mid price. By Theorem~\ref{thm:limit}, however, the SDE stays within $O(\sqrt\gamma)$ of the solution $\bar X$ of the ODE $d\bar X=\tilde F(\bar X)\,dt$. We show that a crossing therefore occurs only with probability $O(\gamma)$ whenever $\bar X$ keeps its orders away from the mid price. Let $\bar x_{i,j}=\sigma_i\bar y_{i,j}$ denote the distances under $\bar X$, and let $\bar r_T=\sup_{t\le T}\max_{i,j}\bar x_{i,j}(t)$.

\begin{corollary}[No volume crosses the mid price]\label{cor:sides}
Let the assumptions of Theorem~\ref{thm:limit} hold, and let $\delta>0$ satisfy $\bar x_{i,j}(t)\ge2\delta$ for every order and $t\le T$. Then the event $E_\delta=\{\sup_{t\le T}|X(t)-\bar X(t)|<\delta\}$ satisfies $\mathbb P(E_\delta^c)\le C\gamma/\delta^2$. On $E_\delta$, for every side $\sigma$, every $\varepsilon<\delta$ and $t\le T$,
\begin{enumerate}[label=(\roman*),leftmargin=*]
\item every order satisfies $x_{i,j}>\delta$, and the volume of each side stays one;
\item $d\Phi^{\varepsilon,\sigma}_t(r)=0$ for $r\ge\bar r_T+\delta+\varepsilon$.
\end{enumerate}
\end{corollary}

Up to an event of probability $O(\gamma)$, the SDE thus keeps every order away from the mid price and lets no volume flow far from it.

\begin{proof}
By Chebyshev's inequality and Theorem~\ref{thm:limit}(ii), we obtain
\begin{equation*}
 \mathbb P(E_\delta^c)\le\frac{1}{\delta^2}\,\mathbb E\sup_{t\le T}|X(t)-\bar X(t)|^2\le\frac{C\gamma}{\delta^2}.
\end{equation*}
On $E_\delta$ every order satisfies $|x_{i,j}-\bar x_{i,j}|<\delta$ for $t\le T$, such that
\begin{equation*}
 x_{i,j}>\bar x_{i,j}-\delta\ge\delta,\qquad x_{i,j}<\bar r_T+\delta .
\end{equation*}
Every order therefore keeps its side, such that the volume of each side stays $\sum_{i:\sigma_i=\sigma}Lm_i=1$. Further, every order stays farther than $\varepsilon$ from each $r\ge\bar r_T+\delta+\varepsilon$. At these distances every term of \eqref{eq:smoothflow} carries the factor $\rho_\varepsilon(x_{i,j}-r)=0$, such that the drift of the smoothed flow vanishes on $E_\delta$. Its noise is a continuous martingale with quadratic variation
\begin{equation*}
 \int_0^t\sum_{(i,j),(i',j')}m_im_{i'}\,\rho_\varepsilon(x_{i,j}-r)\,\rho_\varepsilon(x_{i',j'}-r)\,Q_{(i,j),(i',j')}\,ds,
\end{equation*}
which vanishes on $E_\delta$. By \cite[Proposition~IV.1.13]{RY99}, a continuous martingale is constant wherever its quadratic variation is constant, hence $d\Phi^{\varepsilon,\sigma}_t(r)=0$ at these distances. 
\end{proof}

\runin{The mid price in the continuum}We now show that Proposition~\ref{prop:mid} holds in the continuum as well. On the event $E_\delta$ every order stays farther than $\varepsilon$ from the mid price, such that smoothing the distances over the width $\varepsilon$ rescales every weight $e^{-x/\ell}$ by the same constant
\begin{equation*}
 \kappa_\varepsilon=\int\rho_\varepsilon(z)\,e^{-z/\ell}\,dz=1+O\Bigl(\frac{\varepsilon^2}{\ell^2}\Bigr).
\end{equation*}

\begin{proposition}[Motion of the mid price in the continuum]\label{prop:midcont}
Under the assumptions of Corollary~\ref{cor:sides}, on the event $E_\delta$, for every $\varepsilon<\delta$ and $t\le T$,
\begin{equation*}
 dv_t=\frac{\bar c}{\kappa_\varepsilon}\int_0^\infty\frac{e^{-r/\ell}}{\ell}\,\bigl(d\Phi^{\varepsilon,+}_t(r)-d\Phi^{\varepsilon,-}_t(r)\bigr)\,dr .
\end{equation*}
\end{proposition}

The velocity of the mid price thus responds only to the smoothed flows, with the weights (Proposition~\ref{prop:mid}) up to the factor $\kappa_\varepsilon$, which tends to one as $\varepsilon\to0$.

\begin{proof}
For $x\ge\varepsilon$, every $z$ in the support of $\rho_\varepsilon$ satisfies $x+z\ge0$. Writing $1-\psi_\varepsilon(x-r)=\psi_\varepsilon(r-x)$, which holds as $\rho$ is even, we obtain
\begin{equation*}
 \int_0^\infty\frac{e^{-r/\ell}}{\ell}\,\psi_\varepsilon(r-x)\,dr=\int\rho_\varepsilon(z)\int_{x+z}^\infty\frac{e^{-r/\ell}}{\ell}\,dr\,dz=\int\rho_\varepsilon(z)\,e^{-(x+z)/\ell}\,dz=\kappa_\varepsilon\,e^{-x/\ell}.
\end{equation*}
On $E_\delta$ every order satisfies $x_{i,j}>\delta>\varepsilon$. Summing over the orders of side $\sigma$ with $\sum_{i:\sigma_i=\sigma}Lm_i=1$, we therefore obtain
\begin{equation*}
 \kappa_\varepsilon Z^\sigma_t=\int_0^\infty\frac{e^{-r/\ell}}{\ell}\Bigl(1-\sum_{i:\,\sigma_i=\sigma}m_i\sum_{j=1}^{L}\psi_\varepsilon(x_{i,j}-r)\Bigr)dr .
\end{equation*}
The sum in the integrand is the smoothed volume of side $\sigma$ beyond $r$, whose differential is $d\Phi^{\varepsilon,\sigma}_t(r)$ by the proof of Proposition~\ref{prop:flowcont}. Since the integrand is bounded and continuous in $r$, and $\ell^{-1}e^{-r/\ell}dr$ is a probability measure, the stochastic Fubini theorem \cite[Theorem~2.2]{Ver12} allows us to interchange the integral over $r$ with the stochastic integrals, and we obtain
\begin{equation*}
 \kappa_\varepsilon\,dZ^\sigma_t=-\int_0^\infty\frac{e^{-r/\ell}}{\ell}\,d\Phi^{\varepsilon,\sigma}_t(r)\,dr .
\end{equation*}
Substituting into $v_t=-\bar c\,(Z^+_t-Z^-_t)$ proves the claim. 
\end{proof}

\runin{Feedback of the mid price}The shape of the two distributions sets the velocity $v$ of the mid price, while the motion of the mid price moves every order through the shift and through the trends $\pi_k$. We now write this loop as one equation for $v$. To this end we apply It\^o's formula to $v=-\bar c\sum_i\sigma_im_i\sum_je^{-x_{i,j}/\ell}$ directly, and we then split the drift of the distances into its causes.

\begin{proposition}[Feedback of the mid price]\label{prop:feedback}
For distances that move as in Proposition~\ref{prop:flowcont},
\begin{equation*}
 dv_t=\frac{\bar c}{\ell}\sum_i\sigma_im_i\sum_{j=1}^{L}e^{-x_{i,j}/\ell}\Bigl(\eta_{i,j}\,dt+dN_{i,j}-\frac{1}{2\ell}\,A_{i,j}\,dt\Bigr).
\end{equation*}
For the SDE (Theorem~\ref{thm:limit}) with the state in $K_T$, the drift of $v$ splits as
\begin{equation*}
 \begin{aligned}
 \frac{\bar c}{\ell}\Bigl(&-\bigl(Z^+_t+Z^-_t\bigr)\,v_t+\Pi\sum_{k=1}^{K}p_k\,\pi_k\,Z_{k,t}\\
 &+\sum_i\sigma_ip_{k(i)}m_i\sum_{j=1}^{L}e^{-x_{i,j}/\ell}\bigl(x^*_{i,j}-x_{i,j}\bigr)-\frac{1}{2\ell}\sum_i\sigma_im_i\sum_{j=1}^{L}e^{-x_{i,j}/\ell}A_{i,j}\Bigr),
 \end{aligned}
\end{equation*}
where $Z_{k,t}=\sum_{i:\,k(i)=k}m_i\sum_je^{-x_{i,j}/\ell}$ is the volume of cluster $k$ near the mid price.
\end{proposition}

The first term restores the balance of the two sides. A rising mid price shifts the asks toward it and the bids away from it, such that the asks press harder and $v$ relaxes at the rate $\bar c\,(Z^++Z^-)/\ell$. The second term reinforces the motion through the memory of the desks. A positive trend $\pi_k$ makes the desks of cluster $k$ move their asks away and their bids closer, which raises $v$. The trend in turn follows $v$ through $d\pi_k=(v-\pi_k)\,dt/\tau_k$. The third term pulls each side toward the shape of its targets, with a weight set by the probability of acting $p_k$. The fourth term arises from the fluctuations of the orders, as the convexity of $e^{-x/\ell}$ makes fluctuating orders press harder on average. Lastly, the noise term carries the random timing of the desks near the mid price into $v$. The shape of the two distributions thus sets $v$. In turn, $v$ reshapes the two distributions through the shift and the trends, and the reshaped distributions set the next velocity.

\begin{proof}
It\^o's formula for each distance gives
\begin{equation*}
 de^{-x_{i,j}/\ell}=-\frac1\ell\,e^{-x_{i,j}/\ell}\bigl(\eta_{i,j}\,dt+dN_{i,j}\bigr)+\frac{1}{2\ell^2}\,e^{-x_{i,j}/\ell}A_{i,j}\,dt .
\end{equation*}
Multiplying by $-\bar c\,\sigma_im_i$ and summing over all orders, we obtain the first display. For the SDE (Theorem~\ref{thm:limit}) on $K_T$, the drift of each distance reads
\begin{equation*}
 \eta_{i,j}=\sigma_i\bigl(p_{k(i)}b_{i,j}-v\bigr)=p_{k(i)}\bigl(x^*_{i,j}-x_{i,j}\bigr)+p_{k(i)}\,\sigma_i\Pi\pi_{k(i)}-\sigma_iv,
\end{equation*}
since $\sigma_ib_{i,j}=-(x_{i,j}-x^*_{i,j})+\sigma_i\Pi\pi_{k(i)}$. Substituting and using $\sigma_i^2=1$, the term $-\sigma_iv$ contributes $-(Z^++Z^-)\,v$ and the term $p_{k(i)}\sigma_i\Pi\pi_{k(i)}$ contributes $\Pi\sum_kp_k\pi_kZ_k$, which proves the split. 
\end{proof}

\begin{remark}[Smoothing and the grid of ticks]\label{rem:ticks}
A smooth step of width about one tick matches a real book, where every order sits on a grid of ticks \cite{BMP02}. A distance between two ticks is crossed only when a desk moves an order, as in the protocol. The infinitely many crossings of the continuum instead come from arbitrarily small moves below one tick, and thus only from treating distances as continuous in Section~\ref{sec:book}. The width enters Propositions~\ref{prop:flowcont} and~\ref{prop:midcont} only through $\kappa_\varepsilon=1+O(\varepsilon^2/\ell^2)$. The smoothing is therefore most accurate for small-tick stocks, whose spread, unlike that of large-tick stocks, often exceeds one tick \cite{HLR15}. Their volume spreads over many ticks, such that one tick is small against $\ell$ and $\kappa_\varepsilon$ is close to one.
\end{remark}

The two distributions thus move the mid price only through flows of volume, and Proposition~\ref{prop:feedback} closes the loop between their shape and the velocity $v$. Its noise term comes from the random timing of the desks, whose probabilities of acting $p_k$ grow with the estimates $U_k$ of market activity by \eqref{eq:batch}. Every action enters $I_n$ in \eqref{eq:push} and through \eqref{eq:memory} the estimates $U_k$, such that each action can raise the probability that other desks act in later steps. We study this self-excitation of the order flow next.

\section{From the order book to self-exciting order flow}\label{sec:flow}
\subsection{Linear response}\label{sec:response}
In Section~\ref{sec:conservation} we found that the noise of the mid price comes from the random timing of the desks. Yet this timing depends on the past. By \eqref{eq:batch} each desk acts with a probability that grows with its estimate $U_k$ of market activity, while by \eqref{eq:memory} this estimate is a moving average of the size of the updates $I_n$ in \eqref{eq:push}. The order flow thus excites itself, as measured in many order books \cite{FS12,HBB13}. Roughness requires that the order flow excites itself strongly, with a memory that fades as a power law \cite{EFR18,JR16}. Models of order flow usually choose this memory to fit the data \cite{BMM15}. In our market, \eqref{eq:memory} and \eqref{eq:batch} instead fix the horizons over which the desks average the size of the updates, while the book fixes how large each update is. We can therefore compute the memory of the order flow from the desks. We first state the structure that this memory takes.

\runin{Self-exciting sequences}To make this memory precise, we use the structure of a Hawkes process \cite{Haw71} in discrete time \cite{Kir16}. Unlike a Poisson process, whose events occur at a constant rate, a Hawkes process lets each event raise the rate of later events. As the size of the updates takes real values rather than counting events, we state this structure for real values.

\begin{definition}[Self-exciting sequence]\label{def:selfexc}
Let $(Y_n)_{n\ge0}$ be a stochastic process with values in $[0,\infty)$, such that $Y_n$ is $\mathcal F_{n+1}$-measurable for the $\sigma$-algebra $\mathcal F_n$ generated by $\xi(0),\dots,\xi(n-1)$ (Section~\ref{sec:spde}). We call $(Y_n)$ \emph{self-exciting} with base rate $\mu\ge0$ and kernel $\phi:\mathbb N\setminus\{0\}\to[0,\infty)$ with $\|\phi\|=\sum_{s\in\mathbb N\setminus\{0\}}\phi(s)<\infty$ if, for every step $n$,
\begin{equation*}
 \mathbb E\bigl[Y_n\mid\mathcal F_n\bigr]=\mu+\sum_{m=0}^{n-1}\phi(n-m)\,Y_m .
\end{equation*}
\end{definition}

With $\phi=0$ the expected value stays at $\mu$, as for a Poisson process. With $\phi>0$, each value $Y_m$ instead raises the expected value of $Y_{m+s}$ by $\phi(s)\,Y_m$, and of all later values together by $\|\phi\|\,Y_m$. The values raised in this way raise later values in turn. Repeating this argument, $Y_m>0$ raises all later values together in expectation by
\begin{equation*}
 Y_m\bigl(\|\phi\|+\|\phi\|^2+\|\phi\|^3+\cdots\bigr)=\begin{cases}Y_m\,\dfrac{\|\phi\|}{1-\|\phi\|},&\|\phi\|<1,\\[2mm]\infty,&\|\phi\|\ge1 .\end{cases}
\end{equation*}
Each value thus raises later values by a finite amount on average if $\|\phi\|<1$, while otherwise the values grow without bound.

\runin{Activity}The size of the updates grows with the number of desks that act in a step. A desk with more volume moves more of the book. We therefore count each desk by its volume. Let
\begin{equation*}
 \mathcal A_n=\sum_{k=1}^{K}w_k\,p_k(n)
\end{equation*}
denote the activity in step $n$, where $p_k(n)$ is the probability of acting \eqref{eq:batch} of cluster $k$. As the desks of cluster $k$ on each side together hold the volume $w_k$ by Section~\ref{sec:info}, the activity is the expected volume of the desks that act in step $n$ on each side.

\runin{Size of the updates per unit activity}The size of the updates of an acting desk depends on where its orders sit, as a step deep in the book changes $Z^\sigma$ little. Let $\Delta Z_i(n)$ denote the change $\Delta Z_i$ (Section~\ref{sec:rule}) that desk $d_i$ causes if it acts in step $n$, and let
\begin{equation*}
 \kappa_l(n)=\frac{c}{w_l}\sum_{i:\,k(i)=l}m_i\,\bigl|\Delta Z_i(n)\bigr|
\end{equation*}
denote the size of the updates per unit activity of cluster $l$.

\begin{lemma}[Size of the updates per unit activity]\label{lem:size}
For every step $n$,
\begin{equation*}
 \mathbb E\bigl[I_n\mid\mathcal F_n\bigr]=\sum_{l=1}^{K}\kappa_l(n)\,w_l\,p_l(n).
\end{equation*}
\end{lemma}

The expected size of the updates thus grows with the activity of every cluster, at a rate $\kappa_l$ that is large when the orders of cluster $l$ sit near the mid price.

\begin{proof}
By Section~\ref{sec:rule}, $\Delta Z_i(n)$ depends only on the distances of the orders of $d_i$ before its step and on the trend $P_{k(i)}$. Both are part of the state $X_n$, such that $\Delta Z_i(n)$ is $\mathcal F_n$-measurable. Starting from \eqref{eq:push} and using $\mathbb E[\xi_i(n)\mid\mathcal F_n]=p_{k(i)}(n)$, we obtain
\begin{equation*}
 \begin{aligned}
\mathbb E\bigl[I_n\mid\mathcal F_n\bigr]&=\sum_i\mathbb E\bigl[\xi_i(n)\mid\mathcal F_n\bigr]\,c\,m_i\bigl|\Delta Z_i(n)\bigr|\\
&=\sum_{l=1}^{K}p_l(n)\sum_{i:\,k(i)=l}c\,m_i\bigl|\Delta Z_i(n)\bigr|=\sum_{l=1}^{K}\kappa_l(n)\,w_l\,p_l(n),
\end{aligned}
\end{equation*}
where the second line groups the desks by cluster. 
\end{proof}

\runin{Activity driven by earlier updates}Lemma~\ref{lem:size} links the size of the updates in a step to the activity of the same step. To close the loop, we now link the activity to the size of the updates in earlier steps. By \eqref{eq:memory}, each cluster averages the size of the updates over its horizon $T_k$, and we let
\begin{equation*}
 \omega_k(s)=\frac{1}{T_k}\Bigl(1-\frac{1}{T_k}\Bigr)^{s-1},\qquad s\in\mathbb N\setminus\{0\},
\end{equation*}
denote the weight that this average gives to the size of the updates from $s$ steps ago. As $T_k>1$, these weights are positive, sum to one over $s$ and decay over about $T_k$ steps.

\begin{proposition}[Activity driven by earlier updates]\label{prop:activity}
Under Assumption~\ref{ass:interior}(a), for every step $n$,
\begin{equation*}
 \mathcal A_n=p_0+\beta\sum_{m=0}^{n-1}\Bigl(\sum_{k=1}^{K}w_k\,\omega_k(n-m)\Bigr)I_m .
\end{equation*}
\end{proposition}

The size of the updates in every step thus raises the activity of later steps through the memories of all clusters, each weighted by the volume of its cluster.

\begin{proof}
By Assumption~\ref{ass:interior}(a) we have $p_0+\beta U_k(n)=p_0+\bar\beta u_k(n)\le\bar p<1$, such that \eqref{eq:batch} reads $p_k(n)=p_0+\beta U_k(n)$. By \eqref{eq:memory} the estimates evolve as $U_k(n)=(1-1/T_k)\,U_k(n-1)+I_{n-1}/T_k$ from $U_k=I_{-1}=0$ in Section~\ref{sec:protocol}. We show by induction that
\begin{equation*}
 U_k(n)=\sum_{m=0}^{n-1}\omega_k(n-m)\,I_m .
\end{equation*}
For $n=0$ both sides vanish. If the identity holds for $n-1$, then using $(1-1/T_k)\,\omega_k(s)=\omega_k(s+1)$ and $\omega_k(1)=1/T_k$, we obtain
\begin{equation*}
 \begin{aligned}
U_k(n)&=\Bigl(1-\frac{1}{T_k}\Bigr)\sum_{m=0}^{n-2}\omega_k(n-1-m)\,I_m+\frac{I_{n-1}}{T_k}\\
&=\sum_{m=0}^{n-2}\omega_k(n-m)\,I_m+\omega_k(1)\,I_{n-1}.
\end{aligned}
\end{equation*}
Multiplying by $\beta w_k$, summing over $k$ and using $\sum_kw_k=1$ from \eqref{eq:weights}, we obtain the claim. 
\end{proof}

\runin{Two time scales}Lemma~\ref{lem:size} and Proposition~\ref{prop:activity} together make the activity depend on its own past, except that $\kappa_l(n)$ depends on the book. A larger size of the updates moves the mid price further. By \eqref{eq:shift}, every order then shifts from $x_{i,j}$ to $x_{i,j}-\sigma_i\theta_{n+1}>0$, and $\kappa_l$ changes with the book. We now compare how fast the book recovers from such a shift with how long the shift stays in the moving averages \eqref{eq:memory}.

\begin{lemma}[Two time scales]\label{lem:timescales}
Let Assumptions~\ref{ass:regime} and~\ref{ass:interior} hold with $\gamma<\tau_1$, and let $o_{i,j}$ be an order of a desk of cluster $k$. Then
\begin{enumerate}[label=(\roman*),leftmargin=*]
\item its gap to the target after step $n$ satisfies
\begin{equation*}
 \mathbb E\bigl[x^{\mathrm{new}}_{i,j}-x^*_{i,j}\mid\mathcal F_n\bigr]=\bigl(1-\gamma p_k(n)\bigr)\bigl(x_{i,j}-x^*_{i,j}\bigr)+p_k(n)\,\sigma_i\Pi P_k ;
\end{equation*}
\item for every $s\in\mathbb N\setminus\{0\}$, with $t=(s-1)\gamma$,
\begin{equation*}
 \frac{\gamma}{\tau_k}\,e^{-t/(\tau_k-\gamma)}\le\omega_k(s)\le\frac{\gamma}{\tau_k}\,e^{-t/\tau_k}.
\end{equation*}
\end{enumerate}
\end{lemma}

In expectation, each step thus shrinks the gap of an order to its target by the factor $1-\gamma p_k$, apart from the move with the trend $P_k$. Over the time $t$, that is, over $t/\gamma$ steps, the gap therefore shrinks by $(1-\gamma p_k)^{t/\gamma}\le e^{-p_kt}$. By (ii), the memory of cluster $k$ instead decays as $e^{-t/\tau_k}$.

\begin{proof}
For (i), by Assumption~\ref{ass:interior}(b) we have $x_{i,j}-\xi_i(n)\,\gamma\,\partial_j\mathcal L_i(x_i)>0$, such that \eqref{eq:sgd} reads $x^{\mathrm{new}}_{i,j}=x_{i,j}-\xi_i(n)\,\gamma\,\partial_j\mathcal L_i(x_i)$. By \eqref{eq:loss} we have $\gamma\,\partial_j\mathcal L_i(x_i)=\gamma(x_{i,j}-x^*_{i,j})-\sigma_i\Pi P_k$, and hence
\begin{equation*}
 \begin{aligned}
x^{\mathrm{new}}_{i,j}-x^*_{i,j}&=x_{i,j}-x^*_{i,j}-\xi_i(n)\Bigl(\gamma\bigl(x_{i,j}-x^*_{i,j}\bigr)-\sigma_i\Pi P_k\Bigr)\\
&=\bigl(1-\gamma\,\xi_i(n)\bigr)\bigl(x_{i,j}-x^*_{i,j}\bigr)+\xi_i(n)\,\sigma_i\Pi P_k .
\end{aligned}
\end{equation*}
As $x_{i,j}$ and $P_k$ are part of $X_n$, we obtain (i) by taking the conditional expectation with $\mathbb E[\xi_i(n)\mid\mathcal F_n]=p_k(n)$. For (ii), $T_k=\tau_k/\gamma$ by Assumption~\ref{ass:regime}, such that $\omega_k(s)=(\gamma/\tau_k)(1-\gamma/\tau_k)^{s-1}$. For $0\le y<1$ we have $-y/(1-y)\le\ln(1-y)\le-y$. With $y=\gamma/\tau_k$, multiplying by $s-1$ and taking exponentials, we obtain
\begin{equation*}
 e^{-(s-1)\gamma/(\tau_k-\gamma)}\le\Bigl(1-\frac{\gamma}{\tau_k}\Bigr)^{s-1}\le e^{-(s-1)\gamma/\tau_k},
\end{equation*}
which is (ii). 
\end{proof}

The book thus recovers from a shift within a time of order $1/p_k$, while the moving average \eqref{eq:memory} of cluster $k$ decays over a time of order $\tau_k$. Whenever $\tau_kp_k\gg1$, the book recovers its shape long before the moving averages decay. We therefore replace $\kappa_l(n)$ by its average $\kappa_l$ along the path, which we verify in Section~\ref{sec:simulation}. With these averages we write
\begin{equation}\label{eq:kernel}
 \phi(s)=\beta\sum_{l=1}^{K}\kappa_l\,w_l\,\omega_l(s),\qquad\bar\kappa=\sum_{l=1}^{K}\kappa_l\,w_l ,
\end{equation}
which combine the memories $\omega_l$, the volumes $w_l$ and the sizes of the updates per unit activity $\kappa_l$ of all clusters.

\begin{proposition}[Self-excitation]\label{prop:selfexc}
Let the assumptions of Proposition~\ref{prop:activity} hold, and let $\kappa_l(n)=\kappa_l$ for every cluster $l$ and step $n$. Then the size of the updates $(I_n)$ is self-exciting in the sense of Definition~\ref{def:selfexc}, with the base rate $\mu=p_0\bar\kappa$ and the kernel $\phi(s)=\beta\sum_l\kappa_lw_l\,\omega_l(s)$ of \eqref{eq:kernel}. In particular,
\begin{equation}\label{eq:mean}
 \mathbb E[I_n]=p_0\,\bar\kappa+\sum_{m=0}^{n-1}\phi(n-m)\,\mathbb E[I_m] .
\end{equation}
\end{proposition}

Although no kernel enters the protocol, the desks thus make the size of the updates excite itself, with a kernel built from their memories, their volumes and the book.

\begin{proof}
The size of the updates $I_n$ is nonnegative by \eqref{eq:push} and $\mathcal F_{n+1}$-measurable. Combining Lemma~\ref{lem:size} with the formula for $U_l(n)$ from the proof of Proposition~\ref{prop:activity}, we obtain
\begin{equation*}
 \begin{aligned}
\mathbb E\bigl[I_n\mid\mathcal F_n\bigr]&=\sum_{l=1}^{K}\kappa_l\,w_l\Bigl(p_0+\beta\sum_{m=0}^{n-1}\omega_l(n-m)\,I_m\Bigr)\\
&=p_0\,\bar\kappa+\sum_{m=0}^{n-1}\Bigl(\beta\sum_{l=1}^{K}\kappa_l\,w_l\,\omega_l(n-m)\Bigr)I_m ,
\end{aligned}
\end{equation*}
where the inner sum is $\phi(n-m)$ by \eqref{eq:kernel}. Taking expectations, we obtain \eqref{eq:mean}. 
\end{proof}

\runin{Response to one extra update}We now follow a single extra update through the market. To this end we add $E>0$ to $I_{m_0}$ in \eqref{eq:memory} and leave everything else unchanged. The estimates of step $m_0+1$ then receive $I_{m_0}+E$, and the estimates of every later step $n$ contain the additional term $E\,\omega_l(n-m_0)$. Let $\delta_n$ denote the difference of $\mathbb E[I_n]$ between the disturbed and the undisturbed market, with $\delta_n=0$ for $n\le m_0$. By \eqref{eq:mean}, we obtain for $s\in \mathbb N\setminus\{0\}$
\begin{equation*}
 \delta_{m_0+s}=E\,\phi(s)+\sum_{r=1}^{s-1}\phi(s-r)\,\delta_{m_0+r}.
\end{equation*}
Substituting this identity into its own right-hand side, we obtain
\begin{equation*}
 \begin{aligned}
\delta_{m_0+s}&=E\,\phi(s)+E\sum_{r=1}^{s-1}\phi(s-r)\,\phi(r)+\sum_{r=1}^{s-1}\sum_{q=1}^{r-1}\phi(s-r)\,\phi(r-q)\,\delta_{m_0+q}\\
&=E\,\phi^{*1}(s)+E\,\phi^{*2}(s)+E\,\phi^{*3}(s)+\cdots,
\end{aligned}
\end{equation*}
where $\phi^{*1}=\phi$ and $\phi^{*(j+1)}(s)=\sum_{r=1}^{s-1}\phi(s-r)\,\phi^{*j}(r)$. The term $E\,\phi^{*j}(s)$ is the contribution of $E$ through $j-1$ intermediate updates, and it vanishes for $j>s$. We therefore define the total response
\begin{equation*}
 R(s)=\sum_{j\in \mathbb N\setminus\{0\}}\phi^{*j}(s),\qquad s\in \mathbb N\setminus\{0\} .
\end{equation*}

\begin{corollary}[Response to one extra update]\label{cor:response}
Under the assumptions of Proposition~\ref{prop:selfexc}, adding $E$ to $I_{m_0}$ in \eqref{eq:memory} leaves $\mathbb E[I_n]$ unchanged for $n\le m_0$ and raises $\mathbb E[I_{m_0+s}]$ by $E\,R(s)$ for every $s\in \mathbb N\setminus\{0\}$.
\end{corollary}

The response to one extra update is thus fixed by $\phi$ alone, which involves only the horizons, the volumes, $\beta$ and the sizes of the updates per unit activity.

\begin{proof}
Up to step $m_0$ nothing changes. Splitting off the term $j=1$ in the definition of $R$, we obtain
\begin{equation*}
 \begin{aligned}
R(s)&=\phi(s)+\sum_{j\in \mathbb N\setminus\{0\}}\phi^{*(j+1)}(s)=\phi(s)+\sum_{r=1}^{s-1}\phi(s-r)\sum_{j\in \mathbb N\setminus\{0\}}\phi^{*j}(r)\\
&=\phi(s)+\sum_{r=1}^{s-1}\phi(s-r)\,R(r).
\end{aligned}
\end{equation*}
The difference $\delta_{m_0+s}$ and $E\,R(s)$ thus satisfy the same recursion. For $s=1$ both equal $E\,\phi(1)$, and by induction on $s$ they agree for every $s\in \mathbb N\setminus\{0\}$. 
\end{proof}

The desks thus produce a self-exciting order flow whose kernel we know explicitly. We next compute $\|\phi\|$ for our market, which decides whether the activity stays finite.

\subsection{Norm of the kernel and mean activity}\label{sec:kernel}
By Definition~\ref{def:selfexc}, the size of the updates stays finite on average only if $\|\phi\|<1$. By \cite{JR16}, roughness in turn requires self-excitation close to this threshold, with $\|\phi\|$ close to one. Whether our market can come close to the threshold depends on what sets $\|\phi\|$. We now compute $\|\phi\|$ in terms of the desks and the book, show how it fixes the mean activity, and write the kernel as a function of time.

\runin{Norm of the kernel}Each memory $\omega_l$ sums to one over $s$, such that summing \eqref{eq:kernel} over $s$ leaves only the sensitivity $\beta$ and the sizes of the updates per unit activity.

\begin{proposition}[Norm of the kernel]\label{prop:strength}
Under the assumptions of Proposition~\ref{prop:selfexc},
\begin{equation*}
 \|\phi\|=\beta\,\bar\kappa=\beta\sum_{l=1}^{K}\kappa_l\,w_l .
\end{equation*}
Moreover, a unit of activity of cluster $l$ in step $m$ raises the expected activity of all later steps together by $\beta\kappa_l$.
\end{proposition}

A unit of activity of cluster $l$ thus sets off on average $\beta\kappa_l$ units of activity in later steps. Since $\kappa_l$ is large when the orders of cluster $l$ sit near the mid price, a book with more volume near the mid price excites itself more strongly.

\begin{proof}
As $\sum_{s\in\mathbb N\setminus\{0\}}(1-1/T_l)^{s-1}=T_l$, each memory satisfies $\sum_s\omega_l(s)=1$. Summing \eqref{eq:kernel} over $s$, we obtain
\begin{equation*}
 \|\phi\|=\beta\sum_{l=1}^{K}\kappa_l\,w_l\sum_{s\in\mathbb N\setminus\{0\}}\omega_l(s)=\beta\sum_{l=1}^{K}\kappa_l\,w_l=\beta\,\bar\kappa .
\end{equation*}
For the second claim, by Lemma~\ref{lem:size} a unit of activity of cluster $l$ in step $m$ adds $\kappa_l$ to the expected size of the updates in step $m$. By Proposition~\ref{prop:activity} this addition raises the activity of step $n>m$ by $\beta\kappa_l\sum_kw_k\,\omega_k(n-m)$. Summing over $n>m$ and using $\sum_kw_k=1$ from \eqref{eq:weights}, we obtain
\begin{equation*}
 \sum_{n>m}\beta\,\kappa_l\sum_{k=1}^{K}w_k\,\omega_k(n-m)=\beta\,\kappa_l\sum_{k=1}^{K}w_k\sum_{s\in\mathbb N\setminus\{0\}}\omega_k(s)=\beta\,\kappa_l . \qedhere
\end{equation*}
\end{proof}

\runin{Mean activity}By Definition~\ref{def:selfexc}, the threshold $\|\phi\|=1$ decides whether the size of the updates stays finite on average. As the base rate $p_0\bar\kappa$ of our market does not change over time, we can solve \eqref{eq:mean} exactly and follow the activity on both sides of the threshold.

\begin{proposition}[Mean activity]\label{prop:longrun}
Let the assumptions of Proposition~\ref{prop:selfexc} hold with $\bar\kappa>0$. Then for every step $n$,
\begin{equation*}
 \mathbb E[I_n]=p_0\,\bar\kappa\Bigl(1+\sum_{s=1}^{n}R(s)\Bigr).
\end{equation*}
\begin{enumerate}[label=(\roman*),leftmargin=*]
\item If $\|\phi\|<1$, then $\sum_{s\in\mathbb N\setminus\{0\}}R(s)=\|\phi\|/(1-\|\phi\|)$, and as $n\to\infty$
\begin{equation*}
 \mathbb E[I_n]\uparrow\frac{p_0\,\bar\kappa}{1-\|\phi\|},\qquad\mathbb E[p_k(n)]\to\frac{p_0}{1-\|\phi\|}\quad\text{for every cluster }k,\qquad\mathbb E[\mathcal A_n]\to\frac{p_0}{1-\|\phi\|}.
\end{equation*}
In particular, Assumption~\ref{ass:interior}(a) implies $\|\phi\|\le1-p_0/\bar p$.
\item If $\|\phi\|\ge1$, then there exist a cluster $k\in\{1,\dots,K\}$ and a step $n\in\mathbb N$ with $\mathbb P\bigl(p_0+\beta U_k(n)>\bar p\bigr)>0$.
\end{enumerate}
\end{proposition}

Below the threshold, the desks thus multiply the activity $p_0$, which comes from information outside the desks, by the factor $1/(1-\|\phi\|)$. As $\|\phi\|$ approaches one, almost all activity is set off by earlier actions of the desks, which by (i) requires $p_0$ to be small. However, above this threshold, the probabilities of acting grow until they exceed $\bar p$, which leads to the failure of Assumption~\ref{ass:interior}(a).

\begin{proof}
We first show the formula for $\mathbb E[I_n]$ by induction on $n$. For $n=0$ we obtain $\mathbb E[I_0]=p_0\bar\kappa$ from \eqref{eq:mean}. If the formula holds below $n$, then substituting it into \eqref{eq:mean} with $u=n-m$, we obtain
\begin{equation*}
 \begin{aligned}
\mathbb E[I_n]&=p_0\,\bar\kappa+\sum_{m=0}^{n-1}\phi(n-m)\,p_0\,\bar\kappa\Bigl(1+\sum_{r=1}^{m}R(r)\Bigr)\\
&=p_0\,\bar\kappa\Bigl(1+\sum_{u=1}^{n}\phi(u)+\sum_{r=1}^{n-1}R(r)\sum_{u=1}^{n-r}\phi(u)\Bigr)\\
&=p_0\,\bar\kappa\Bigl(1+\sum_{s=1}^{n}\Bigl(\phi(s)+\sum_{r=1}^{s-1}\phi(s-r)\,R(r)\Bigr)\Bigr)=p_0\,\bar\kappa\Bigl(1+\sum_{s=1}^{n}R(s)\Bigr),
\end{aligned}
\end{equation*}
where the third line collects the terms with $r+u=s$ and the last equality is the recursion $R(s)=\phi(s)+\sum_{r=1}^{s-1}\phi(s-r)\,R(r)$, which we showed in the proof of Corollary~\ref{cor:response}. Next, for functions $\phi,\psi\colon\mathbb N\setminus\{0\}\to[0,\infty)$ we have $\|\phi*\psi\|=\|\phi\|\,\|\psi\|$, as both sides sum the same nonnegative terms. Hence $\|\phi^{*j}\|=\|\phi\|^j$, and
\begin{equation*}
 \sum_{s\in\mathbb N\setminus\{0\}}R(s)=\sum_{j\in\mathbb N\setminus\{0\}}\|\phi\|^j=\begin{cases}\|\phi\|/(1-\|\phi\|),&\|\phi\|<1,\\ \infty,&\|\phi\|\ge1 .\end{cases}
\end{equation*}
For (i), the partial sums of $R\ge0$ increase to $\|\phi\|/(1-\|\phi\|)$, such that $\mathbb E[I_n]$ increases to $I^*=p_0\bar\kappa/(1-\|\phi\|)$. By the proof of Proposition~\ref{prop:activity}, $\mathbb E[p_k(n)]=p_0+\beta\sum_{s=1}^{n}\omega_k(s)\,\mathbb E[I_{n-s}]$. Each term converges to $\beta\,\omega_k(s)\,I^*$ and is bounded by it, while $\sum_s\omega_k(s)=1$. By dominated convergence and Proposition~\ref{prop:strength}, we therefore obtain
\begin{equation*}
 \mathbb E[p_k(n)]\to p_0+\beta\,I^*=p_0\Bigl(1+\frac{\beta\bar\kappa}{1-\|\phi\|}\Bigr)=\frac{p_0}{1-\|\phi\|},
\end{equation*}
and $\mathbb E[\mathcal A_n]=\sum_kw_k\,\mathbb E[p_k(n)]$ has the same limit. By Assumption~\ref{ass:interior}(a) we have $p_k(n)\le\bar p$ in every step, such that the limit satisfies $p_0/(1-\|\phi\|)\le\bar p$, that is, $\|\phi\|\le1-p_0/\bar p$. For (ii), suppose that $p_0+\beta U_k(n)\le\bar p$ almost surely for every cluster $k$ and step $n$, such that Assumption~\ref{ass:interior}(a) holds in every step. By Lemma~\ref{lem:size} and Assumption~\ref{ass:interior}(a), we then obtain $\mathbb E[I_n]=\sum_l\kappa_lw_l\,\mathbb E[p_l(n)]\le\bar p\,\bar\kappa$ for every $n$. By the formula for $\mathbb E[I_n]$ and $\sum_sR(s)=\infty$, however, $\mathbb E[I_n]\to\infty$, which is a contradiction. 
\end{proof}

\runin{The kernel as a function of time}The kernel $\phi(s)$ measures the lag $s$ in steps. Note, however, that in the continuum order book (Proposition~\ref{prop:dk}) each step lasts only the time $\gamma$, while by Assumption~\ref{ass:regime} the moving average \eqref{eq:memory} of cluster $l$ extends over $T_l=\tau_l/\gamma$ steps. Therefore, as $\gamma\to0^+$, each value $\phi(s)$ vanishes, although each moving average still covers the fixed time $T_l\gamma=\tau_l$. We thus write the kernel as a function of the elapsed time $t=(s-1)\gamma$. Expanding the exponentials in $\Delta Z_i$ to first order in $\gamma$, an acting desk changes its volume near the mid price by $|\Delta Z_i|=\gamma|a_i|+O(\gamma^2)$ with $a_i=-(\sigma_i/\ell)\sum_{j=1}^{L}e^{-\sigma_iy_{i,j}/\ell}\,b_{i,j}$, while $c=\gamma\bar c$ by Assumption~\ref{ass:regime}. The sizes of the updates per unit activity thus satisfy
\begin{equation*}
 \kappa_l=\gamma^2\,\frac{\bar c}{w_l}\sum_{i:\,k(i)=l}m_i\,|a_i|+O(\gamma^3),
\end{equation*}
such that $\|\phi\|=\beta\bar\kappa$ with $\beta=\bar\beta/\gamma^2$ stays of order one as $\gamma\to0^+$. Let
\begin{equation*}
 \bar\phi(t)=\|\phi\|\sum_{l=1}^{K}\varpi_l\,\frac{e^{-t/\tau_l}}{\tau_l},\qquad\varpi_l=\frac{\kappa_l\,w_l}{\bar\kappa},\qquad t\ge0,
\end{equation*}
denote the kernel as a function of time, where the weights $\varpi_l$ sum to one.

\begin{proposition}[The kernel as a function of time]\label{prop:kerneltime}
Under the assumptions of Proposition~\ref{prop:selfexc} and Assumption~\ref{ass:regime} with $\gamma<\tau_1$, for every $s\in\mathbb N\setminus\{0\}$ and $t=(s-1)\gamma$,
\begin{equation*}
 \gamma\,\|\phi\|\sum_{l=1}^{K}\varpi_l\,\frac{e^{-t/(\tau_l-\gamma)}}{\tau_l}\le\phi(s)\le\gamma\,\bar\phi(t),\qquad\int_0^\infty\bar\phi(t)\,dt=\|\phi\| .
\end{equation*}
\end{proposition}

As a function of time, the kernel is thus a mixture of exponentials, one for each horizon $\tau_l$, where cluster $l$ enters with the weight $\varpi_l=\kappa_lw_l/\bar\kappa$. Its total weight $\|\phi\|$ does not depend on $\gamma$ and thus survives the limit $\gamma\to0^+$.

\begin{proof}
By Proposition~\ref{prop:strength} we have $\beta\kappa_lw_l=\|\phi\|\varpi_l$. Multiplying Lemma~\ref{lem:timescales}(ii) by $\beta\kappa_lw_l$ and summing over $l$, we obtain the two bounds from \eqref{eq:kernel}. Lastly, as $\int_0^\infty e^{-t/\tau_l}\,dt/\tau_l=1$ and $\sum_l\varpi_l=1$, we obtain $\int_0^\infty\bar\phi(t)\,dt=\|\phi\|$. 
\end{proof}

We next ask how this mixture fades when the horizons of the desks spread over many scales.

\subsection{A power-law kernel from many horizons}\label{sec:powerlaw}
By Proposition~\ref{prop:kerneltime}, the kernel of our market is a mixture of exponentials, one for each horizon $\tau_l$,
\begin{equation*}
 \bar\phi(t)=\|\phi\|\sum_{l=1}^{K}\varpi_l\,\frac{e^{-t/\tau_l}}{\tau_l},\qquad t\ge0 .
\end{equation*}
For a single horizon, $K=1$, the weight of the kernel beyond the time $x$ decays exponentially,
\begin{equation*}
 \int_x^\infty\bar\phi(t)\,dt=\|\phi\|\,e^{-x/\tau_1},\qquad x\ge0 .
\end{equation*}
The self-excitation of the size of the updates therefore extends only over times of order $\tau_1$. In this subsection we show that horizons on many scales instead make $\bar\phi$ decay as the power $t^{-1-\alpha}$ with the exponent $\alpha$ of the volumes \eqref{eq:weights}.

The horizons of the clusters, however, spread over many scales, as by Section~\ref{sec:info} and Assumption~\ref{ass:regime}
\begin{equation*}
 \tau_k=\gamma\,T_k=\tau_1\,\Omega^{k-1},\qquad k\in\{1,\dots,K\}.
\end{equation*}
Further, the clusters differ in their sizes of the updates per unit activity, and the bounds below depend on these sizes only through their extremes $\kappa_-=\min_l\kappa_l$ and $\kappa_+=\max_l\kappa_l$.

\begin{proposition}[Power-law kernel]\label{prop:powerlaw}
Let the assumptions of Proposition~\ref{prop:kerneltime} hold with $\kappa_l>0$ for every cluster $l$. Let
\begin{equation*}
 \mathcal I(t)=\int_{t/\tau_K}^{t/\tau_1}u^{\alpha}e^{-u}\,du,\qquad\mathcal I_\infty=\int_0^\infty u^{\alpha}e^{-u}\,du,\qquad C_\phi=\frac{\|\phi\|}{\ln\Omega\,\sum_{l=1}^{K}\tau_l^{-\alpha}},
\end{equation*}
where $C_\phi$ depends only on $\|\phi\|$, $\alpha$, $\tau_1$, $\Omega$ and $K$. Then the following hold.
\begin{enumerate}[label=(\roman*),leftmargin=*]
\item For every $t>0$,
\begin{equation*}
 \frac{\kappa_-}{\kappa_+}\,\frac{\mathcal I(t)}{\Omega^{1+\alpha}}\le\frac{t^{1+\alpha}\,\bar\phi(t)}{C_\phi}\le\frac{\kappa_+}{\kappa_-}\Bigl(\Omega^{1+\alpha}\,\mathcal I(t)+\ln\Omega\,\Bigl(\frac{t}{\tau_K}\Bigr)^{1+\alpha}e^{-t/\tau_K}\Bigr).
\end{equation*}
\item Let $t$, $\tau_1$ and $K$ vary such that $\tau_1/t\to0$ and $t/\tau_K\to0$, while $\Omega$, $\alpha$, $\kappa_-$ and $\kappa_+$ stay fixed. Then
\begin{equation*}
 \frac{\kappa_-}{\kappa_+}\,\frac{\mathcal I_\infty}{\Omega^{1+\alpha}}\le\liminf\frac{t^{1+\alpha}\,\bar\phi(t)}{C_\phi}\le\limsup\frac{t^{1+\alpha}\,\bar\phi(t)}{C_\phi}\le\frac{\kappa_+}{\kappa_-}\,\Omega^{1+\alpha}\,\mathcal I_\infty .
\end{equation*}
\end{enumerate}
\end{proposition}

Between the shortest and the longest horizon, the kernel thus decays as $t^{-1-\alpha}$. As $\tau_1^{-\alpha}\le\sum_l\tau_l^{-\alpha}\le\tau_1^{-\alpha}/(1-\Omega^{-\alpha})$, the constant $C_\phi$ stays bounded as $K\to\infty$. Clusters with longer horizons hence extend the power law to longer times.

Notice that on the logarithmic scale of the horizons the sum in $\bar\phi$ is a Riemann sum,
\begin{equation*}
 \sum_{l=1}^{K}\tau_l^{-1-\alpha}e^{-t/\tau_l}\approx\frac{1}{\ln\Omega}\int_{\tau_1}^{\tau_K}\tau^{-2-\alpha}e^{-t/\tau}\,d\tau=\frac{t^{-1-\alpha}}{\ln\Omega}\,\mathcal I(t).
\end{equation*}

\begin{proof}
By \eqref{eq:weights} and Assumption~\ref{ass:regime}, the factor $\gamma^{\alpha}$ cancels in $w_l=T_l^{-\alpha}/\sum_jT_j^{-\alpha}$, such that $w_l=\tau_l^{-\alpha}/\sum_j\tau_j^{-\alpha}$. As $\bar\kappa=\sum_l\kappa_lw_l$ with $\sum_lw_l=1$, we further have $\kappa_-\le\bar\kappa\le\kappa_+$. Now, let $f(\tau)=\tau^{-1-\alpha}e^{-t/\tau}$ for $\tau>0$. Inserting $\varpi_l=\kappa_lw_l/\bar\kappa$ into $\bar\phi$, we obtain
\begin{equation*}
 \frac{\kappa_-}{\kappa_+}\,C_\phi\ln\Omega\sum_{l=1}^{K}f(\tau_l)\le\bar\phi(t)=\|\phi\|\sum_{l=1}^{K}\frac{\kappa_l}{\bar\kappa}\,\frac{\tau_l^{-\alpha}}{\sum_j\tau_j^{-\alpha}}\,\frac{e^{-t/\tau_l}}{\tau_l}\le\frac{\kappa_+}{\kappa_-}\,C_\phi\ln\Omega\sum_{l=1}^{K}f(\tau_l).
\end{equation*}
Next, let $l\in\{1,\dots,K-1\}$ and $\tau\in[\tau_l,\tau_{l+1}]$. Since $\tau^{-1-\alpha}$ decreases and $e^{-t/\tau}$ increases in $\tau$, we obtain from $\tau_{l+1}=\Omega\tau_l$
\begin{equation*}
 \Omega^{-1-\alpha}f(\tau_l)=(\Omega\tau_l)^{-1-\alpha}e^{-t/\tau_l}\le f(\tau)\le(\tau_{l+1}/\Omega)^{-1-\alpha}e^{-t/\tau_{l+1}}=\Omega^{1+\alpha}f(\tau_{l+1}).
\end{equation*}
Integrating against $d\tau/\tau$ over $[\tau_l,\tau_{l+1}]$, summing over $l$ and substituting $u=t/\tau$, we obtain
\begin{equation*}
 \begin{aligned}
\frac{\ln\Omega}{\Omega^{1+\alpha}}\sum_{l=1}^{K-1}f(\tau_l)\le\int_{\tau_1}^{\tau_K}f(\tau)\,\frac{d\tau}{\tau}&=\int_{t/\tau_K}^{t/\tau_1}\Bigl(\frac{u}{t}\Bigr)^{1+\alpha}e^{-u}\,\frac{du}{u}\\
&=t^{-1-\alpha}\,\mathcal I(t)\le\Omega^{1+\alpha}\ln\Omega\sum_{l=2}^{K}f(\tau_l).
\end{aligned}
\end{equation*}
Adding $f(\tau_K)=t^{-1-\alpha}(t/\tau_K)^{1+\alpha}e^{-t/\tau_K}$ to the left sum and $f(\tau_1)\ge0$ to the right sum, we obtain
\begin{equation*}
 \frac{t^{-1-\alpha}\,\mathcal I(t)}{\Omega^{1+\alpha}\ln\Omega}\le\sum_{l=1}^{K}f(\tau_l)\le t^{-1-\alpha}\Bigl(\frac{\Omega^{1+\alpha}}{\ln\Omega}\,\mathcal I(t)+\Bigl(\frac{t}{\tau_K}\Bigr)^{1+\alpha}e^{-t/\tau_K}\Bigr).
\end{equation*}
Inserting these bounds into the first display, we obtain (i). For (ii), the limits of integration of $\mathcal I(t)$ tend to $0$ and $\infty$, such that $\mathcal I(t)\to\mathcal I_\infty$ by monotone convergence. Moreover, $(t/\tau_K)^{1+\alpha}e^{-t/\tau_K}\le(t/\tau_K)^{1+\alpha}\to0$. Lastly, noting that the bounds in (i) depend on $t$, $\tau_1$ and $K$ only through $\mathcal I(t)$ and $t/\tau_K$, we obtain (ii). 
\end{proof}

\runin{Many desks}We now show that the bounds of Proposition~\ref{prop:powerlaw}(ii) close in a market with many desks, which gives $\bar\phi(t)/\|\phi\|$ an exact tail $k_0\,t^{-1-\alpha}$. The two bounds on $t^{1+\alpha}\bar\phi(t)/C_\phi$ in (ii) differ by the factor
\begin{equation*}
 \frac{(\kappa_+/\kappa_-)\,\Omega^{1+\alpha}\,\mathcal I_\infty}{(\kappa_-/\kappa_+)\,\Omega^{-1-\alpha}\,\mathcal I_\infty}=\Bigl(\frac{\kappa_+}{\kappa_-}\Bigr)^{2}\,\Omega^{2(1+\alpha)}>1 .
\end{equation*}
Here, the factor $\Omega^{2(1+\alpha)}$ grows with the ratio $\Omega=\tau_{l+1}/\tau_l$ of neighboring horizons, while $(\kappa_+/\kappa_-)^2$ grows with the differences between the clusters in the size of the updates per unit activity. Both factors therefore approach one when many clusters with nearly equal horizons and nearly equal sizes of the updates per unit activity fill the range of horizons.

A liquid market comes close to this situation, as many desks with horizons on many scales quote there at the same time. In our model, by contrast, the clusters group these desks into the horizons $\tau_k=\tau_1\Omega^{k-1}$, which keep the fixed ratio $\Omega$ between neighbors. We therefore describe a liquid market by the limit of many desks.\footnote{Rough volatility is usually observed in liquid markets \cite{GJR18}, which makes this limit the relevant one for our question.} In this limit, the number of clusters grows, $K\to\infty$, and neighboring horizons approach each other, $\Omega\to1^+$. With $N$ desks per cluster and side, the market then holds $2NK\to\infty$ desks. For the second factor $(\kappa_+/\kappa_-)^2$, we note that the protocol treats two clusters alike except for their horizons and volumes, while $\kappa_l$ divides out the volume $w_l$. If the targets of all clusters follow the same distribution, clusters with nearly equal horizons thus have nearly equal $\kappa_l$. In the limit of many desks we therefore let $\kappa_l$ vary continuously with the horizon.

Now, let $\kappa\colon[\tau_1,\infty)\to(0,\infty)$ be continuous with $\inf\kappa>0$, $\sup\kappa<\infty$ and $\kappa(\tau)\to\kappa_\infty$ as $\tau\to\infty$, and let $\kappa_l=\kappa(\tau_l)$. Further, let
\begin{equation*}
 \mathcal K(t)=\frac{\bar\phi(t)}{\|\phi\|}=\sum_{l=1}^{K}\varpi_l\,\frac{e^{-t/\tau_l}}{\tau_l},\qquad t\ge0,
\end{equation*}
denote the normalized kernel, which satisfies $\int_0^\infty\mathcal K(t)\,dt=1$ by Proposition~\ref{prop:kerneltime}. By \eqref{eq:weights}, the weight $\varpi_l$ is proportional to $\kappa(\tau_l)\,\tau_l^{-\alpha}$, such that $\mathcal K(t)$ is proportional to the sum on the left below. Since $\ln\tau_{l+1}-\ln\tau_l=\ln\Omega$, we obtain in the limit of many desks
\begin{equation*}
 \ln\Omega\sum_{l=1}^{K}\kappa(\tau_l)\,\tau_l^{-1-\alpha}e^{-t/\tau_l}\to\int_{\tau_1}^{\infty}\kappa(\tau)\,\tau^{-2-\alpha}e^{-t/\tau}\,d\tau=t^{-1-\alpha}\int_0^{t/\tau_1}\kappa\Bigl(\frac{t}{u}\Bigr)u^{\alpha}e^{-u}\,du,
\end{equation*}
where we substituted $u=t/\tau$ in the last step. As $\kappa(t/u)\to\kappa_\infty$, the last integral converges to $\kappa_\infty\mathcal I_\infty$ as $t\to\infty$.

\begin{proposition}[Kernel of many desks]\label{prop:manydesks}
Under the assumptions of Proposition~\ref{prop:powerlaw}, let $\kappa_l=\kappa(\tau_l)$ for every cluster $l$, and let
\begin{equation*}
 \begin{aligned}
 \varpi(\tau)&=\frac{\alpha\,\tau_1^{\alpha}\,\kappa(\tau)\,\tau^{-1-\alpha}}{\bar\kappa_\infty},\qquad\bar\kappa_\infty=\int_{\tau_1}^{\infty}\alpha\,\tau_1^{\alpha}\,\kappa(\tau)\,\tau^{-1-\alpha}\,d\tau,\\
 \mathcal K_\infty(t)&=\int_{\tau_1}^{\infty}\varpi(\tau)\,\frac{e^{-t/\tau}}{\tau}\,d\tau .
 \end{aligned}
\end{equation*}
Then the following hold.
\begin{enumerate}[label=(\roman*),leftmargin=*]
\item As $\Omega\to1^+$ and $K\to\infty$ with $\tau_K\to\infty$, we have $\mathcal K(t)\to\mathcal K_\infty(t)$ for every $t>0$, and $\int_0^\infty\mathcal K_\infty(t)\,dt=1$.
\item As $t\to\infty$,
\begin{equation*}
 t^{1+\alpha}\,\mathcal K_\infty(t)\to k_0,\qquad\alpha\,x^{\alpha}\int_x^\infty\mathcal K_\infty(t)\,dt\to k_0\quad\text{as }x\to\infty,\qquad k_0=\frac{\alpha\,\tau_1^{\alpha}\,\kappa_\infty\,\mathcal I_\infty}{\bar\kappa_\infty}.
\end{equation*}
\end{enumerate}
\end{proposition}

The exponent of the tail is thus set by the volumes \eqref{eq:weights} alone, while the sizes of the updates per unit activity enter only through the constant $k_0$. If all clusters have the same size of the updates per unit activity, $\kappa=\kappa_\infty$, then $\bar\kappa_\infty=\kappa_\infty$, such that $k_0=\alpha\,\tau_1^{\alpha}\,\mathcal I_\infty$ depends only on $\alpha$ and the shortest horizon.

\begin{proof}
For (i), let $g$ be continuous on $[\tau_1,\infty)$ with $|g(\tau)|\le\sup\kappa\,\tau^{-\alpha}$. We first show that
\begin{equation*}
 \ln\Omega\sum_{l=1}^{K}g(\tau_l)\to\int_{\tau_1}^{\infty}g(\tau)\,\frac{d\tau}{\tau}\qquad\text{as }\Omega\to1^+,\ \tau_K\to\infty .
\end{equation*}
As $\ln\tau_{l+1}-\ln\tau_l=\ln\Omega$, the left side is a Riemann sum of the function $\ln\tau\mapsto g(\tau)$ with the step $\ln\Omega$. For a horizon $\hat\tau>\tau_1$, the terms with $\tau_l\le\hat\tau$ thus converge to $\int_{\tau_1}^{\hat\tau}g(\tau)\,d\tau/\tau$ by uniform continuity on $[\ln\tau_1,\ln\hat\tau]$. The remaining terms and the remaining integral satisfy
\begin{equation*}
 \begin{aligned}
\ln\Omega\sum_{l:\,\tau_l>\hat\tau}|g(\tau_l)|&\le\sup\kappa\,\ln\Omega\sum_{j\in\mathbb N}\bigl(\hat\tau\,\Omega^{j}\bigr)^{-\alpha}=\sup\kappa\,\frac{\ln\Omega}{1-\Omega^{-\alpha}}\,\hat\tau^{-\alpha},\\
\int_{\hat\tau}^\infty|g(\tau)|\,\frac{d\tau}{\tau}&\le\sup\kappa\,\frac{\hat\tau^{-\alpha}}{\alpha},
\end{aligned}
\end{equation*}
where $\ln\Omega/(1-\Omega^{-\alpha})\to1/\alpha$. Letting first $\Omega\to1^+$ and then $\hat\tau\to\infty$, we obtain the claim. Next, by \eqref{eq:weights} and Assumption~\ref{ass:regime} we have $\varpi_l=\kappa_l\tau_l^{-\alpha}/\sum_j\kappa_j\tau_j^{-\alpha}$, such that
\begin{equation*}
 \mathcal K(t)=\frac{\ln\Omega\sum_{l}\kappa(\tau_l)\,\tau_l^{-\alpha}\,e^{-t/\tau_l}/\tau_l}{\ln\Omega\sum_{j}\kappa(\tau_j)\,\tau_j^{-\alpha}}\to\frac{\int_{\tau_1}^{\infty}\kappa(\tau)\,\tau^{-2-\alpha}\,e^{-t/\tau}\,d\tau}{\int_{\tau_1}^{\infty}\kappa(\tau)\,\tau^{-1-\alpha}\,d\tau}=\mathcal K_\infty(t),
\end{equation*}
where we applied the claim to $g(\tau)=\kappa(\tau)\tau^{-1-\alpha}e^{-t/\tau}$ and to $g(\tau)=\kappa(\tau)\tau^{-\alpha}$. Moreover, by Fubini's theorem and $\int_0^\infty e^{-t/\tau}\,dt/\tau=1$, we obtain $\int_0^\infty\mathcal K_\infty(t)\,dt=\int_{\tau_1}^\infty\varpi(\tau)\,d\tau=1$. For (ii), substituting $u=t/\tau$, we obtain
\begin{equation*}
 t^{1+\alpha}\,\mathcal K_\infty(t)=\frac{\alpha\,\tau_1^{\alpha}}{\bar\kappa_\infty}\int_0^{t/\tau_1}\kappa\Bigl(\frac{t}{u}\Bigr)\,u^{\alpha}e^{-u}\,du .
\end{equation*}
For every $u>0$, $\kappa(t/u)\to\kappa_\infty$ as $t\to\infty$, while the integrand is bounded by $\sup\kappa\,u^{\alpha}e^{-u}$. By dominated convergence, $t^{1+\alpha}\mathcal K_\infty(t)\to k_0$, that is, $\mathcal K_\infty(t)=t^{-1-\alpha}\bigl(k_0+o(1)\bigr)$. Lastly, integrating from $x$ to $\infty$, we obtain
\begin{equation*}
 \alpha\,x^{\alpha}\int_x^\infty\mathcal K_\infty(t)\,dt=\alpha\,x^{\alpha}\int_x^\infty t^{-1-\alpha}\bigl(k_0+o(1)\bigr)\,dt=k_0+o(1)\qquad\text{as }x\to\infty . \qedhere
\end{equation*}
\end{proof}

\runin{Spacing of the horizons}Proposition~\ref{prop:manydesks} closes the bounds of Proposition~\ref{prop:powerlaw} only in the limit $\Omega\to1^+$. We now show that already a few clusters per decade of horizons make the kernel a power law to high accuracy. To isolate the role of the ratio $\Omega$, we let all clusters have the same size of the updates per unit activity and let the horizons $\tau_k=\tau_1\Omega^{k-1}$ run over all $k\in\mathbb Z$. A market with $K$ clusters drops the terms with $k\le0$ and $k>K$. As in the proof of Proposition~\ref{prop:powerlaw}(ii), these terms vanish as $\tau_1/t\to0$ and $t/\tau_K\to0$. Now, let
\begin{equation*}
 \mathcal J_m=\int_0^\infty u^{\alpha+2\pi i m/\ln\Omega}\,e^{-u}\,du,\qquad m\in\mathbb N,
\end{equation*}
which extend $\mathcal I_\infty$ to complex powers of $u$, such that $\mathcal J_0=\mathcal I_\infty$.

\begin{remark}[Log-periodic corrections]\label{rem:logperiodic}
For every $t>0$,
\begin{equation*}
 \frac{\ln\Omega}{\mathcal I_\infty}\,t^{1+\alpha}\sum_{k\in\mathbb Z}\tau_k^{-1-\alpha}e^{-t/\tau_k}=1+\frac{2}{\mathcal I_\infty}\sum_{m\in\mathbb N\setminus\{0\}}\mathrm{Re}\Bigl(\overline{\mathcal J_m}\,e^{2\pi i m\ln(t/\tau_1)/\ln\Omega}\Bigr).
\end{equation*}
The left side is thus periodic in $\ln t$ with the period $\ln\Omega$ and oscillates around one. Hence, the relative error of the power law $(\mathcal I_\infty/\ln\Omega)\,t^{-1-\alpha}$ satisfies
\begin{equation*}
 \Bigl|\frac{\ln\Omega}{\mathcal I_\infty}\,t^{1+\alpha}\sum_{k\in\mathbb Z}\tau_k^{-1-\alpha}e^{-t/\tau_k}-1\Bigr|\le\frac{2}{\mathcal I_\infty}\sum_{m\in\mathbb N\setminus\{0\}}|\mathcal J_m| .
\end{equation*}
By Stirling's formula, the terms $|\mathcal J_m|$ of this bound decay as $e^{-\pi^2m/\ln\Omega}$ up to a power of $m$. Moreover, $|\mathcal J_m|$ increases with $\Omega$. As each decade of horizons holds $\ln10/\ln\Omega$ clusters, more clusters per decade thus give a smaller relative error. With at least $\ln10/\ln2\approx3.3$ clusters per decade, that is, with $\Omega\le2$, we obtain a relative error of at most $9.1\times10^{-5}$ for every $\alpha\in(1/2,1)$. Lastly, \cite{Sor98} traces oscillations that are periodic in the logarithm of time to discrete scale invariance. In our market, the fixed ratio $\Omega$ of neighboring horizons produces this invariance.
\end{remark}

\begin{proof}
Let $F(z)=\exp\bigl((1+\alpha)z-e^{z}\bigr)$ for $z\in\mathbb R$. Writing $y=\ln(t/\tau_1)$ and $\ln\tau_k=\ln\tau_1+(k-1)\ln\Omega$, we obtain
\begin{equation*}
 t^{1+\alpha}\sum_{k\in\mathbb Z}\tau_k^{-1-\alpha}e^{-t/\tau_k}=\sum_{k\in\mathbb Z}F\bigl(y-(k-1)\ln\Omega\bigr)=\sum_{k\in\mathbb Z}F\bigl(y-k\ln\Omega\bigr).
\end{equation*}
Substituting $u=e^z$, we obtain for the Fourier transform of $F$
\begin{equation*}
 \hat F(\xi)=\int_{\mathbb R}F(z)\,e^{-i\xi z}\,dz=\int_0^\infty u^{\alpha-i\xi}e^{-u}\,du,
\end{equation*}
such that $\hat F(0)=\mathcal I_\infty$, $\hat F(2\pi m/\ln\Omega)=\overline{\mathcal J_m}$ and $\hat F(-\xi)=\overline{\hat F(\xi)}$. As $F$ decays exponentially, and $\hat F$ by Stirling's formula, we obtain from the Poisson summation formula
\begin{equation*}
 \sum_{k\in\mathbb Z}F\bigl(y-k\ln\Omega\bigr)=\frac{1}{\ln\Omega}\sum_{m\in\mathbb Z}\hat F\Bigl(\frac{2\pi m}{\ln\Omega}\Bigr)e^{2\pi i my/\ln\Omega}=\frac{1}{\ln\Omega}\Bigl(\mathcal I_\infty+2\sum_{m\ge1}\mathrm{Re}\bigl(\overline{\mathcal J_m}\,e^{2\pi imy/\ln\Omega}\bigr)\Bigr),
\end{equation*}
where we combined the terms $m$ and $-m$. Multiplying by $\ln\Omega/\mathcal I_\infty$, we obtain the identity. Moreover, using $|\mathrm{Re}(z)|\le|z|$, we obtain the bound. Lastly, with $y_m=2\pi m/\ln\Omega$, the product formula for $|\mathcal J_m|$ gives
\begin{equation*}
 \frac{|\mathcal J_m|^2}{\mathcal I_\infty^2}=\prod_{n\in\mathbb N}\Bigl(1+\frac{y_m^2}{(1+\alpha+n)^2}\Bigr)^{-1}.
\end{equation*}
The right side decreases in $y_m$ and thus increases in $\Omega$. 
\end{proof}

We next combine the power law $t^{-1-\alpha}$ with $\|\phi\|$ close to one to obtain the volatility of the mid price.

\section{Roughness}\label{sec:rough}
By Proposition~\ref{prop:manydesks}, a market with many desks has a normalized kernel with the tail $k_0\,t^{-1-\alpha}$. By Proposition~\ref{prop:longrun}, its norm $\|\phi\|$ can approach one only if $p_0$ vanishes, as $p_0\le\bar p\,(1-\|\phi\|)$. In this section we let $\|\phi\|$ approach one and show that the expected size of the updates is then rough with the Hurst parameter $H=\alpha-1/2$. As the size of the updates moves the mid price, the volatility of the mid price is then rough as well.

\subsection{Self-excitation close to the threshold}\label{sec:threshold}

By Proposition~\ref{prop:manydesks}, the normalized kernel of a market with many desks has the tail
\begin{equation*}
 \lim_{x\to\infty}\alpha\,x^{\alpha}\int_x^\infty\mathcal K_\infty(t)\,dt=k_0 .
\end{equation*}
For $\alpha\in(0,1)$, the Karamata Tauberian theorem \cite[Corollary~8.1.7]{BGT87} turns this tail into the behavior of the Laplace transform of $\mathcal K_\infty$ near zero,
\begin{equation*}
 \hat{\mathcal K}_\infty(z)=\int_0^\infty e^{-zt}\,\mathcal K_\infty(t)\,dt=1-\mathcal C_0\,z^{\alpha}+o(z^{\alpha})\quad\text{as }z\to0^+,\qquad\mathcal C_0=\frac{k_0}{\alpha}\int_0^\infty u^{-\alpha}e^{-u}\,du .
\end{equation*}
As in Corollary~\ref{cor:response}, the total response of the kernel $\|\phi\|\,\mathcal K_\infty$ thus has the Laplace transform
\begin{equation*}
 \hat R(z)=\frac{\|\phi\|\,\hat{\mathcal K}_\infty(z)}{1-\|\phi\|\,\hat{\mathcal K}_\infty(z)}=\frac{\|\phi\|\bigl(1+o(1)\bigr)}{(1-\|\phi\|)\bigl(1+(\mathcal Tz)^{\alpha}\bigr)+o(z^{\alpha})},\qquad\mathcal T=\Bigl(\frac{\|\phi\|\,\mathcal C_0}{1-\|\phi\|}\Bigr)^{1/\alpha}.
\end{equation*}
As $\hat R(z)$ weights the response up to times of order $1/z$, the response reaches its total $\|\phi\|/(1-\|\phi\|)$ only over times longer than $\mathcal T$. This time diverges as $\|\phi\|\to1^-$. By Proposition~\ref{prop:longrun}, the mean activity tends to $p_0/(1-\|\phi\|)$. We therefore measure time in units of $\mathcal T$ and the activity in units of its mean.

On times shorter than $\mathcal T$, the response has not yet reached its total. For $1/\mathcal T\ll z\ll1/\tau_1$, the Laplace transform $\hat R(z)$ becomes
\begin{equation*}
 \hat R(z)\approx\frac{\|\phi\|}{(1-\|\phi\|)\,(\mathcal Tz)^{\alpha}}=\frac{1}{\mathcal C_0\,z^{\alpha}}=\int_0^\infty e^{-zt}\,\frac{t^{\alpha-1}}{\mathcal C_0\int_0^\infty u^{\alpha-1}e^{-u}\,du}\,dt,
\end{equation*}
such that the response decays as $t^{\alpha-1}$ on these times. Now, let $\tilde I_n=I_n-\mathbb E[I_n\mid\mathcal F_n]$ denote the deviation of the size of the updates from its expectation, which has the mean zero given $\mathcal F_n$. Solving the self-excitation (Proposition~\ref{prop:selfexc}) as in the proof of Proposition~\ref{prop:longrun}, we obtain
\begin{equation*}
 \mathbb E[I_n\mid\mathcal F_n]=p_0\,\bar\kappa\Bigl(1+\sum_{s=1}^{n}R(s)\Bigr)+\sum_{m=0}^{n-1}R(n-m)\,\tilde I_m .
\end{equation*}
On times shorter than $\mathcal T$, the last sum thus weights each deviation $\tilde I_m$ with $(n-m)^{\alpha-1}$ up to a constant factor. We next state the limit of this representation for self-excitation in continuous time.

In a market that runs in continuous time, the desks act at random times. The number $N^{\mathcal T}_t$ of their actions up to the time $t\in[0,\mathcal T]$ then starts at zero and increases by one at each action. Let $\mathcal F^{\mathcal T}_t$ be the $\sigma$-algebra generated by $(N^{\mathcal T}_s)_{s\le t}$. The conditional expectation $\mathbb E[Y_n\mid\mathcal F_n]$ (Definition~\ref{def:selfexc}) then becomes the intensity $\lambda^{\mathcal T}\ge0$, the adapted process with
\begin{equation*}
 \mathbb E\bigl[(N^{\mathcal T}_b-N^{\mathcal T}_a)\,\mathbf 1_A\bigr]=\mathbb E\Bigl[\int_a^b\lambda^{\mathcal T}_s\,\mathbf 1_A\,ds\Bigr],\qquad0\le a<b\le\mathcal T,\quad A\in\mathcal F^{\mathcal T}_a .
\end{equation*}

\begin{definition}[Hawkes process]\label{def:hawkes}
The number of actions $N^{\mathcal T}$ is a \emph{Hawkes process} with the base rate $\mu_{\mathcal T}>0$ and the kernel $\mathcal K_{\mathcal T}\colon[0,\infty)\to[0,\infty)$ with $\|\mathcal K_{\mathcal T}\|=\int_0^\infty\mathcal K_{\mathcal T}(s)\,ds<1$ if its intensity is
\begin{equation*}
 \lambda^{\mathcal T}_t=\mu_{\mathcal T}+\int_0^t\mathcal K_{\mathcal T}(t-s)\,dN^{\mathcal T}_s,\qquad t\in[0,\mathcal T].
\end{equation*}
\end{definition}

Let $R_{\mathcal T}=\sum_{j\in\mathbb N\setminus\{0\}}\mathcal K_{\mathcal T}^{*j}$ denote the total response of the Hawkes process, and let $M^{\mathcal T}_t=N^{\mathcal T}_t-\int_0^t\lambda^{\mathcal T}_s\,ds$ denote the deviation of the number of actions from its integrated intensity, which is a martingale. Solving the self-excitation as for the expected size of the updates, we obtain
\begin{equation*}
 \lambda^{\mathcal T}_t=\mu_{\mathcal T}\Bigl(1+\int_0^tR_{\mathcal T}(s)\,ds\Bigr)+\int_0^tR_{\mathcal T}(t-s)\,dM^{\mathcal T}_s .
\end{equation*}

The time scale of the response satisfies $\mathcal T^{\alpha}(1-\|\phi\|)=\|\phi\|\,\mathcal C_0$. As $\mathcal T\to\infty$, the norm of the kernel thus approaches one at the rate $\mathcal T^{-\alpha}$. We therefore consider, for every $\mathcal T>0$, a Hawkes process $N^{\mathcal T}$ with the kernel $\mathcal K_{\mathcal T}=\|\mathcal K_{\mathcal T}\|\,\mathcal K$ and $\|\mathcal K_{\mathcal T}\|\to1^-$. As $\mathcal K_\infty$ does by Proposition~\ref{prop:manydesks}, the kernel $\mathcal K$ with $\|\mathcal K\|=1$ has a tail
\begin{equation*}
 \lim_{x\to\infty}\alpha\,x^{\alpha}\int_x^\infty\mathcal K(s)\,ds=k_0 .
\end{equation*}
In units of $\mathcal T$ and of the mean intensity $\mu_{\mathcal T}/(1-\|\mathcal K_{\mathcal T}\|)$, the number of actions becomes
\begin{equation*}
 \mathcal X^{\mathcal T}_t=\frac{1-\|\mathcal K_{\mathcal T}\|}{\mathcal T\,\mu_{\mathcal T}}\,N^{\mathcal T}_{\mathcal Tt},\qquad t\in[0,1].
\end{equation*}
On the same time scale, the total response $R_{\mathcal T}$ normalized to the total weight one is $\rho^{\mathcal T}(x)=\mathcal T R_{\mathcal T}(\mathcal Tx)/\|R_{\mathcal T}\|$. By the Karamata Tauberian theorem, this tail gives the Laplace transform of $\mathcal K$ near zero with the same constant $\mathcal C_0$ as for $\mathcal K_\infty$,
\begin{equation*}
 \hat{\mathcal K}(z)=\int_0^\infty e^{-zt}\,\mathcal K(t)\,dt=1-\mathcal C_0\,z^{\alpha}+o(z^{\alpha})\qquad\text{as }z\to0^+ .
\end{equation*}
Inserting this expansion, we find
\begin{equation*}
 \int_0^\infty e^{-zx}\rho^{\mathcal T}(x)\,dx=\frac{(1-\|\mathcal K_{\mathcal T}\|)\,\hat{\mathcal K}(z/\mathcal T)}{1-\|\mathcal K_{\mathcal T}\|\,\hat{\mathcal K}(z/\mathcal T)}\to\frac{\mathcal C_1}{\mathcal C_1+z^{\alpha}}\qquad\text{if }\ \frac{\mathcal T^{\alpha}\bigl(1-\|\mathcal K_{\mathcal T}\|\bigr)}{\mathcal C_0}\to\mathcal C_1>0 .
\end{equation*}
We let $\mathcal G$ denote the function on $(0,\infty)$ with this limit as its Laplace transform.

\begin{theorem}[{Hawkes processes close to the threshold \cite[Theorems~3.1 and~3.2]{JR16}}]\label{thm:jr16}
For every $\mathcal T>0$, let $N^{\mathcal T}$ be a Hawkes process with the base rate $\mu_{\mathcal T}$ and the kernel $\|\mathcal K_{\mathcal T}\|\,\mathcal K$, where $\|\mathcal K\|=1$. Assume that
\begin{enumerate}[label=(J\arabic*),leftmargin=*]
\item $\displaystyle\lim_{x\to\infty}\alpha\,x^{\alpha}\int_x^\infty\mathcal K(s)\,ds=k_0$ for some $\alpha\in(1/2,1)$ and $k_0>0$;
\item $\mathcal T^{\alpha}\bigl(1-\|\mathcal K_{\mathcal T}\|\bigr)\to\mathcal C_1\mathcal C_0$ and $\mathcal T^{1-\alpha}\mu_{\mathcal T}\to\mathcal C_2/\mathcal C_0$ as $\mathcal T\to\infty$, for some $\mathcal C_1,\mathcal C_2>0$.
\end{enumerate}
Then every sequence $\mathcal T_k\to\infty$ has a subsequence along which $\mathcal X^{\mathcal T_k}$ converges in law. Every such limit $\mathcal X$ is differentiable, and there is a Brownian motion $B$ such that
\begin{equation*}
 \mathcal X'_t=\int_0^t\mathcal G(s)\,ds+\frac{1}{\sqrt{\mathcal C_1\mathcal C_2}}\int_0^t\mathcal G(t-s)\sqrt{\mathcal X'_s}\,dB_s,\qquad t\in[0,1].
\end{equation*}
Moreover, for every $h<\alpha-1/2$,
\begin{equation*}
 \sup_{0\le s<t\le1}\frac{|\mathcal X'_t-\mathcal X'_s|}{(t-s)^{h}}<\infty\qquad\text{almost surely}.
\end{equation*}
\end{theorem}

Our normalization of $\mathcal X^{\mathcal T}$ differs from that in \cite{JR16} by the factor $\mathcal C_0\,\mathcal T^{1-\alpha}\mu_{\mathcal T}/\mathcal C_2$, which tends to one by (J2) and leaves the limit unchanged. The paths of fractional Brownian motion with the Hurst parameter $H$ satisfy the bound in Theorem~\ref{thm:jr16} for every $h<H$ \cite{GJR18}. In the limit, the intensity in units of its mean is thus as rough as fractional Brownian motion with the Hurst parameter $H=\alpha-1/2$.

The proof of Theorem~\ref{thm:jr16} passes to the limit in the representation of the intensity $\lambda^{\mathcal T}$ in three steps. Let
\begin{equation*}
 \mathcal Z^{\mathcal T}_t=\Bigl(\frac{1-\|\mathcal K_{\mathcal T}\|}{\mathcal T\,\mu_{\mathcal T}}\Bigr)^{1/2}M^{\mathcal T}_{\mathcal Tt},\qquad t\in[0,1],
\end{equation*}
denote the rescaled deviation of the number of actions from its integrated intensity. Since $N^{\mathcal T}$ jumps by one, the deviation $M^{\mathcal T}$ satisfies
\begin{equation*}
 \langle M^{\mathcal T}\rangle_t=\int_0^t\lambda^{\mathcal T}_s\,ds,\qquad[M^{\mathcal T},M^{\mathcal T}]_t=N^{\mathcal T}_t ,
\end{equation*}
where $\langle M^{\mathcal T}\rangle$ is the process that makes $(M^{\mathcal T})^2-\langle M^{\mathcal T}\rangle$ a martingale, and $[M^{\mathcal T},M^{\mathcal T}]$ is the sum of the squared jumps of $M^{\mathcal T}$. In the units of $\mathcal X^{\mathcal T}$, these two identities read
\begin{equation*}
 \langle\mathcal Z^{\mathcal T}\rangle_t=\mathcal X^{\mathcal T}_t-\Bigl(\frac{1-\|\mathcal K_{\mathcal T}\|}{\mathcal T\,\mu_{\mathcal T}}\Bigr)^{1/2}\mathcal Z^{\mathcal T}_t,\qquad[\mathcal Z^{\mathcal T},\mathcal Z^{\mathcal T}]_t=\mathcal X^{\mathcal T}_t .
\end{equation*}
In the first step, we obtain $\mathbb E[\sup_{t\le1}|\mathcal Z^{\mathcal T}_t|^2]\le4\,\mathbb E[\langle\mathcal Z^{\mathcal T}\rangle_1]$ from Doob's inequality. The last term in $\langle\mathcal Z^{\mathcal T}\rangle$ thus vanishes as $\mathcal T\to\infty$. Every limit $\mathcal Z$ is then a continuous martingale with the quadratic variation $\mathcal X$. In the second step, we insert the representation of $\lambda^{\mathcal T}$ into $\mathcal X^{\mathcal T}$ and let $\rho^{\mathcal T}\to\mathcal G$, which yields the equation in Theorem~\ref{thm:jr16}. Its Brownian motion $B$ comes from $\mathcal Z$, as a continuous martingale with the quadratic variation $\mathcal X$ is a Brownian motion run at the time $\mathcal X$,
\begin{equation*}
 \mathcal Z_t=B_{\mathcal X_t},\qquad t\in[0,1].
\end{equation*}
In the third step, the bound on the increments of $\mathcal X'$ follows from the singularity of $\mathcal G$ at zero. Of these steps, only the two identities for $M^{\mathcal T}$ depend on the desks acting one at a time. In Section~\ref{sec:correspondence} we derive their analogues for the size of the updates of our market.

\subsection{The size of the updates as a number of actions}\label{sec:correspondence}

The proof of Theorem~\ref{thm:jr16} depends on single actions only through two identities for the deviation $M^{\mathcal T}$ (Section~\ref{sec:threshold}). The first, $\langle M^{\mathcal T}\rangle_t=\int_0^t\lambda^{\mathcal T}_s\,ds$, states that the conditional variance of the number of actions equals its intensity. For our market, we therefore compute the conditional variance of the size of the updates. As for $\kappa_l$ in Proposition~\ref{prop:selfexc}, we freeze the contribution $c\,m_i|\Delta Z_i|$ of each acting desk $i$ to the size of the updates \eqref{eq:push} at its average. Now, let
\begin{equation*}
 \nu_l=\frac{1}{\kappa_l\,w_l}\sum_{i:\,k(i)=l}\bigl(c\,m_i|\Delta Z_i|\bigr)^2
\end{equation*}
denote the contribution of one desk of cluster $l$, averaged with the weights $c\,m_i|\Delta Z_i|/(\kappa_lw_l)$. Since the desks act independently given $\mathcal F_n$ (Section~\ref{sec:batch}), we obtain as in Lemma~\ref{lem:size}
\begin{equation*}
 \mathbb E[I_n\mid\mathcal F_n]=\sum_{l=1}^{K}\kappa_l\,w_l\,p_l(n),\qquad\mathrm{Var}[I_n\mid\mathcal F_n]=\sum_{l=1}^{K}\nu_l\bigl(1-p_l(n)\bigr)\,\kappa_l\,w_l\,p_l(n).
\end{equation*}
If the clusters share the probability of acting, $p_l(n)=p(n)$, the conditional variance becomes
\begin{equation*}
 \mathrm{Var}[I_n\mid\mathcal F_n]=\bar\nu\,\bigl(1-p(n)\bigr)\,\mathbb E[I_n\mid\mathcal F_n],\qquad\bar\nu=\frac{\sum_l\nu_l\,\kappa_l\,w_l}{\sum_l\kappa_l\,w_l}.
\end{equation*}
In units of $\bar\nu$, the size of the updates thus has the conditional variance of a number of actions, up to the factor $1-p(n)$. If the $2N$ desks of cluster $l$ contribute equally, then $\nu_l=\kappa_lw_l/(2N)$, such that
\begin{equation*}
 \bar\nu=\frac{1}{2N}\,\frac{\sum_l(\kappa_l\,w_l)^2}{\sum_l\kappa_l\,w_l}
\end{equation*}
decreases as $1/N$.

In units of $\bar\nu$, the size of the updates accumulated up to the time $t=n\gamma$ thus takes the role of the number of actions. By Proposition~\ref{prop:selfexc}, the size of the updates has the base rate $p_0\bar\kappa$ per step. By Proposition~\ref{prop:kerneltime}, the kernel per unit of time is $\bar\phi=\|\phi\|\,\mathcal K$, which tends to $\|\phi\|\,\mathcal K_\infty$ for many desks (Proposition~\ref{prop:manydesks}). The quantities in Theorem~\ref{thm:jr16} thus correspond to ours as
\begin{equation*}
 N^{\mathcal T}_t\;\longleftrightarrow\;\frac{1}{\bar\nu}\sum_{n\le t/\gamma}I_n,\qquad\mu_{\mathcal T}\;\longleftrightarrow\;\frac{p_0\,\bar\kappa}{\bar\nu\,\gamma},\qquad\|\mathcal K_{\mathcal T}\|\,\mathcal K\;\longleftrightarrow\;\|\phi\|\,\mathcal K_\infty .
\end{equation*}
Condition (J1) holds by Proposition~\ref{prop:manydesks} for $\alpha\in(1/2,1)$. With the time scale $\mathcal T$ (Section~\ref{sec:threshold}), we have $\mathcal T^{\alpha}(1-\|\phi\|)=\|\phi\|\,\mathcal C_0\to\mathcal C_0$, such that the first condition in (J2) holds with $\mathcal C_1=1$. Inserting $\mathcal T^{-\alpha}=(1-\|\phi\|)/(\|\phi\|\,\mathcal C_0)$, we write the second condition in (J2) as
\begin{equation*}
 \frac{\mathcal T\,(1-\|\phi\|)\,p_0}{\|\phi\|\,\gamma\,\bar\nu/\bar\kappa}\to\mathcal C_2 .
\end{equation*}
By Proposition~\ref{prop:longrun}, we have $p_0\le\bar p\,(1-\|\phi\|)$, such that
\begin{equation*}
 \gamma\,\frac{\bar\nu}{\bar\kappa}\le\frac{\mathcal T\,(1-\|\phi\|)^2\,\bar p}{\|\phi\|\,\mathcal C_2}\bigl(1+o(1)\bigr)=\frac{\|\phi\|\,\mathcal C_0^2\,\bar p}{\mathcal C_2}\,\mathcal T^{1-2\alpha}\bigl(1+o(1)\bigr)\to0\qquad\text{for }\alpha>1/2 .
\end{equation*}
Since $\bar\nu/\bar\kappa$ is proportional to $1/N$, the ratio $\gamma/N$ of the step length to the number of desks per cluster and side must vanish at least as fast as $\mathcal T^{1-2\alpha}$. A market observed over longer times must therefore hold more desks or act in shorter steps.

\subsection{The limit of the size of the updates}\label{sec:limitupdates}

The correspondence of Section~\ref{sec:correspondence} expresses the conditions (J1) and (J2) of Theorem~\ref{thm:jr16} in the quantities of our market. Yet the proof of Theorem~\ref{thm:jr16} also uses the two identities for the deviation $M^{\mathcal T}$ (Section~\ref{sec:threshold}), which depend on the desks acting one at a time. In our market, every cluster instead enters the conditional variance of the size of the updates with its own probability of acting $p_l(n)$ and the factor $1-p_l(n)$ (Section~\ref{sec:correspondence}). Moreover, the size of the updates in units of $\bar\nu$ jumps by $I_n/\bar\nu$ rather than by one. In the limit of many desks with vanishing mean activity $p_0/(1-\|\phi\|)$, both identities hold up to vanishing errors. We also let the horizons extend beyond $\mathcal T$ and the steps shrink relative to $\mathcal T$, such that the kernel keeps its power law up to $\mathcal T$ and sums over steps become integrals. Inserting the correspondence into $\mathcal X^{\mathcal T}$ (Section~\ref{sec:threshold}), we obtain
\begin{equation*}
 \mathcal X^{\mathcal T}_t=\frac{1-\|\phi\|}{\mathcal T}\,\frac{\bar\nu\,\gamma}{p_0\,\bar\kappa}\,\frac{1}{\bar\nu}\sum_{n\le\mathcal Tt/\gamma}I_n=\frac{\gamma}{\mathcal T}\sum_{n\le\mathcal Tt/\gamma}\frac{I_n}{\bar I},\qquad t\in[0,1],
\end{equation*}
where $\bar I=p_0\bar\kappa/(1-\|\phi\|)$ is the limit of $\mathbb E[I_n]$ (Proposition~\ref{prop:longrun}). With both identities, the rescaled size of the updates $\mathcal X^{\mathcal T}$ then has the same limit as the rescaled number of actions in Theorem~\ref{thm:jr16}.

\begin{theorem}[Size of the updates close to the threshold]\label{thm:rough}
For every $\mathcal T>0$, consider a market under Assumptions~\ref{ass:regime} and~\ref{ass:interior} with $\gamma<\tau_1$, in which the $2N$ desks of each cluster make equal frozen contributions (Section~\ref{sec:correspondence}) and the sizes of the updates per unit activity are proportional to $\kappa(\tau_l)$, with $\kappa$ as in Proposition~\ref{prop:manydesks} and independent of $\mathcal T$. Let $\mathcal T^{\alpha}(1-\|\phi\|)=\|\phi\|\,\mathcal C_0$ with $\alpha\in(1/2,1)$, and assume that, as $\mathcal T\to\infty$,
\begin{enumerate}[label=(R\arabic*),leftmargin=*]
\item $\Omega\to1^+$, $\tau_K/\mathcal T\to\infty$ and $\gamma/\mathcal T\to0$, while $\tau_1$ stays fixed;
\item $p_0/(1-\|\phi\|)\to0$ and $\mathcal T\,(1-\|\phi\|)\,p_0/(\|\phi\|\,\gamma\,\bar\nu/\bar\kappa)\to\mathcal C_2$ for some $\mathcal C_2>0$.
\end{enumerate}
Then every sequence $\mathcal T_k\to\infty$ has a subsequence along which $\mathcal X^{\mathcal T_k}$ converges in law. Every such limit $\mathcal X$ is differentiable, and there is a Brownian motion $B$ such that
\begin{equation*}
 \mathcal X'_t=\int_0^t\mathcal G(s)\,ds+\frac{1}{\sqrt{\mathcal C_2}}\int_0^t\mathcal G(t-s)\sqrt{\mathcal X'_s}\,dB_s,\qquad t\in[0,1],
\end{equation*}
where $\mathcal G$ has the Laplace transform $1/(1+z^{\alpha})$. Moreover, for every $h<\alpha-1/2$,
\begin{equation*}
 \sup_{0\le s<t\le1}\frac{|\mathcal X'_t-\mathcal X'_s|}{(t-s)^{h}}<\infty\qquad\text{almost surely}.
\end{equation*}
\end{theorem}

In the limit, the expected size of the updates in units of $\bar I$ is thus as rough as fractional Brownian motion with the Hurst parameter $H=\alpha-1/2$.

The first identity sums the conditional variance over all steps up to the time $\mathcal Tt$. By Section~\ref{sec:correspondence}, each cluster $l$ enters this sum through $\sum_rp_l(r)\bigl(1-p_l(r)\bigr)$. At a given step, clusters with different horizons have different probabilities of acting, as their moving averages $U_l$ weigh the past size of the updates differently \eqref{eq:memory}. Summed over all later steps, however, these weights equal one for every cluster (Section~\ref{sec:response}). We therefore compare the probabilities of acting summed over steps.

\begin{lemma}[Equal total response of the clusters]\label{lem:summedprob}
Under Assumption~\ref{ass:interior}(a), for every cluster $l$ and every step $n\in\mathbb N$,
\begin{equation*}
 \sum_{r=0}^{n}p_l(r)=(n+1)\,p_0+\beta\sum_{m=0}^{n-1}I_m-\beta\,(T_l-1)\,U_l(n).
\end{equation*}
\end{lemma}

Every cluster thus responds to each past update with the same total, and the clusters differ only in the part of this response that falls after the step $n$.

\begin{proof}
By Assumption~\ref{ass:interior}(a), $p_l(r)=p_0+\beta U_l(r)$ with $U_l(r)=\sum_{m=0}^{r-1}\omega_l(r-m)\,I_m$ (proof of Proposition~\ref{prop:activity}). Exchanging the sums and summing the geometric weights $\omega_l(s)=(1-1/T_l)^{s-1}/T_l$, we obtain
\begin{equation*}
 \sum_{r=0}^{n}U_l(r)=\sum_{m=0}^{n-1}I_m\sum_{s=1}^{n-m}\omega_l(s)=\sum_{m=0}^{n-1}I_m\Bigl(1-\Bigl(1-\frac{1}{T_l}\Bigr)^{n-m}\Bigr).
\end{equation*}
Since $(1-1/T_l)^{s}=(T_l-1)\,\omega_l(s)$ for every $s\in\mathbb N\setminus\{0\}$, the last term satisfies
\begin{equation*}
 \sum_{m=0}^{n-1}I_m\Bigl(1-\frac{1}{T_l}\Bigr)^{n-m}=(T_l-1)\sum_{m=0}^{n-1}\omega_l(n-m)\,I_m=(T_l-1)\,U_l(n).
\end{equation*}
Multiplying by $\beta$ and adding $(n+1)\,p_0$, we obtain the claim.
\end{proof}

Inserting Lemma~\ref{lem:summedprob} into the summed conditional variance, we obtain $\bar\nu$ times the summed conditional mean up to two errors.

\begin{lemma}[Conditional variance summed over steps]\label{lem:summedvar}
Under Assumption~\ref{ass:interior}(a) and with frozen contributions of the desks (Section~\ref{sec:correspondence}), for every step $n\in\mathbb N$,
\begin{equation*}
 \begin{aligned}\sum_{r=0}^{n}\mathrm{Var}\bigl[I_r\mid\mathcal F_r\bigr]&=\bar\nu\sum_{r=0}^{n}\mathbb E\bigl[I_r\mid\mathcal F_r\bigr]-\beta\sum_{l=1}^{K}(\nu_l-\bar\nu)\,\kappa_l\,w_l\,(T_l-1)\,U_l(n)\\&\quad-\sum_{l=1}^{K}\nu_l\,\kappa_l\,w_l\sum_{r=0}^{n}p_l(r)^2 .\end{aligned}
\end{equation*}
\end{lemma}

By the proof of Lemma~\ref{lem:summedprob} and Assumption~\ref{ass:regime}, the part of the response of cluster $l$ after the step $n$ is
\begin{equation*}
 \beta\,(T_l-1)\,U_l(n)=\beta\sum_{m=0}^{n-1}\Bigl(1-\frac{\gamma}{\tau_l}\Bigr)^{n-m}I_m\le\beta\sum_{m=0}^{n-1}e^{-(n-m)\gamma/\tau_l}\,I_m .
\end{equation*}
As $p_l(r)^2\le p_l(r)\max_{l',r'}p_{l'}(r')$, the second error satisfies
\begin{equation*}
 \sum_{l=1}^{K}\nu_l\,\kappa_l\,w_l\sum_{r=0}^{n}p_l(r)^2\le\max_{l}\frac{\nu_l}{\bar\nu}\,\max_{l,\,r\le n}p_l(r)\;\bar\nu\sum_{r=0}^{n}\mathbb E\bigl[I_r\mid\mathcal F_r\bigr].
\end{equation*}

\begin{proof}
By Section~\ref{sec:correspondence}, the conditional variance and the conditional mean of the size of the updates are $\sum_l\nu_l\kappa_lw_lp_l(r)(1-p_l(r))$ and $\sum_l\kappa_lw_lp_l(r)$. Summing over $r$ and subtracting $\bar\nu$ times the second sum from the first, we obtain
\begin{equation*}
 \sum_{r=0}^{n}\mathrm{Var}\bigl[I_r\mid\mathcal F_r\bigr]-\bar\nu\sum_{r=0}^{n}\mathbb E\bigl[I_r\mid\mathcal F_r\bigr]=\sum_{l=1}^{K}(\nu_l-\bar\nu)\,\kappa_l\,w_l\sum_{r=0}^{n}p_l(r)-\sum_{l=1}^{K}\nu_l\,\kappa_l\,w_l\sum_{r=0}^{n}p_l(r)^2 .
\end{equation*}
By the definition of $\bar\nu$, the weights $(\nu_l-\bar\nu)\kappa_lw_l$ sum to zero over $l$. When we insert Lemma~\ref{lem:summedprob} into the first sum on the right, the part $(n+1)\,p_0+\beta\sum_{m<n}I_m$ therefore cancels, and only $-\beta(T_l-1)U_l(n)$ remains.
\end{proof}

Taking expectations in the second error of Lemma~\ref{lem:summedvar} and inserting $U_l$ from the proof of Proposition~\ref{prop:activity}, we obtain
\begin{equation*}
 \mathbb E\Bigl[\sum_{l=1}^{K}\nu_l\,\kappa_l\,w_l\sum_{r=0}^{n}p_l(r)^2\Bigr]=\sum_{l=1}^{K}\nu_l\,\kappa_l\,w_l\sum_{r=0}^{n}\mathbb E\Bigl[\Bigl(p_0+\beta\sum_{m=0}^{r-1}\omega_l(r-m)\,I_m\Bigr)^2\Bigr].
\end{equation*}
Since the deviations $\tilde I_m=I_m-\mathbb E[I_m\mid\mathcal F_m]$ are uncorrelated, we obtain from the representation of the expected size of the updates (Section~\ref{sec:threshold})
\begin{equation*}
 \begin{aligned}\mathbb E\bigl[I_n\mid\mathcal F_n\bigr]&=p_0\,\bar\kappa\Bigl(1+\sum_{s=1}^{n}R(s)\Bigr)+\sum_{m=0}^{n-1}R(n-m)\,\tilde I_m,\\\mathrm{Var}\bigl(\mathbb E[I_n\mid\mathcal F_n]\bigr)&=\sum_{m=0}^{n-1}R(n-m)^2\,\mathbb E\bigl[\tilde I_m^2\bigr]\le\max_{m<n}\mathbb E\bigl[\tilde I_m^2\bigr]\sum_{s\in\mathbb N\setminus\{0\}}R(s)^2 .\end{aligned}
\end{equation*}

\begin{lemma}[Square-summable response]\label{lem:responsesq}
Under the assumptions of Theorem~\ref{thm:rough}, there is a constant $C_R$, depending only on $\alpha$, $\tau_1$, $\mathcal C_0$ and $\inf\kappa/\sup\kappa$, such that for every sufficiently large $\mathcal T$
\begin{equation*}
 \sum_{s\in\mathbb N\setminus\{0\}}R(s)^2\le C_R\,\frac{\gamma}{(1-\|\phi\|)^2\,\mathcal T}.
\end{equation*}
\end{lemma}

By Proposition~\ref{prop:longrun}, $\sum_sR(s)=\|\phi\|/(1-\|\phi\|)$, such that the lemma reads
\begin{equation*}
 \sum_{s\in\mathbb N\setminus\{0\}}R(s)^2\le\frac{C_R}{\|\phi\|^2}\,\frac{\gamma}{\mathcal T}\,\Bigl(\sum_{s\in\mathbb N\setminus\{0\}}R(s)\Bigr)^2 .
\end{equation*}

\begin{proof}
By Proposition~\ref{prop:longrun}, $\sum_sR(s)=\|\phi\|/(1-\|\phi\|)<\infty$. Parseval's identity thus applies to $\hat R(z)=\sum_sR(s)z^s$, and as $R$ is real,
\begin{equation*}
 \sum_{s\in\mathbb N\setminus\{0\}}R(s)^2=\frac{1}{2\pi}\int_{-\pi}^{\pi}\bigl|\hat R(e^{i\theta})\bigr|^2\,d\theta=\frac1\pi\int_0^\pi\bigl|\hat R(e^{i\theta})\bigr|^2\,d\theta .
\end{equation*}
By Assumption~\ref{ass:regime}, $\omega_l(s)=(1-a_l)\,a_l^{s-1}$ with $a_l=1-\gamma/\tau_l$. Using $\beta\kappa_lw_l=\|\phi\|\varpi_l$ (Proposition~\ref{prop:strength}) in the kernel \eqref{eq:kernel} and $R=\sum_j\phi^{*j}$ (Corollary~\ref{cor:response}), we obtain for $|z|\le1$
\begin{equation*}
 \begin{aligned}\hat\phi(z)&=\|\phi\|\sum_{l=1}^{K}\varpi_l\sum_{s\in\mathbb N\setminus\{0\}}(1-a_l)\,a_l^{s-1}z^s=\|\phi\|\sum_{l=1}^{K}\varpi_l\,\frac{(1-a_l)\,z}{1-a_lz},\\1-\hat\phi(z)&=(1-\|\phi\|)+\|\phi\|\sum_{l=1}^{K}\varpi_l\,\frac{1-z}{1-a_lz},\\\hat R(z)&=\sum_{j\in\mathbb N\setminus\{0\}}\hat\phi(z)^j=\frac{\hat\phi(z)}{1-\hat\phi(z)},\end{aligned}
\end{equation*}
where the last series converges as $|\hat\phi(z)|\le\|\phi\|<1$.

Next, let $x=1-\cos\theta$ and $y(\theta)=\pi\gamma/(2\theta)$. Every cluster with $\tau_l\ge y(\theta)$ satisfies $(\gamma/\tau_l)^2\le2x$, as $x\ge2\theta^2/\pi^2$ for $|\theta|\le\pi$. Hence
\begin{equation*}
 \begin{aligned}\mathrm{Re}\bigl(1-\hat\phi(e^{i\theta})\bigr)&=(1-\|\phi\|)+\|\phi\|\sum_{l=1}^{K}\varpi_l\,\frac{(1+a_l)\,x}{(1-a_l)^2+2a_lx}\\&\ge(1-\|\phi\|)+\|\phi\|\sum_{l=1}^{K}\varpi_l\,\frac{x}{(\gamma/\tau_l)^2+2x}\ge(1-\|\phi\|)+\frac{\|\phi\|}{4}\sum_{l:\,\tau_l\ge y(\theta)}\varpi_l .\end{aligned}
\end{equation*}
By the proof of Proposition~\ref{prop:manydesks}, $\varpi_l=\kappa_l\tau_l^{-\alpha}/\sum_j\kappa_j\tau_j^{-\alpha}$. For $\tau_1\le y\le\tau_K/2$ and $\Omega\le2$, let $l_0$ be the first cluster with $\tau_{l_0}\ge y$, such that $\tau_{l_0}<\Omega y$. Summing the geometric series in $\tau_l^{-\alpha}$ and using $\tau_K=\tau_{l_0}\Omega^{K-l_0}$, we obtain
\begin{equation*}
 \begin{aligned}\sum_{l:\,\tau_l\ge y}\varpi_l&\ge\frac{\inf\kappa}{\sup\kappa}\Bigl(\frac{\tau_{l_0}}{\tau_1}\Bigr)^{-\alpha}\bigl(1-\Omega^{-\alpha(K-l_0+1)}\bigr)\\&\ge\frac{\inf\kappa}{\sup\kappa}\Bigl(\frac{\Omega y}{\tau_1}\Bigr)^{-\alpha}\Bigl(1-\Bigl(\frac{y}{\tau_K}\Bigr)^{\alpha}\Bigr)\ge c_\varpi\Bigl(\frac{\tau_1}{y}\Bigr)^{\alpha},\qquad c_\varpi=\frac{\inf\kappa}{\sup\kappa}\,2^{-\alpha}\bigl(1-2^{-\alpha}\bigr),\end{aligned}
\end{equation*}
while for $y<\tau_1$ the sum equals one. With $\theta_K=\pi\gamma/\tau_K$, we have $y(\theta)\le\tau_K/2$ for $\theta\ge\theta_K$. Inserting $\tau_1/y(\theta)=2\tau_1\theta/(\pi\gamma)$, we obtain for $\theta_K\le\theta\le\pi$
\begin{equation*}
 \bigl|1-\hat\phi(e^{i\theta})\bigr|\ge(1-\|\phi\|)+c\,\min\Bigl\{1,\Bigl(\frac{\tau_1\theta}{\gamma}\Bigr)^{\alpha}\Bigr\},\qquad c=\frac{\|\phi\|\,c_\varpi}{4}\Bigl(\frac2\pi\Bigr)^{\alpha}.
\end{equation*}

Lastly, we split the integral at $\theta_K$ and at $\gamma/\tau_1$, and use $|\hat R|\le1/(1-\|\phi\|)$, $|\hat\phi|\le1$ and $|1-\hat\phi|\ge c$ on the three parts. Substituting $\theta=(\gamma/\tau_1)\bigl((1-\|\phi\|)/c\bigr)^{1/\alpha}u$ in the second part and using $\sum_s\phi(s)^2\le\phi(1)\,\|\phi\|\le\gamma/\tau_1$ for the decreasing $\phi$ in the third, we obtain with $\mathcal J_\alpha=\int_0^\infty du/(1+u^{\alpha})^2$
\begin{equation*}
 \begin{aligned}\sum_{s}R(s)^2&\le\frac1\pi\int_0^{\theta_K}\frac{d\theta}{(1-\|\phi\|)^2}+\frac1\pi\int_0^{\infty}\frac{d\theta}{\bigl((1-\|\phi\|)+c\,(\tau_1\theta/\gamma)^{\alpha}\bigr)^2}+\frac{1}{\pi c^2}\int_0^\pi\bigl|\hat\phi(e^{i\theta})\bigr|^2\,d\theta\\&\le\frac{\gamma}{\tau_K\,(1-\|\phi\|)^2}+\frac{\gamma}{\pi\,\tau_1}\,c^{-1/\alpha}\,(1-\|\phi\|)^{1/\alpha-2}\,\mathcal J_\alpha+\frac{\gamma}{c^2\,\tau_1}.\end{aligned}
\end{equation*}
Inserting $(1-\|\phi\|)^{1/\alpha}=(\|\phi\|\,\mathcal C_0)^{1/\alpha}/\mathcal T$, we obtain
\begin{equation*}
 \sum_{s}R(s)^2\le\frac{\gamma}{(1-\|\phi\|)^2\,\mathcal T}\Bigl(\frac{\mathcal T}{\tau_K}+\frac{\mathcal J_\alpha}{\pi\,\tau_1}\Bigl(\frac{\|\phi\|\,\mathcal C_0}{c}\Bigr)^{1/\alpha}+\frac{(1-\|\phi\|)^2\,\mathcal T}{c^2\,\tau_1}\Bigr),
\end{equation*}
where $\mathcal J_\alpha$ is finite because $2\alpha>1$. By (R1), $\mathcal T/\tau_K\to0$, $\Omega\le2$ and $\tau_K\ge\pi\tau_1$ for large $\mathcal T$. Moreover, $(1-\|\phi\|)^2\,\mathcal T=(\|\phi\|\,\mathcal C_0)^2\,\mathcal T^{1-2\alpha}\to0$ and $\|\phi\|\to1$. The bracket thus converges to a limit that depends only on $\alpha$, $\tau_1$, $\mathcal C_0$ and $\inf\kappa/\sup\kappa$, and twice this limit is a valid $C_R$.
\end{proof}

Inserting Lemma~\ref{lem:responsesq} into the variance of the expected size of the updates $\mathbb E[I_n\mid\mathcal F_n]$, we obtain
\begin{equation*}
 \mathrm{Var}\bigl(\mathbb E[I_n\mid\mathcal F_n]\bigr)\le\max_{m<n}\mathbb E\bigl[\tilde I_m^2\bigr]\;C_R\,\frac{\gamma}{(1-\|\phi\|)^2\,\mathcal T}.
\end{equation*}
With the conditional variance of Section~\ref{sec:correspondence} and the mean $\mathbb E[I_m]\le\bar I$ (Proposition~\ref{prop:longrun}), the second moment of each deviation $\tilde I_m$ satisfies
\begin{equation*}
 \mathbb E\bigl[\tilde I_m^2\bigr]=\mathbb E\Bigl[\sum_{l=1}^{K}\nu_l\bigl(1-p_l(m)\bigr)\kappa_l\,w_l\,p_l(m)\Bigr]\le\max_l\nu_l\;\mathbb E\bigl[I_m\bigr]\le\max_l\frac{\nu_l}{\bar\nu}\;\bar\nu\,\bar I .
\end{equation*}

\begin{lemma}[Second moments]\label{lem:moments}
Under the assumptions of Theorem~\ref{thm:rough}, let $q=p_0/(1-\|\phi\|)$ denote the mean activity of Proposition~\ref{prop:longrun}, and let
\begin{equation*}
 C_\nu=4\Bigl(\frac{\sup\kappa}{\inf\kappa}\Bigr)^{3},\qquad C_\lambda=1+\frac{2\,C_\nu\,C_R}{\mathcal C_2}.
\end{equation*}
Then for every sufficiently large $\mathcal T$, every cluster $l$ and every step $n$,
\begin{equation*}
 \begin{aligned}&\text{(i)}\ \ \nu_l\le C_\nu\,\bar\nu,\qquad\text{(ii)}\ \ \mathbb E\Bigl[\mathbb E\bigl[I_n\mid\mathcal F_n\bigr]^2\Bigr]\le C_\lambda\,\bar I^2,\\&\text{(iii)}\ \ \mathbb E\bigl[p_l(n)^2\bigr]\le3\,(1+C_\lambda)\,q^2+3\,C_\nu\,q\,\frac{\gamma\,\bar\nu}{\bar\kappa\,\tau_1}.\end{aligned}
\end{equation*}
\end{lemma}

By (R2), $q\to0$, and by Section~\ref{sec:correspondence}, $\gamma\bar\nu/\bar\kappa\to0$. Hence $\mathbb E[p_l(n)^2]/q\to0$ uniformly in $l$ and $n$.

\begin{proof}
For (i), the $2N$ desks of cluster $l$ contribute equally, such that $\nu_l=\kappa_lw_l/(2N)$ and $\nu_l/\bar\nu=\varpi_l/\sum_k\varpi_k^2$ (Section~\ref{sec:correspondence}). Bounding $\kappa_l$ by $\inf\kappa$ and $\sup\kappa$ in $\varpi_l=\kappa_l\tau_l^{-\alpha}/\sum_j\kappa_j\tau_j^{-\alpha}$ and summing the geometric series in $\tau_k=\tau_1\Omega^{k-1}$, we obtain
\begin{equation*}
 \frac{\nu_l}{\bar\nu}\le\Bigl(\frac{\sup\kappa}{\inf\kappa}\Bigr)^{3}\frac{\tau_1^{-\alpha}\sum_j\tau_j^{-\alpha}}{\sum_k\tau_k^{-2\alpha}}\le\Bigl(\frac{\sup\kappa}{\inf\kappa}\Bigr)^{3}\frac{1-\Omega^{-2\alpha}}{(1-\Omega^{-\alpha})\bigl(1-\Omega^{-2\alpha K}\bigr)}=\Bigl(\frac{\sup\kappa}{\inf\kappa}\Bigr)^{3}\frac{1+\Omega^{-\alpha}}{1-\Omega^{-2\alpha K}} .
\end{equation*}
As $\Omega^{K-1}=\tau_K/\tau_1\to\infty$ by (R1), the last fraction is at most $4$ for large $\mathcal T$.

For (ii), we insert (i) into the bound on $\mathbb E[\tilde I_m^2]$ and this bound into the bound on $\mathrm{Var}(\mathbb E[I_n\mid\mathcal F_n])$. With $\bar I=p_0\bar\kappa/(1-\|\phi\|)$, we obtain
\begin{equation*}
 \frac{\mathrm{Var}\bigl(\mathbb E[I_n\mid\mathcal F_n]\bigr)}{\bar I^2}\le C_\nu\,\frac{\bar\nu}{\bar I}\,C_R\,\frac{\gamma}{(1-\|\phi\|)^2\,\mathcal T}=C_\nu\,C_R\,\frac{\gamma\,\bar\nu/\bar\kappa}{\mathcal T\,(1-\|\phi\|)\,p_0}\to\frac{C_\nu\,C_R}{\mathcal C_2},
\end{equation*}
where the limit follows from the second condition in (R2) and $\|\phi\|\to1$. Adding $\mathbb E[\mathbb E[I_n\mid\mathcal F_n]]^2=\mathbb E[I_n]^2\le\bar I^2$, we obtain (ii) for large $\mathcal T$.

For (iii), we split $U_l$ with $I_m=\mathbb E[I_m\mid\mathcal F_m]+\tilde I_m$. Since the weights $\omega_l$ sum to at most one and the deviations $\tilde I_m$ are uncorrelated, we obtain from Minkowski's inequality
\begin{equation*}
 \begin{aligned}\mathbb E\bigl[p_l(n)^2\bigr]^{1/2}&\le p_0+\beta\sum_{m=0}^{n-1}\omega_l(n-m)\,\mathbb E\Bigl[\mathbb E\bigl[I_m\mid\mathcal F_m\bigr]^2\Bigr]^{1/2}+\beta\Bigl(\sum_{m=0}^{n-1}\omega_l(n-m)^2\,\mathbb E\bigl[\tilde I_m^2\bigr]\Bigr)^{1/2}\\&\le q+\beta\,\sqrt{C_\lambda}\,\bar I+\beta\Bigl(\frac{\gamma}{\tau_l}\,C_\nu\,\bar\nu\,\bar I\Bigr)^{1/2},\end{aligned}
\end{equation*}
where we used $p_0\le q$, (ii), $\omega_l\le\gamma/\tau_l$ (Lemma~\ref{lem:timescales}) and the bound on $\mathbb E[\tilde I_m^2]$ above. By Proposition~\ref{prop:strength}, $\beta\bar I=\|\phi\|\,q\le q$ and $\beta^2\bar\nu\,\bar I=\|\phi\|^2q\,\bar\nu/\bar\kappa$. Squaring with $(a+b+c)^2\le3(a^2+b^2+c^2)$ and using $\tau_l\ge\tau_1$, we obtain (iii).
\end{proof}

Splitting $I_n=\mathbb E[I_n\mid\mathcal F_n]+\tilde I_n$ in $\mathcal X^{\mathcal T}$ into the two quantities of Lemma~\ref{lem:moments}, we obtain
\begin{equation*}
 \mathcal X^{\mathcal T}_t=\frac{\gamma}{\mathcal T}\sum_{n\le\mathcal Tt/\gamma}\frac{\mathbb E[I_n\mid\mathcal F_n]}{\bar I}+\varepsilon_{\mathcal T}\,\mathcal Z^{\mathcal T}_t,\qquad\mathcal Z^{\mathcal T}_t=\Bigl(\frac{\gamma}{\mathcal T\,\bar I\,\bar\nu}\Bigr)^{1/2}\sum_{n\le\mathcal Tt/\gamma}\tilde I_n,\qquad\varepsilon_{\mathcal T}=\Bigl(\frac{\gamma\,\bar\nu}{\mathcal T\,\bar I}\Bigr)^{1/2},
\end{equation*}
where $\mathcal Z^{\mathcal T}$ is the rescaled deviation of Section~\ref{sec:threshold} under the correspondence of Section~\ref{sec:correspondence}. By the second condition in (R2),
\begin{equation*}
 \varepsilon_{\mathcal T}^2=\frac{(1-\|\phi\|)\,\gamma\,\bar\nu/\bar\kappa}{\mathcal T\,p_0}=\frac{(1-\|\phi\|)^2}{\|\phi\|\,\mathcal C_2}\bigl(1+o(1)\bigr)\to0 .
\end{equation*}

\begin{lemma}[Increments of the size of the updates]\label{lem:increments}
Under the assumptions of Theorem~\ref{thm:rough}, for every sufficiently large $\mathcal T$ and all $0\le u\le t\le1$,
\begin{equation*}
 \mathbb E\Bigl[\bigl(\mathcal X^{\mathcal T}_t-\mathcal X^{\mathcal T}_u\bigr)^2\Bigr]\le2\,C_\lambda\Bigl(t-u+\frac{\gamma}{\mathcal T}\Bigr)^2+2\,C_\nu\,\varepsilon_{\mathcal T}^2\Bigl(t-u+\frac{\gamma}{\mathcal T}\Bigr).
\end{equation*}
Moreover, for every $\delta\in[\gamma/\mathcal T,1]$ and $\eta>0$,
\begin{equation*}
 \mathbb P\Bigl(\sup_{0\le u\le t\le1,\ t-u\le\delta}\bigl(\mathcal X^{\mathcal T}_t-\mathcal X^{\mathcal T}_u\bigr)>\eta\Bigr)\le\frac{36\,C_\lambda\,\delta+12\,C_\nu\,\varepsilon_{\mathcal T}^2}{\eta^2}.
\end{equation*}
\end{lemma}

As the right side vanishes for $\mathcal T\to\infty$ and then $\delta\to0$, the rescaled size of the updates has no jumps in the limit.

\begin{proof}
The interval $(u,t]$ holds at most $\mathcal T(t-u)/\gamma+1$ steps $n$ with $\mathcal Tu/\gamma<n\le\mathcal Tt/\gamma$. For the expected size of the updates, we obtain from Minkowski's inequality and Lemma~\ref{lem:moments}(ii)
\begin{equation*}
 \Bigl\|\frac{\gamma}{\mathcal T}\sum_{\mathcal Tu/\gamma<n\le\mathcal Tt/\gamma}\frac{\mathbb E[I_n\mid\mathcal F_n]}{\bar I}\Bigr\|_{L^2}\le\frac{\gamma}{\mathcal T}\Bigl(\frac{\mathcal T(t-u)}{\gamma}+1\Bigr)\sqrt{C_\lambda}=\sqrt{C_\lambda}\Bigl(t-u+\frac{\gamma}{\mathcal T}\Bigr).
\end{equation*}
As the deviations are uncorrelated, we obtain with $\mathbb E[\tilde I_m^2]\le C_\nu\bar\nu\,\bar I$ from Lemma~\ref{lem:moments}(i)
\begin{equation*}
 \varepsilon_{\mathcal T}^2\,\mathbb E\Bigl[\bigl(\mathcal Z^{\mathcal T}_t-\mathcal Z^{\mathcal T}_u\bigr)^2\Bigr]=\frac{\gamma^2}{\mathcal T^2\,\bar I^2}\sum_{\mathcal Tu/\gamma<n\le\mathcal Tt/\gamma}\mathbb E\bigl[\tilde I_n^2\bigr]\le C_\nu\,\varepsilon_{\mathcal T}^2\Bigl(t-u+\frac{\gamma}{\mathcal T}\Bigr).
\end{equation*}
With $(a+b)^2\le2a^2+2b^2$, we obtain the first claim. For the second claim, let $t_k=\min\{k\delta,1\}$ for $k=0,\dots,\lceil1/\delta\rceil$. As $\mathcal X^{\mathcal T}$ does not decrease, every pair $u\le t$ with $t-u\le\delta$ lies in some $[t_k,t_{k+2}]$, such that
\begin{equation*}
 \sup_{t-u\le\delta}\bigl(\mathcal X^{\mathcal T}_t-\mathcal X^{\mathcal T}_u\bigr)\le\max_k\bigl(\mathcal X^{\mathcal T}_{t_{k+2}}-\mathcal X^{\mathcal T}_{t_k}\bigr).
\end{equation*}
Summing Chebyshev's inequality over the at most $2/\delta$ values of $k$ and inserting the first claim with $t_{k+2}-t_k+\gamma/\mathcal T\le3\delta$, we obtain
\begin{equation*}
 \mathbb P\Bigl(\max_k\bigl(\mathcal X^{\mathcal T}_{t_{k+2}}-\mathcal X^{\mathcal T}_{t_k}\bigr)>\eta\Bigr)\le\frac{2}{\delta\,\eta^2}\Bigl(18\,C_\lambda\,\delta^2+6\,C_\nu\,\varepsilon_{\mathcal T}^2\,\delta\Bigr)=\frac{36\,C_\lambda\,\delta+12\,C_\nu\,\varepsilon_{\mathcal T}^2}{\eta^2}.\qedhere
\end{equation*}
\end{proof}

For the rescaled deviation $\mathcal Z^{\mathcal T}$ of our market, the process $\langle\mathcal Z^{\mathcal T}\rangle$ of Section~\ref{sec:threshold} is the sum of the conditional variances of its increments,
\begin{equation*}
 \langle\mathcal Z^{\mathcal T}\rangle_t=\frac{\gamma}{\mathcal T\,\bar I\,\bar\nu}\sum_{n\le\mathcal Tt/\gamma}\mathrm{Var}\bigl[I_n\mid\mathcal F_n\bigr],\qquad t\in[0,1].
\end{equation*}
Let $n_t=\lfloor\mathcal Tt/\gamma\rfloor$. Dividing Lemma~\ref{lem:summedvar} by $\mathcal T\bar I\bar\nu/\gamma$ and inserting $(\gamma/\mathcal T)\sum_{n\le n_t}\mathbb E[I_n\mid\mathcal F_n]/\bar I=\mathcal X^{\mathcal T}_t-\varepsilon_{\mathcal T}\mathcal Z^{\mathcal T}_t$, we obtain
\begin{equation*}
 \begin{aligned}\langle\mathcal Z^{\mathcal T}\rangle_t&=\mathcal X^{\mathcal T}_t-\varepsilon_{\mathcal T}\,\mathcal Z^{\mathcal T}_t-D^{\mathcal T}_t,\\D^{\mathcal T}_t&=\frac{\gamma}{\mathcal T\,\bar I\,\bar\nu}\Bigl(\beta\sum_{l=1}^{K}(\nu_l-\bar\nu)\,\kappa_l\,w_l\,(T_l-1)\,U_l(n_t)+\sum_{l=1}^{K}\nu_l\,\kappa_l\,w_l\sum_{r=0}^{n_t}p_l(r)^2\Bigr).\end{aligned}
\end{equation*}

\begin{lemma}[Conditional variance in the limit]\label{lem:condvarlimit}
Under the assumptions of Theorem~\ref{thm:rough}, as $\mathcal T\to\infty$,
\begin{equation*}
 \sup_{t\le1}\bigl|D^{\mathcal T}_t\bigr|\to0\qquad\text{and}\qquad\sup_{t\le1}\bigl|\langle\mathcal Z^{\mathcal T}\rangle_t-\mathcal X^{\mathcal T}_t\bigr|\to0\qquad\text{in probability}.
\end{equation*}
\end{lemma}

In units of $\bar\nu$, the conditional variance of the size of the updates thus equals its conditional mean in the limit, as for a number of actions.

\begin{proof}
We first bound the part of $D^{\mathcal T}$ with the response after the step $n_t$. Let $\Delta\mathcal X_m=\gamma I_m/(\mathcal T\bar I)$ denote the increments of $\mathcal X^{\mathcal T}$. Using $|\nu_l-\bar\nu|\le C_\nu\bar\nu$ (Lemma~\ref{lem:moments}(i)), $\beta\kappa_lw_l=\|\phi\|\varpi_l$ (Proposition~\ref{prop:strength}) and the proof of Lemma~\ref{lem:summedprob}, we obtain
\begin{equation*}
 \frac{\gamma}{\mathcal T\,\bar I\,\bar\nu}\,\beta\sum_{l=1}^{K}\bigl|\nu_l-\bar\nu\bigr|\,\kappa_l\,w_l\,(T_l-1)\,U_l(n_t)\le C_\nu\sum_{l=1}^{K}\varpi_l\,S_l(t),\qquad S_l(t)=\sum_{m<n_t}e^{-(n_t-m)\gamma/\tau_l}\,\Delta\mathcal X_m .
\end{equation*}
For $\delta\in[\gamma/\mathcal T,1]$, the increments within the time $\delta$ before $t$ sum to at most the largest increment $w_{\mathcal T}(2\delta)$ of $\mathcal X^{\mathcal T}$ over intervals of length $2\delta$, while the earlier increments carry the weight at most $e^{-\delta\mathcal T/\tau_l}$. Bounding the weights of the long horizons by the geometric series in $\tau_l^{-\alpha}$ as in the proof of Lemma~\ref{lem:moments}(i), we thus obtain
\begin{equation*}
 \sum_{l=1}^{K}\varpi_l\,S_l(t)\le w_{\mathcal T}(2\delta)+\mathcal X^{\mathcal T}_1\Bigl(e^{-1/\delta}+\sum_{l:\,\tau_l>\delta^2\mathcal T}\varpi_l\Bigr)\le w_{\mathcal T}(2\delta)+\mathcal X^{\mathcal T}_1\Bigl(e^{-1/\delta}+2\,\frac{\sup\kappa}{\inf\kappa}\Bigl(\frac{\tau_1}{\delta^2\mathcal T}\Bigr)^{\alpha}\Bigr).
\end{equation*}
The right side does not depend on $t$. By Lemma~\ref{lem:increments}, $\mathbb P(w_{\mathcal T}(2\delta)>\eta)\le(72\,C_\lambda\delta+12\,C_\nu\varepsilon_{\mathcal T}^2)/\eta^2$ for every $\eta>0$, while $\mathbb E[\mathcal X^{\mathcal T}_1]\le1+\gamma/\mathcal T$ by Proposition~\ref{prop:longrun}. Letting first $\mathcal T\to\infty$ and then $\delta\to0$, we obtain that this part of $\sup_t|D^{\mathcal T}_t|$ tends to zero in probability.

The part of $D^{\mathcal T}$ with the squared probabilities of acting does not decrease in $t$. With $\nu_l\le C_\nu\bar\nu$, $\bar\kappa/\bar I=1/q$ and Lemma~\ref{lem:moments}(iii), we obtain
\begin{equation*}
 \mathbb E\Bigl[\frac{\gamma}{\mathcal T\,\bar I\,\bar\nu}\sum_{l=1}^{K}\nu_l\,\kappa_l\,w_l\sum_{r=0}^{n_1}p_l(r)^2\Bigr]\le C_\nu\Bigl(1+\frac{\gamma}{\mathcal T}\Bigr)\max_{l,\,r}\frac{\mathbb E\bigl[p_l(r)^2\bigr]}{q}\to0,
\end{equation*}
such that this part tends to zero in probability by Markov's inequality. Lastly, we obtain from Doob's inequality and the proof of Lemma~\ref{lem:increments} that $\mathbb E[\sup_{t\le1}|\mathcal Z^{\mathcal T}_t|^2]\le4\,\mathbb E[(\mathcal Z^{\mathcal T}_1)^2]\le4\,C_\nu(1+\gamma/\mathcal T)$. As $\varepsilon_{\mathcal T}\to0$, the term $\varepsilon_{\mathcal T}\mathcal Z^{\mathcal T}$ thus tends to zero uniformly in probability, and the second claim follows from the first.
\end{proof}

The second identity of Section~\ref{sec:threshold}, $[\mathcal Z^{\mathcal T},\mathcal Z^{\mathcal T}]=\mathcal X^{\mathcal T}$, holds because the number of actions jumps by one. In our market, the jump of $\mathcal Z^{\mathcal T}$ in step $n$ is $(\gamma/(\mathcal T\bar I\bar\nu))^{1/2}\tilde I_n$. The sum of the squared jumps of $\mathcal Z^{\mathcal T}$ is therefore
\begin{equation*}
 [\mathcal Z^{\mathcal T},\mathcal Z^{\mathcal T}]_t=\frac{\gamma}{\mathcal T\,\bar I\,\bar\nu}\sum_{n\le\mathcal Tt/\gamma}\tilde I_n^2 .
\end{equation*}
Subtracting the sum of the conditional variances $\langle\mathcal Z^{\mathcal T}\rangle$, we obtain
\begin{equation*}
 [\mathcal Z^{\mathcal T},\mathcal Z^{\mathcal T}]_t-\langle\mathcal Z^{\mathcal T}\rangle_t=\frac{\gamma}{\mathcal T\,\bar I\,\bar\nu}\sum_{n\le\mathcal Tt/\gamma}\Bigl(\tilde I_n^2-\mathrm{Var}\bigl[I_n\mid\mathcal F_n\bigr]\Bigr).
\end{equation*}

\begin{lemma}[Squared jumps in the limit]\label{lem:jumps}
Under the assumptions of Theorem~\ref{thm:rough}, as $\mathcal T\to\infty$,
\begin{equation*}
 \mathbb E\Bigl[\sup_{t\le1}\bigl([\mathcal Z^{\mathcal T},\mathcal Z^{\mathcal T}]_t-\langle\mathcal Z^{\mathcal T}\rangle_t\bigr)^2\Bigr]\to0,\qquad\max_{n\le\mathcal T/\gamma}\Bigl(\frac{\gamma}{\mathcal T\,\bar I\,\bar\nu}\Bigr)^{1/2}|\tilde I_n|\to0,\qquad\max_{n\le\mathcal T/\gamma}\frac{\gamma\,I_n}{\mathcal T\,\bar I}\to0,
\end{equation*}
where the last two limits hold in probability. In particular, $\sup_{t\le1}\bigl|[\mathcal Z^{\mathcal T},\mathcal Z^{\mathcal T}]_t-\mathcal X^{\mathcal T}_t\bigr|\to0$ in probability.
\end{lemma}

In the limit, the size of the updates thus satisfies both identities of Section~\ref{sec:threshold}, while the jumps of $\mathcal Z^{\mathcal T}$ and $\mathcal X^{\mathcal T}$ vanish.

\begin{proof}
Given $\mathcal F_n$, the deviation $\tilde I_n$ is the sum of the independent contributions $\nu_l\bigl(\xi_i(n)-p_l(n)\bigr)$ of the desks, which have mean zero and satisfy $|\nu_l(\xi_i(n)-p_l(n))|\le C_\nu\bar\nu$ by Lemma~\ref{lem:moments}(i). Expanding the fourth power and using $\mathrm{Var}[I_n\mid\mathcal F_n]\le C_\nu\bar\nu\,\mathbb E[I_n\mid\mathcal F_n]$, we obtain
\begin{equation*}
 \begin{aligned}\mathbb E\bigl[\tilde I_n^4\mid\mathcal F_n\bigr]&\le C_\nu^2\,\bar\nu^2\,\mathrm{Var}\bigl[I_n\mid\mathcal F_n\bigr]+3\,\mathrm{Var}\bigl[I_n\mid\mathcal F_n\bigr]^2\\&\le C_\nu^3\,\bar\nu^3\,\mathbb E\bigl[I_n\mid\mathcal F_n\bigr]+3\,C_\nu^2\,\bar\nu^2\,\mathbb E\bigl[I_n\mid\mathcal F_n\bigr]^2 .\end{aligned}
\end{equation*}
Taking expectations with $\mathbb E[I_n]\le\bar I$ and Lemma~\ref{lem:moments}(ii), and summing over the at most $\mathcal T/\gamma+1$ steps, we obtain with $\varepsilon_{\mathcal T}^2=\gamma\bar\nu/(\mathcal T\bar I)$
\begin{equation*}
 \Bigl(\frac{\gamma}{\mathcal T\,\bar I\,\bar\nu}\Bigr)^2\sum_{n\le\mathcal T/\gamma}\mathbb E\bigl[\tilde I_n^4\bigr]\le\Bigl(1+\frac{\gamma}{\mathcal T}\Bigr)\Bigl(C_\nu^3\,\varepsilon_{\mathcal T}^2+3\,C_\nu^2\,C_\lambda\,\frac{\gamma}{\mathcal T}\Bigr)\to0 .
\end{equation*}
The differences $\tilde I_n^2-\mathrm{Var}[I_n\mid\mathcal F_n]$ have conditional mean zero and second moment at most $\mathbb E[\tilde I_n^4]$. By Doob's inequality, the expectation in the first limit is thus at most four times this sum of fourth moments. Moreover, we obtain from Markov's inequality with the fourth power
\begin{equation*}
 \mathbb P\Bigl(\max_{n\le\mathcal T/\gamma}\Bigl(\frac{\gamma}{\mathcal T\,\bar I\,\bar\nu}\Bigr)^{1/2}|\tilde I_n|>\eta\Bigr)\le\frac{1}{\eta^4}\Bigl(\frac{\gamma}{\mathcal T\,\bar I\,\bar\nu}\Bigr)^2\sum_{n\le\mathcal T/\gamma}\mathbb E\bigl[\tilde I_n^4\bigr]\to0 .
\end{equation*}
For the jumps of $\mathcal X^{\mathcal T}$, we write $\gamma I_n/(\mathcal T\bar I)=\gamma\,\mathbb E[I_n\mid\mathcal F_n]/(\mathcal T\bar I)+\varepsilon_{\mathcal T}(\gamma/(\mathcal T\bar I\bar\nu))^{1/2}\tilde I_n$. The second term vanishes by the previous limit and $\varepsilon_{\mathcal T}\to0$, while we obtain from Chebyshev's inequality and Lemma~\ref{lem:moments}(ii)
\begin{equation*}
 \mathbb P\Bigl(\max_{n\le\mathcal T/\gamma}\frac{\gamma\,\mathbb E[I_n\mid\mathcal F_n]}{\mathcal T\,\bar I}>\eta\Bigr)\le\frac{1}{\eta^2}\Bigl(\frac{\mathcal T}{\gamma}+1\Bigr)\frac{\gamma^2}{\mathcal T^2}\,C_\lambda=\frac{C_\lambda}{\eta^2}\,\frac{\gamma}{\mathcal T}\Bigl(1+\frac{\gamma}{\mathcal T}\Bigr)\to0 .
\end{equation*}
Lastly, the last claim follows from the first limit and Lemma~\ref{lem:condvarlimit}.
\end{proof}

For our market, the normalized response $\rho^{\mathcal T}$ of Section~\ref{sec:threshold} becomes the total of $R$ up to the time $\mathcal Tx$ in units of $\sum_sR(s)=\|\phi\|/(1-\|\phi\|)$ (Proposition~\ref{prop:longrun}),
\begin{equation*}
 F^{\mathcal T}(x)=\frac{1-\|\phi\|}{\|\phi\|}\sum_{s\le\mathcal Tx/\gamma}R(s),\qquad x\ge0 .
\end{equation*}

\begin{lemma}[Response in the limit]\label{lem:responselimit}
Under the assumptions of Theorem~\ref{thm:rough}, let $F(x)=\int_0^x\mathcal G(y)\,dy$. Then, as $\mathcal T\to\infty$,
\begin{equation*}
 \sup_{x\in[0,1]}\bigl|F^{\mathcal T}(x)-F(x)\bigr|\to0 .
\end{equation*}
\end{lemma}

On the time scale $\mathcal T$, the response of our market thus has the same limit $\mathcal G$ as the response of the Hawkes processes in Section~\ref{sec:threshold}.

\begin{proof}
Let $z>0$, $w=e^{-z\gamma/\mathcal T}$ and $h(y)=y/(1+y)$. By the proof of Lemma~\ref{lem:responsesq}, $\hat R=\hat\phi/(1-\hat\phi)$ with $1-\hat\phi(w)/\|\phi\|=\sum_l\varpi_l(1-w)/(1-a_lw)$. Dividing the numerator and the denominator of each term by $(1-a_l)\,w=\gamma w/\tau_l$, we obtain
\begin{equation*}
 1-\frac{\hat\phi(w)}{\|\phi\|}=\sum_{l=1}^{K}\varpi_l\,\frac{(1-w)/(\gamma w/\tau_l)}{(1-w)/(\gamma w/\tau_l)+1}=\sum_{l=1}^{K}\varpi_l\,h\Bigl(\frac{z_{\mathcal T}\,\tau_l}{\mathcal T}\Bigr)=:S_{\mathcal T}(z_{\mathcal T}),\qquad z_{\mathcal T}=\frac{\mathcal T}{\gamma}\bigl(e^{z\gamma/\mathcal T}-1\bigr).
\end{equation*}
With $1-\|\phi\|=\|\phi\|\,\mathcal C_0\,\mathcal T^{-\alpha}$, the Laplace transform of the measure with the distribution function $F^{\mathcal T}$ thus becomes
\begin{equation*}
 \sum_{s\in\mathbb N\setminus\{0\}}e^{-zs\gamma/\mathcal T}\,\frac{1-\|\phi\|}{\|\phi\|}\,R(s)=\frac{1-\|\phi\|}{\|\phi\|}\,\frac{\hat\phi(w)}{1-\hat\phi(w)}=\frac{1-S_{\mathcal T}(z_{\mathcal T})}{1+\mathcal C_0^{-1}\,\mathcal T^{\alpha}\,S_{\mathcal T}(z_{\mathcal T})}.
\end{equation*}

We next show that $\mathcal T^{\alpha}S_{\mathcal T}(z')\to\mathcal C_0\,z'^{\alpha}$ for every $z'>0$. By the proof of Proposition~\ref{prop:manydesks}, $\varpi_l=\kappa_l\tau_l^{-\alpha}/\sum_j\kappa_j\tau_j^{-\alpha}$ with $\kappa_l$ proportional to $\kappa(\tau_l)$. Multiplying the numerator and the denominator by $\ln\Omega$ and writing $x_l=\tau_l/\mathcal T$, we obtain
\begin{equation*}
 \mathcal T^{\alpha}S_{\mathcal T}(z')=\frac{\ln\Omega\sum_{l=1}^{K}\kappa(\mathcal Tx_l)\,x_l^{-\alpha}\,h(z'x_l)}{\ln\Omega\sum_{j=1}^{K}\kappa(\tau_j)\,\tau_j^{-\alpha}}.
\end{equation*}
By the proof of Proposition~\ref{prop:manydesks}, the denominator converges to $\int_{\tau_1}^\infty\kappa(\tau)\,\tau^{-1-\alpha}\,d\tau=\bar\kappa_\infty/(\alpha\tau_1^{\alpha})$. The numerator is a Riemann sum with the step $\ln\Omega$ in $\ln x$, over $x_l$ from $\tau_1/\mathcal T\to0$ to $\tau_K/\mathcal T\to\infty$ by (R1). For $0<x_0<X$, $\kappa(\mathcal Tx)\to\kappa_\infty$ uniformly on $[x_0,X]$, such that the terms with $x_l\in[x_0,X]$ converge to $\kappa_\infty\int_{x_0}^Xx^{-\alpha}h(z'x)\,dx/x$. As $h(z'x)\le z'x$ and $h\le1$, the remaining terms satisfy
\begin{equation*}
 \begin{aligned}&\ln\Omega\sum_{l:\,x_l<x_0}\kappa(\mathcal Tx_l)\,x_l^{-\alpha}h(z'x_l)\le\sup\kappa\,\frac{z'\,\ln\Omega}{1-\Omega^{-(1-\alpha)}}\,x_0^{1-\alpha},\\&\ln\Omega\sum_{l:\,x_l>X}\kappa(\mathcal Tx_l)\,x_l^{-\alpha}h(z'x_l)\le\sup\kappa\,\frac{\ln\Omega}{1-\Omega^{-\alpha}}\,X^{-\alpha},\end{aligned}
\end{equation*}
where $\ln\Omega/(1-\Omega^{-\beta})\to1/\beta$ as $\Omega\to1^+$. Letting $\mathcal T\to\infty$, then $x_0\to0$ and $X\to\infty$, substituting $y=z'x$ and inserting $k_0=\alpha\tau_1^{\alpha}\kappa_\infty\Gamma(1+\alpha)/\bar\kappa_\infty$ (Proposition~\ref{prop:manydesks}) into $\mathcal C_0=(k_0/\alpha)\Gamma(1-\alpha)$ (Section~\ref{sec:threshold}), we obtain
\begin{equation*}
 \mathcal T^{\alpha}S_{\mathcal T}(z')\to\frac{\alpha\,\tau_1^{\alpha}\,\kappa_\infty}{\bar\kappa_\infty}\int_0^\infty x^{-\alpha}h(z'x)\,\frac{dx}{x}=\frac{\alpha\,\tau_1^{\alpha}\,\kappa_\infty}{\bar\kappa_\infty}\,\Gamma(\alpha)\,\Gamma(1-\alpha)\,z'^{\alpha}=\mathcal C_0\,z'^{\alpha}.
\end{equation*}

Lastly, $S_{\mathcal T}$ increases in its argument, and $z_{\mathcal T}\to z$ as $\gamma/\mathcal T\to0$. For every $\eta\in(0,1)$, we thus have $\mathcal C_0(1-\eta)^{\alpha}z^{\alpha}\le\liminf\mathcal T^{\alpha}S_{\mathcal T}(z_{\mathcal T})\le\limsup\mathcal T^{\alpha}S_{\mathcal T}(z_{\mathcal T})\le\mathcal C_0(1+\eta)^{\alpha}z^{\alpha}$, while $S_{\mathcal T}(z_{\mathcal T})\to0$. The Laplace transform above therefore converges to $1/(1+z^{\alpha})$, which is the Laplace transform of $\mathcal G$ (Section~\ref{sec:threshold}). By the continuity theorem for Laplace transforms \cite[Section~XIII.1]{Fel71}, the measures with the distribution functions $F^{\mathcal T}$ thus converge weakly to the measure with the density $\mathcal G$. As $F$ is continuous, $F^{\mathcal T}$ then converges to $F$ uniformly.
\end{proof}

\begin{proof}[Proof of Theorem~\ref{thm:rough}]
We follow the proof of Theorem~\ref{thm:jr16} in \cite[Sections~4.1--4.4]{JR16}, where the Hawkes process enters in three places. The bound on the mean of $N^{\mathcal T}$ \cite[Lemma~4.1]{JR16} becomes the bound on the increments of $\mathcal X^{\mathcal T}$ in Lemma~\ref{lem:increments}. The two identities for $M^{\mathcal T}$ (Section~\ref{sec:threshold}) become their limits in Lemmas~\ref{lem:condvarlimit} and~\ref{lem:jumps}. The limit of the normalized response \cite[Lemma~4.3]{JR16} becomes Lemma~\ref{lem:responselimit}. With sums over steps in place of integrals over time, the remaining steps apply unchanged. By the correspondence of Section~\ref{sec:correspondence}, the limit has the constants $\mathcal C_1=1$ and $\mathcal C_2$ of (R2).
\end{proof}

\runin{The mid price}In step $n$, each desk $d_i$ with $\xi_i(n)=1$ changes $Z^{\sigma_i}$ by $m_i\Delta Z_i(n)$ (Section~\ref{sec:rule}). By \eqref{eq:mid}, the next move of the mid price is therefore
\begin{equation*}
 \theta_{n+1}=-c\bigl(\tilde Z^+_n-\tilde Z^-_n\bigr)-c\sum_i\sigma_i\,\xi_i(n)\,m_i\,\Delta Z_i(n),
\end{equation*}
where $\tilde Z^\sigma_n$ is $Z^\sigma$ after the shift of step $n$ (Section~\ref{sec:protocol}).

\begin{corollary}[Volatility of the mid price]\label{cor:mid}
Under the assumptions of Theorem~\ref{thm:rough}, let
\begin{equation*}
 \mathcal P^{\mathcal T}_t=\Bigl(\frac{\gamma}{\mathcal T\,\bar I\,\bar\nu}\Bigr)^{1/2}\sum_{n\le\mathcal Tt/\gamma}\Bigl(\theta_{n+1}-\mathbb E\bigl[\theta_{n+1}\mid\mathcal F_n\bigr]\Bigr),\qquad t\in[0,1],
\end{equation*}
denote the rescaled fluctuation of the mid price. Then every sequence $\mathcal T_k\to\infty$ has a subsequence along which $(\mathcal P^{\mathcal T_k},\mathcal X^{\mathcal T_k})$ converges in law, and every limit satisfies
\begin{equation*}
 \mathcal P_t=\int_0^t\sqrt{\mathcal X'_s}\,dW_s,\qquad t\in[0,1],
\end{equation*}
for a Brownian motion $W$ and the derivative $\mathcal X'$ of Theorem~\ref{thm:rough}.
\end{corollary}

In the limit, the squared volatility $\mathcal X'$ of the mid price is thus as rough as fractional Brownian motion with the Hurst parameter $H=\alpha-1/2$.

\begin{proof}
Given $\mathcal F_n$, the indicators $\xi_i(n)$ are independent (Section~\ref{sec:spde}), while $\Delta Z_i(n)$ and $\tilde Z^\sigma_n$ are $\mathcal F_n$-measurable (proof of Lemma~\ref{lem:size}). With \eqref{eq:push}, we obtain
\begin{equation*}
 \mathrm{Var}\bigl[\theta_{n+1}\mid\mathcal F_n\bigr]=\sum_i\bigl(c\,m_i|\Delta Z_i(n)|\bigr)^2p_{k(i)}(n)\bigl(1-p_{k(i)}(n)\bigr)=\mathrm{Var}\bigl[I_n\mid\mathcal F_n\bigr].
\end{equation*}
The fluctuation of $\theta_{n+1}$ differs from $\tilde I_n$ only in the signs $\sigma_i$ of its terms. Lemmas~\ref{lem:condvarlimit} and~\ref{lem:jumps} and the proof of Theorem~\ref{thm:rough} thus apply to $\mathcal P^{\mathcal T}$ in place of $\mathcal Z^{\mathcal T}$, such that every limit $\mathcal P$ is a continuous martingale with $[\mathcal P,\mathcal P]=\mathcal X$. By \cite[Chapter~V, (3.8)]{RY99}, we obtain $\mathcal P=\int_0^{\cdot}\sqrt{\mathcal X'_s}\,dW_s$.
\end{proof}

\section{Numerical experiments and results}\label{sec:simulation}

We test each step of our mechanism on a simulation of the protocol of Section~\ref{sec:protocol} over a finite time. The steps lead from the book (Proposition~\ref{prop:dk}) to the activity (Corollary~\ref{cor:response}) and on to the mid price (Corollary~\ref{cor:mid}). For each step we compare a quantity measured in the simulation with its value in the limit.

\subsection{Setting and implementation}\label{sec:simsetting}
We measure every order by its distance $x>0$ from the mid price (Section~\ref{sec:book}). As a function of this distance, the volume of average order books is known empirically to peak away from the best price \cite{BMP02,GCZ08,PB03}. We therefore draw each target from the Gamma density $x\,e^{-x/1.5}/1.5^{2}$ (Table~\ref{tab:hyper}). This density vanishes at $x=0$ and peaks at $x=1.5$.

\subsection{The Dean--Kawasaki check}\label{sec:dkcheck}
For the test function $\varphi_d(x)=e^{-(x-d)^2/2}$ at the depth $d\in\{1,3,6\}$ we follow the volume of the asks around $d$,
\[
V_n(d)=\sum_{i:\,\sigma_i=+1}m_i\sum_{j=1}^{L}\varphi_d\bigl(x_{i,j}(n)\bigr).
\]
By Proposition~\ref{prop:dk}, its change $\Delta_n(d)=V_{n+1}(d)-V_n(d)$ has the conditional mean $\mu_n(d)$. By Lemma~\ref{lem:noise}, the orders of one desk move together, such that the conditional variance sums over desks rather than over orders,
\[
\begin{aligned}
s_n^2&=\sum_{i:\,\sigma_i=+1}p_{k(i)}\bigl(1-p_{k(i)}\bigr)\Bigl(m_i\sum_{j}\varphi_d'(x_{i,j})\,g_{i,j}\Bigr)^2,\\
\bigl(s^{\mathrm{ind}}_n\bigr)^2&=\sum_{i:\,\sigma_i=+1}\sum_{j}p_{k(i)}\bigl(1-p_{k(i)}\bigr)\bigl(m_i\,\varphi_d'(x_{i,j})\,g_{i,j}\bigr)^2 .
\end{aligned}
\]
In the limit the slope of $\Delta_n(d)$ on $\mu_n(d)$ is one, and $(\Delta_n(d)-\mu_n(d))/s_n$ has variance one.

\begin{table}[H]
\centering\small
\caption{Change of the book against Proposition~\protect\ref{prop:dk} ($\beta\bar\kappa=0.5$, $10^5$ steps).}\label{tab:dk}
\begin{tabular}{@{}cc ccc@{}}
\toprule
$L$ & $d$ & slope of $\Delta_n$ on $\mu_n$ & variance with $s_n$ & variance with $s^{\mathrm{ind}}_n$\\
\midrule
6 & 1 & 1.021 & 0.999 & 2.517\\
6 & 3 & 0.993 & 0.999 & 1.365\\
6 & 6 & 1.067 & 0.986 & 1.471\\
12 & 1 & 1.007 & 0.995 & 4.232\\
12 & 3 & 1.016 & 1.001 & 1.180\\
12 & 6 & 1.025 & 0.991 & 1.837\\
\bottomrule
\end{tabular}
\end{table}

The slope lies within 7\% of one, and the variance with $s_n$ within 1.5\% of one (Figure~\ref{fig:sim}a). With independent orders the variance lies between 1.18 and 4.23.

\subsection{The memory kernel}\label{sec:simkernel}
To measure the response $R$, we add $\Delta I=4\cdot10^{-4}$ to $I_{n_0}$ at $\beta\bar\kappa=0.1$ and compare the activity with that of the same run without this addition. All clusters share the target density, such that we use the measured $\bar\kappa$ for every cluster. By Proposition~\ref{prop:activity}, the activity responds linearly as long as no $p_k$ reaches one (Table~\ref{tab:checks}). The difference $D(s)$ of the two activities at $s=n-n_0$ then solves the renewal equation
\[
D(s)=\frac{\Delta I}{\bar\kappa}\,\phi(s)+\sum_{r=1}^{s-1}\phi(s-r)\,D(r),
\qquad
\phi(s)=\beta\bar\kappa\sum_{k}w_k\,\omega_k(s),
\]
such that $D=(\Delta I/\bar\kappa)R$. As both runs of a pair share their random numbers, the ratio of the measured to the predicted response is one up to the sampling error of 600 pairs.

\begin{table}[H]
\centering\small
\caption{Measured over predicted response, $D(s)\big/\bigl((\Delta I/\bar\kappa)R(s)\bigr)$, averaged over 600 pairs.}\label{tab:response}
\begin{tabular}{@{}l ccccccc@{}}
\toprule
$s$ & 1 & 4 & 17 & 55 & 173 & 545 & 1713\\
\midrule
ratio & 1.00 & 0.99 & 0.99 & 1.01 & 1.01 & 1.02 & 1.02\\
\bottomrule
\end{tabular}
\end{table}

\begin{figure}[H]
\centering
\includegraphics[width=0.75\textwidth]{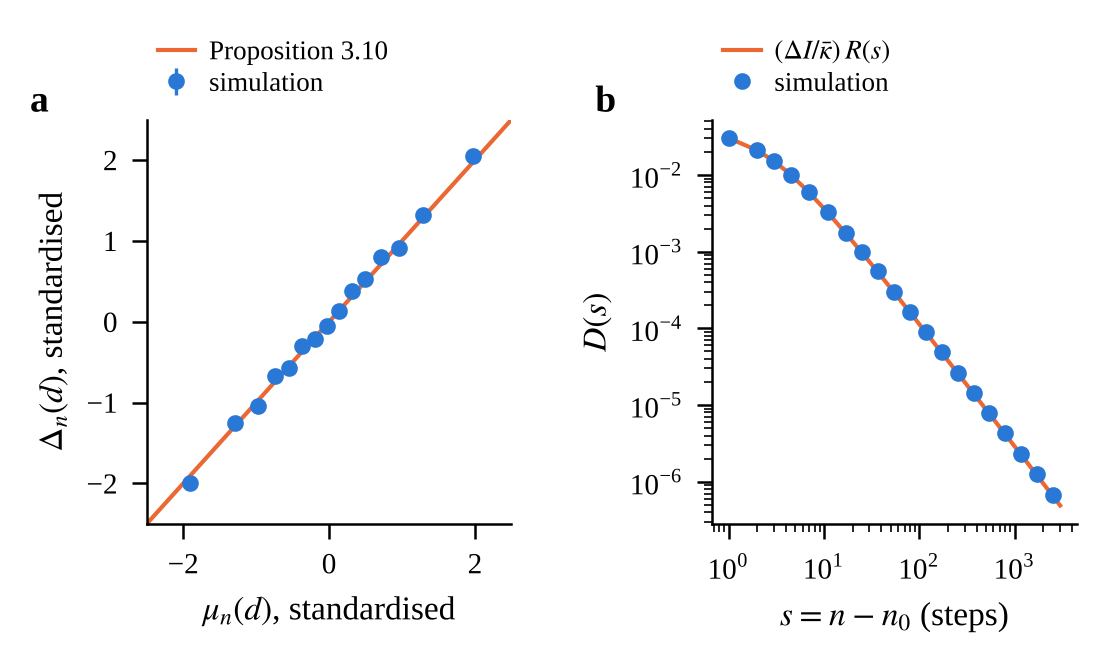}
\caption{Simulation against Proposition~\protect\ref{prop:dk} and Corollary~\protect\ref{cor:response}. (a) Change $\Delta_n(d)$ of the book against its conditional mean $\mu_n(d)$ of Proposition~\protect\ref{prop:dk}, both divided by the standard deviation of $\mu_n(d)$, in 15 bins of equal count over $d\in\{1,3,6\}$ ($L=6$, Table~\protect\ref{tab:dk}). Bars are standard errors. (b) Response $D(s)$ of the activity to $\Delta I=4\cdot10^{-4}$ added at step $n_0$, averaged over 600 pairs, against $(\Delta I/\bar\kappa)R(s)$ (Table~\protect\ref{tab:response}).}\label{fig:sim}
\end{figure}

The ratio stays within 2\% of one over three decades of lags (Figure~\ref{fig:sim}b). The horizons of Table~\ref{tab:hyper} grow by the factor $\Omega=1.49$. By Remark~\ref{rem:logperiodic}, the kernel then deviates from a power law by at most
\[
b(\Omega,\alpha)=\frac{2}{\mathcal I_\infty}\sum_{m\ge1}\lvert\mathcal J_m\rvert=\frac{2}{\Gamma(1+\alpha)}\sum_{m\ge1}\Bigl\lvert\Gamma\Bigl(1+\alpha+\frac{2\pi i m}{\ln\Omega}\Bigr)\Bigr\rvert ,
\]
which we evaluate numerically.

\begin{table}[H]
\centering\small
\caption{Bound $b(\Omega,\alpha)$ on the deviation of the kernel from a power law.}\label{tab:logperiodic}
\begin{tabular}{@{}cc cc@{}}
\toprule
$\Omega$ & clusters per decade & $\alpha=0.6$ & $\alpha=0.7$\\
\midrule
4 & 1.7 & $2.4\cdot10^{-2}$ & $2.8\cdot10^{-2}$\\
3 & 2.1 & $4.8\cdot10^{-3}$ & $5.7\cdot10^{-3}$\\
2 & 3.3 & $4.2\cdot10^{-5}$ & $5.1\cdot10^{-5}$\\
1.49 & 5.8 & $2.3\cdot10^{-9}$ & $3.0\cdot10^{-9}$\\
\bottomrule
\end{tabular}
\end{table}

For $\Omega=1.49$ the bound lies below $3.1\cdot10^{-9}$, such that for lags well inside the range of horizons the kernel is a power law to this accuracy.

\subsection{Roughness}\label{sec:simrough}
We compute two series over bins $b$ of 50 steps,
\[
A_b=\sum_{n\in b}\mathbb E[\mathcal A_n\mid\mathcal F_n],\qquad V_b=\sum_{n\in b}\operatorname{Var}[\theta_{n+1}\mid\mathcal F_n].
\]
We use the conditional mean of the activity because the actions of one step add noise without memory to the variogram. For each series $Y\in\{A,V\}$ we estimate the Hurst parameter $H$ from the variogram
\[
m_Y(\Delta)=\mathbb E\bigl[(\ln Y_{b+\Delta}-\ln Y_b)^2\bigr]\propto\Delta^{2H}
\]
over the lags of Table~\ref{tab:rough} (Appendix~\ref{app:estimators}). By Theorem~\ref{thm:rough}, the activity has $H=\alpha-\tfrac12$ as $\mathcal T\to\infty$. By the proof of Corollary~\ref{cor:mid}, $V_b$ equals $\sum_{n\in b}\operatorname{Var}[I_n\mid\mathcal F_n]$. We set
\[
\bar q_b=\frac{\sum_{n\in b}\operatorname{Var}[I_n\mid\mathcal F_n]}{\sum_{n\in b}\mathbb E[I_n\mid\mathcal F_n]} .
\]
With $\mathbb E[I_n\mid\mathcal F_n]=\bar\kappa\,\mathbb E[\mathcal A_n\mid\mathcal F_n]$ (Proposition~\ref{prop:selfexc}), we obtain
\[
\ln V_b=\ln\bar\kappa+\ln A_b+\ln\bar q_b .
\]
By Lemma~\ref{lem:condvarlimit}, $\bar q_b$ is close to the constant $\bar\nu$. Both series thus have the same variogram up to the variation of $\bar q_b$, and hence the same $H$.

\begin{table}[H]
\centering\small
\caption{Hurst parameters of the activity and of the squared volatility of the mid price (bins of 50 steps, Table~\protect\ref{tab:experiments}).}\label{tab:rough}
\begin{tabular}{@{}l ccc@{}}
\toprule
 & $\alpha=0.6$ & $\alpha=0.7$ & single cluster\\
\midrule
lags (steps) & $[10^3,\,5\cdot10^3]$ & $[10^3,\,5\cdot10^3]$ & $[300,\,2000]$\\
$H$ in the limit & $\alpha-\tfrac12=0.10$ & $\alpha-\tfrac12=0.20$ & $\tfrac12$\\
$H$ of $A_b$ & 0.152 & 0.217 & 0.473\\
$H$ of $V_b$ & 0.147 & 0.201 & 0.470\\
relative s.d.\ of $\bar q_b$ & 0.089 & 0.135 & 0.059\\
\bottomrule
\end{tabular}
\end{table}

The two estimates agree within 0.016. For $\alpha=0.7$ both equal $\alpha-\tfrac12$. For $\alpha=0.6$ both exceed it by 0.05 at our finite $\mathcal T$. With a single cluster the kernel has the finite mean $\sum_{s\ge1}s\,\omega(s)=T$. By \cite[Theorem~2.2]{JR15}, the activity then tends to the solution of $dX_t=a(\mu-X_t)\,dt+\sigma\sqrt{X_t}\,dB_t$, which has $H=\tfrac12$ by Lemma~\ref{lem:cir}. Both estimates lie within 0.03 of this limit.

\subsection{The role of the book}\label{sec:simbook}
The squared volatility of the mid price in Section~\ref{sec:simrough} is the conditional variance of one function of the book. For the changes of the book at the depths $d$ of Section~\ref{sec:dkcheck} we sum the same variance over each bin,
\[
S_b(d)=\sum_{n\in b}\ \sum_{i:\,\sigma_i=+1}p_k(n)\bigl(1-p_k(n)\bigr)\Bigl(m_i\sum_{j}\bigl[\varphi_d(x^{\mathrm{new}}_{i,j})-\varphi_d(x_{i,j})\bigr]\Bigr)^2,
\]
where $k$ is the cluster of desk $i$. With $r_b(d)=S_b(d)/A_b$ we obtain
\[
\ln S_b(d)=\ln A_b+\ln r_b(d).
\]
The book thus has the activity's $H$ when the variogram of $\ln r_b(d)$ is small against that of $\ln A_b$. By Lemma~\ref{lem:timescales}, the book recovers from a shift within about $1/p_k$ steps, while the memory of cluster $k$ lasts about $T_k$ steps. With the mean activity of 0.05 (Proposition~\ref{prop:longrun}), the product $T_kp_k$ lies between 50 and 250 for horizons in the range of lags of Table~\ref{tab:rough}. A single cluster with $T=200$ instead has $Tp_k\approx10$. The book therefore recovers its shape within the memory of the desks for 24 clusters but not for a single cluster.

\begin{table}[H]
\centering\small
\caption{Hurst parameters of the activity and of the changes of the book (bins of 50 steps, lags of Table~\protect\ref{tab:rough}). Each range runs over the depths $d\in\{1,3,6\}$ and, in the third row, also over the shortest and the longest lag.}\label{tab:booknoise}
\begin{tabular}{@{}l ccc@{}}
\toprule
 & $\alpha=0.6$ & $\alpha=0.7$ & single cluster\\
\midrule
$H$ of $A_b$ & 0.152 & 0.217 & 0.473\\
$H$ of $S_b(d)$ & 0.137--0.138 & 0.201 & 0.23--0.29\\
$m_{\ln r(d)}\,/\,m_{\ln A}$ & 0.05--0.08 & 0.03--0.04 & 0.5--4.1\\
correlation of $\ln S_b(d)$ and $\ln A_b$ & 0.98 & 0.99 & 0.93--0.96\\
\bottomrule
\end{tabular}
\end{table}

For 24 clusters the variogram of $\ln r_b(d)$ is at most 0.08 times that of $\ln A_b$. At every depth the book then has the activity's $H$ within 0.016. With a single cluster the ratio of the two variograms lies between 0.5 and 4.1. The $H$ of the book then falls to between 0.23 and 0.29.

The norm $\lVert\phi\rVert=\beta\bar\kappa$ of Proposition~\ref{prop:strength} depends on the book through $\bar\kappa=\sum_l\kappa_lw_l$. By the definition of $\kappa_l$ before Lemma~\ref{lem:size}, we have
\[
\bar\kappa=c\sum_i m_i\bigl\lvert\Delta Z_i(n)\bigr\rvert\approx\frac{c\gamma}{\ell}\sum_i m_i\Bigl\lvert\sum_j e^{-x_{i,j}/\ell}\bigl(x_{i,j}-x^*_{i,j}\bigr)\Bigr\rvert,
\]
where the second step keeps the first order in $\gamma$ and drops the trend $P_k$. Orders further from the mid price enter this sum with a smaller factor $e^{-x_{i,j}/\ell}$ but with larger gaps $\lvert x_{i,j}-x^*_{i,j}\rvert$ (Table~\ref{tab:shape}). The norm can therefore peak at an intermediate distance of the targets. We test this prediction at a fixed $\beta$ by multiplying every target by $s/1.5$. The density of Table~\ref{tab:hyper} then becomes $x\,e^{-x/s}/s^2$, which peaks at $x=s$.

\begin{table}[H]
\centering\small
\caption{Norm of the kernel as the targets move (24 clusters, $\alpha=0.6$, $\beta$ fixed at $\beta\bar\kappa=0.5$ for $s=1.5$). Here $\bar g=\sum_{i,j}m_i\lvert x_{i,j}-x^*_{i,j}\rvert/\sum_{i,j}m_i$ is the mean gap, and both $Z^++Z^-$ and $\bar g$ are averages along the run.}\label{tab:shape}
\begin{tabular}{@{}l ccccccc@{}}
\toprule
$s$ & 0.5 & 0.75 & 1 & 1.5 & 2 & 3 & 4\\
\midrule
$Z^++Z^-$ & 1.07 & 0.83 & 0.66 & 0.44 & 0.31 & 0.18 & 0.12\\
$\bar g$ & 0.04 & 0.07 & 0.09 & 0.13 & 0.15 & 0.19 & 0.22\\
$\beta\bar\kappa$ & 0.41 & 0.52 & 0.55 & 0.50 & 0.42 & 0.29 & 0.22\\
$\beta\bar\kappa$ / first-order value & 1.04 & 1.05 & 1.05 & 1.05 & 1.04 & 1.03 & 1.03\\
\bottomrule
\end{tabular}
\end{table}

The norm peaks at $s=1$ and falls on both sides. At every $s$ the first-order value lies 3--5\% below the measured norm. Between the tenths of each run $\bar\kappa$ varies by at most 0.23\%, which supports replacing $\kappa_l(n)$ by its average in Section~\ref{sec:response}.

\section{Evidence from Binance}\label{sec:binance}

We test our mechanism on a market whose book market makers change mainly by moving orders. The exchange Binance offers such a market in its perpetual futures of BTC, ETH and SOL against USDT. Trades there rarely reach 1\% from the mid price (Table~\ref{tab:conservation}). Market makers thus change the volume beyond 1\% mainly by placing and cancelling orders. Under price-time priority \cite{GPWMFH13}, an order that a market maker cancels and places again loses only its place in the queue. As queues are short at a small tick \cite{BBDG18}, this place is worth little at the tick of $10^{-6}$ to $10^{-4}$ of the price of these markets (Table~\ref{tab:conservation}). The public data of Binance \cite{Bin26} contain the taker orders of these perpetual futures and of the spot markets of the same coins. As the data contain no actions of market makers, we count the taker orders $N_b$ in each bin $b$ as the activity. In snapshot $n$ of a book, $V_n(d)$ denotes the volume resting between $d\%$ and $(d+1)\%$ from the mid price. The move of the mid price up to snapshot $n+1$ changes $V_n(d)$ by $\Delta^{\mathrm{mid}}_n(d)$ when every order keeps its price (Algorithm~\ref{alg:mid}). With the log return $r_i$ over minute $i$, we measure
\[
\mathrm{RV}_b=\sum_{i\in b}r_i^2,\qquad
\varsigma=\frac{\sum_{n,d}\Delta V_n(d)\,\Delta^{\mathrm{mid}}_n(d)}{\sum_{n,d}\Delta^{\mathrm{mid}}_n(d)^2},\qquad
Q_b=\frac{\sum_{n\in b}\sum_{d=1}^{4}e_n(d)^2}{\sum_{n\in b}\bigl(t_{n+1}-t_n\bigr)},
\]
with $\Delta V_n(d)=V_{n+1}(d)-V_n(d)$ and $e_n(d)=\Delta V_n(d)-\varsigma\,\Delta^{\mathrm{mid}}_n(d)$. A book whose orders all keep their prices has $\varsigma=1$. With the Hurst parameter $H(Y)$ of a series $Y$ (Appendix~\ref{app:estimators}), we then test
\[
\begin{aligned}
&H(N_b)<\tfrac12 &&\text{six markets} &&\text{Theorem~\ref{thm:rough}},\\
&\lVert\phi\rVert\approx1,\quad \phi(s)\propto s^{-1-\alpha} &&\text{six markets} &&\text{Section~\ref{sec:powerlaw}},\\
&H(\mathrm{RV}_b)=H(N_b) &&\text{six markets} &&\text{Corollary~\ref{cor:mid}},\\
&\operatorname{Corr}\bigl(\Delta V_n(d),\Delta V_n(d+1)\bigr)<0 &&\text{perpetual futures} &&\text{Proposition~\ref{prop:dk}},\\
&\varsigma<1 &&\text{perpetual futures} &&\text{Lemma~\ref{lem:timescales}},\\
&H(Q_b)=H(N_b) &&\text{perpetual futures} &&\text{Section~\ref{sec:simbook}}.
\end{aligned}
\]

\subsection{Data}\label{sec:bindata}
The relations above need the activity over time scales from one second to one day. Over the 180 days from 10 April to 6 October 2026, we therefore count per second the aggregated trades of each market. Binance merges into one aggregated trade the fills on one side at one price within 100\,ms \cite{Bin26}. As the shortest lag of 15 minutes needs several bins, $N_b$ counts these trades in bins $b$ of 5 minutes. The sampling noise of $n$ one-minute returns adds to the variogram of $\ln\mathrm{RV}_b$ a constant of $2\psi_1(n/2)$ for Gaussian returns, where $\psi_1$ is the trigamma function (Appendix~\ref{app:estimators}). Measured on these markets, this constant falls from 0.38--0.63 at bins of 15 minutes to 0.12--0.15 at bins of 60 minutes. We therefore compare $H(\mathrm{RV}_b)$ with $H(N_b)$ at bins of 15, 30 and 60 minutes. For the perpetual futures, Binance publishes a snapshot of the book about every 30 seconds with the cumulative volume within 0.2\% and within 1\% to 5\% of the mid price on each side \cite{Bin26}.

The activity of such markets is known empirically to follow a daily and a weekly pattern \cite{BCGL19,HKK24}. A pattern $s_b$ that repeats every day or every week adds to the variogram of $\ln N_b=s_b+y_b$ a second term,
\[
m_{\ln N}(\Delta)=m_{y}(\Delta)+\frac1B\sum_{b=1}^{B}\bigl(s_{b+\Delta}-s_b\bigr)^2 .
\]
This term grows with $\Delta$ up to half a day and raises the estimated $H$. Removing the pattern thus lowers $H$. As estimation error alone can produce $H<\tfrac12$ \cite{CD}, we also estimate $H(N_b)$ without any removal. It then lies between 0.10 and 0.14 (Table~\ref{tab:pattern}). As these markets trade around the clock, $s_b$ has no jump at an open or a close. With the bin $\tau_b\in\{0,\dots,287\}$ of the day and the weekday $\delta_b\in\{1,\dots,7\}$ of bin $b$, we therefore fit by least squares
\[
\ln N_b=\sum_{k=1}^{6}\Bigl(a_k\cos\frac{2\pi k\tau_b}{288}+c_k\sin\frac{2\pi k\tau_b}{288}\Bigr)+w_{\delta_b}+y_b .
\]
We estimate the Hurst parameter of the activity from the remainder $y_b$ (Appendix~\ref{app:estimators}).

\subsection{Activity and volatility}\label{sec:bintrades}
We estimate $H(N_b)$ from the variogram of $y_b$ over lags of 15 minutes to one day, in each of the six blocks of 30 days (Appendix~\ref{app:estimators}). The kernel of our desks in Section~\ref{sec:powerlaw} has $\omega_k(s)=T_k^{-1}(1-T_k^{-1})^{s-1}$, which becomes $e^{-s/T_k}/T_k$ for horizons $T_k$ in seconds. With 80 horizons $T_k=c\,(10^7\,\mathrm{s}/c)^{(k-1)/79}$, we fit to the spectrum of the counts per second over time scales of 1 minute to 4 hours (Appendix~\ref{app:estimators})
\[
\phi_{\mathrm{desk}}(s)=a\sum_{k=1}^{80}w_k\,\frac{e^{-s/T_k}}{T_k},\qquad w_k\propto T_k^{-\alpha},\qquad a=\beta\bar\kappa .
\]
We also fit $\phi_{\mathrm{pow}}(s)=a\,\alpha\,c^{\alpha}(s+c)^{-1-\alpha}$, which has the same norm and tail but another shape at time scales near $c$.

\begin{table}[H]
\centering\small
\caption{Activity and kernel of the six markets, 10 April to 6 October 2026. The gain is the log-likelihood of $\phi_{\mathrm{desk}}$ minus that of $\phi_{\mathrm{pow}}$. The last column gives $H$ of the desks of Table~\protect\ref{tab:reduced} at the fitted $\alpha$, for $T=10^5$ and $10^6$\,s.}\label{tab:binance}
\begin{tabular}{@{}l ccccc@{}}
\toprule
market & $H(N_b)$ & $\alpha$ & $a$, 95\% lower bound & gain & $H$ of the desks\\
\midrule
BTC spot & $0.083\pm0.004$ & $0.40\pm0.02$ & 0.998 & $-0.1$ & 0.050 / 0.056\\
BTC perp & $0.085\pm0.004$ & $0.42\pm0.03$ & 0.996 & 8.9 & 0.059 / 0.065\\
ETH spot & $0.084\pm0.003$ & $0.45\pm0.03$ & 0.979 & 9.5 & 0.069 / 0.073\\
ETH perp & $0.085\pm0.004$ & $0.48\pm0.02$ & 0.971 & 6.4 & 0.081 / 0.088\\
SOL spot & $0.080\pm0.003$ & $0.39\pm0.03$ & 0.971 & $-0.5$ & 0.047 / 0.054\\
SOL perp & $0.071\pm0.006$ & $0.50\pm0.04$ & 0.945 & $-0.0$ & 0.088 / 0.096\\
\bottomrule
\end{tabular}
\end{table}

\begin{figure}[H]
\centering
\includegraphics[width=0.75\textwidth]{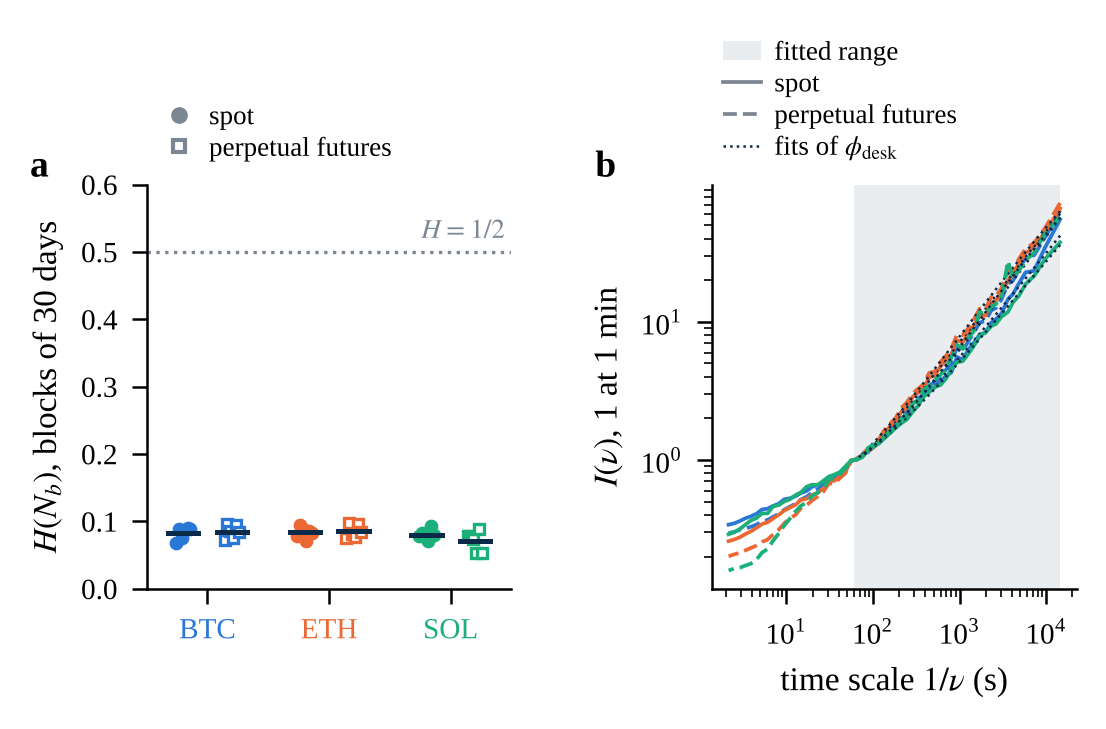}
\caption{Activity and kernel of the six markets, 10 April to 6 October 2026. (a) $H(N_b)$ in each block of 30 days, with the mean over the blocks. (b) Periodogram $I(\nu)$ of the counts per second, scaled to 1 at 1 minute, with the fits of $\phi_{\mathrm{desk}}$ over the shaded range (Appendix~\protect\ref{app:spectrum}).}\label{fig:binance}
\end{figure}

In every market $H(N_b)$ lies between 0.07 and 0.09 (Figure~\ref{fig:binance}a). The 95\% lower bound of the norm $a$ is at least 0.94. The order flow thus lies close to the threshold of Theorem~\ref{thm:rough}. The fitted tails lie between 0.39 and 0.50, where the limit $H=\alpha-\tfrac12$ of that theorem is not positive. At distances $1-a=T^{-\alpha}$ with $T$ of $10^5$ or $10^6$\,s, simulated desks with these tails give values of $H$ that the data exceed by 0.016 and 0.009 on average, with 95\% intervals of $[-0.005,\,0.036]$ and $[-0.012,\,0.030]$ (Table~\ref{tab:binance}). At $T=10^4$\,s the desks fall short of the data by 0.039 $[0.023,\,0.055]$. The data thus agree with desks whose memory reaches beyond the longest lag of one day. As both kernels share their tail, the gain of $\phi_{\mathrm{desk}}$ in three markets reflects its shape near $c$. A similar phenomenon has been observed in the term structure of the at-the-money skew of equity indices \cite{GE23}. There, a power law fits beyond short maturities, while shapes that resemble it but stay finite fit better at short ones.

By Corollary~\ref{cor:mid} the squared volatility of the mid price has the Hurst parameter of the activity. We estimate this squared volatility by $\mathrm{RV}_b$, whose sampling noise adds a constant $\eta$ to the variogram of $\ln\mathrm{RV}_b$ (Section~\ref{sec:bindata}). To measure $\eta$, we split the $n$ minutes of each bin into its odd and its even minutes,
\[
\mathrm{RV}^{\mathrm{odd}}_b=\sum_{i\in b,\ i\ \mathrm{odd}}r_i^2,\qquad \mathrm{RV}^{\mathrm{even}}_b=\sum_{i\in b,\ i\ \mathrm{even}}r_i^2 .
\]
As both halves share the volatility of the bin, the difference of their logarithms contains only sampling noise. With the trigamma function $\psi_1$ of Section~\ref{sec:bindata}, the variance of this difference gives
\[
\hat\eta=2\,\psi_1\bigl(\tfrac n2\bigr)\,\frac{\operatorname{Var}_b\bigl[\ln\mathrm{RV}^{\mathrm{odd}}_b-\ln\mathrm{RV}^{\mathrm{even}}_b\bigr]}{\psi_1\bigl(\lceil n/2\rceil/2\bigr)+\psi_1\bigl(\lfloor n/2\rfloor/2\bigr)} .
\]
We subtract $\hat\eta$ from the variogram of $\ln\mathrm{RV}_b$ and estimate $H(\mathrm{RV}_b)$ (Appendix~\ref{app:estimators}).

\begin{table}[H]
\centering\small
\caption{Hurst parameters of the activity and of the realised variance in the six markets, 10 April to 6 October 2026, at lags of 1 hour to 1 day. Here $N_b$ counts the trades of the one-minute candles, and the pattern removed is the mean of each bin of the day and of each weekday.}\label{tab:rv}
\footnotesize
\begin{tabular}{@{}l ccccc@{}}
\toprule
bin & $\hat\eta$ & $H(N_b)$ & $H(\mathrm{RV}_b)$ & largest $\lvert H(\mathrm{RV}_b)-H(N_b)\rvert$ & correlation of $\ln N_b$, $\ln\mathrm{RV}_b$\\
\midrule
15 min & 0.38--0.63 & 0.078--0.088 & 0.080--0.091 & 0.013 & 0.83--0.90\\
30 min & 0.21--0.28 & 0.094--0.107 & 0.090--0.103 & 0.011 & 0.87--0.93\\
60 min & 0.12--0.15 & 0.123--0.136 & 0.113--0.130 & 0.018 & 0.88--0.95\\
\bottomrule
\end{tabular}
\end{table}

At every bin length $H(\mathrm{RV}_b)$ lies within 0.018 of $H(N_b)$ (Table~\ref{tab:rv}). The logarithms of the two series have correlations of 0.83 to 0.95. From bins of 15 to 60 minutes $\hat\eta$ falls by a factor of four without changing this agreement. Averaging within a bin raises the fitted $H$ of a rough series. A log-intensity with $H=0.08$ at one-minute resolution gives 0.16, 0.19 and 0.23 at bins of 15, 30 and 60 minutes (Table~\ref{tab:binning}).

\subsection{The order book}\label{sec:binbook}
As trades almost never reach 1\% from the mid price (Table~\ref{tab:conservation}), we measure the books of the perpetual futures between 1\% and 5\%. A move of the mid price changes the distance of every order that keeps its price. By Lemma~\ref{lem:timescales}, market makers move their orders back towards their distances. Let the mid price move at a constant speed $v$ over an interval of length $u$. Let $\delta(t)$ denote the displacement of an order from its distance at the start of the interval. If market makers move the order back at the rate $1/\tau$, then
\[
\frac{d\delta}{dt}=v-\frac{\delta}{\tau},\qquad \delta(0)=0\qquad\Longrightarrow\qquad \delta(u)=v\,\tau\,\bigl(1-e^{-u/\tau}\bigr).
\]
As $\Delta V_n(d)$ is linear in the displacements of the orders for small moves, the slope $\varsigma$ equals the ratio of $\delta(u)$ to the displacement $vu$ of an order that keeps its price,
\[
\varsigma=\frac{\delta(u)}{v\,u}=\frac{1-e^{-u/\tau}}{u/\tau}.
\]
With $u$ the median time between snapshots, we solve this relation for the time $\tau$ of re-centring.

Market makers change the volume $e_n(d)=\Delta V_n(d)-\varsigma\,\Delta^{\mathrm{mid}}_n(d)$ that remains after the move of the mid price by placing, cancelling and moving orders. A moved order leaves one interval for its neighbour and thus changes the two volumes in opposite directions, whereas orders placed and cancelled independently leave them uncorrelated. As the size of the changes varies strongly over a day, we correlate them within blocks $B$ of 10 consecutive pairs of snapshots. Averaging over the three neighbouring pairs of the four intervals and taking the median over blocks and both sides of the book, we obtain
\[
\rho=\operatorname*{median}_{B,\,\pm}\ \frac13\sum_{d=1}^{3}\operatorname{Corr}_B\bigl(e_n(d),\,e_n(d+1)\bigr).
\]
This statistic is zero for independent orders and negative for moved orders. Simulated books of our desks, passed through the same four intervals, give its value in our model (Table~\ref{tab:modelbook}). On pairs of snapshots whose move $\mu_n$ of the mid price stays below 0.02\% (Algorithm~\ref{alg:mid}), $\rho$ needs no estimate of $\varsigma$.

By Section~\ref{sec:simbook}, the changes of the book have the Hurst parameter of the activity when the book recovers its shape within the memory of the desks. The sum of the squared remainders $e_n(d)$ over a bin estimates the conditional variance $S_b(d)$ of Section~\ref{sec:simbook}. As the time between snapshots varies, we divide this sum by the time that it covers, which gives $Q_b$. The odd and the even pairs of snapshots in a bin give two such ratios $Q^{\mathrm{odd}}_b$ and $Q^{\mathrm{even}}_b$ with independent sampling noise. As for $\mathrm{RV}_b$, half the variance of $\ln Q^{\mathrm{odd}}_b-\ln Q^{\mathrm{even}}_b$ estimates this noise in the variogram of $\ln Q_b$ (Appendix~\ref{app:estimators}). We subtract it and remove the pattern of Section~\ref{sec:bindata} before we estimate $H(Q_b)$ in bins of 5 minutes over lags of 15 minutes to one day. The time $\tau$ of re-centring and the memory of the activity then decide whether Section~\ref{sec:simbook} predicts $H(Q_b)=H(N_b)$ for these markets.

\begin{table}[H]
\centering\small
\caption{The books of the perpetual futures, 7 September to 6 October 2026. Snapshots come every $u=30$\,s, and the mid price is taken from the one-second candles of the spot market. $H$ is estimated at bins of 5 minutes over lags of 15 minutes to one day.}\label{tab:book}
\begin{tabular}{@{}l ccccccc@{}}
\toprule
market & $\varsigma$ & $\tau$ (s) & $\rho$, all pairs & $\rho$, $\lvert\mu_n\rvert<0.02\%$ & $H(Q_b)$ & $H(N_b)$ & corr.\ of $\ln Q_b$, $\ln N_b$\\
\midrule
BTC perp & 0.37 & 12 & $-0.30$ & $-0.27$ & 0.050 & 0.078 & 0.56\\
ETH perp & 0.51 & 20 & $-0.21$ & $-0.16$ & 0.037 & 0.079 & 0.46\\
SOL perp & 0.53 & 21 & $-0.28$ & $-0.28$ & 0.054 & 0.046 & 0.45\\
\bottomrule
\end{tabular}
\end{table}

In all three books $\varsigma$ lies between 0.37 and 0.53, such that market makers move their orders back within 12 to 21 seconds (Table~\ref{tab:book}). The orders that they move change neighbouring intervals in opposite directions, with $\rho$ between $-0.30$ and $-0.16$ on all pairs and on quiet pairs. This range contains the values $-0.18$ to $-0.17$ of the books of our desks (Table~\ref{tab:modelbook}). As the books recover within seconds and the activity remembers beyond one day (Section~\ref{sec:bintrades}), Section~\ref{sec:simbook} predicts $H(Q_b)=H(N_b)$. SOL meets this equality within 0.008, whereas for BTC and ETH $H(Q_b)$ lies 0.03 and 0.04 below $H(N_b)$. The size of the changes of these books also follows the activity less closely than in our model, with correlations of 0.45 to 0.56 between the logarithms against 0.98 and 0.99 in Table~\ref{tab:booknoise}.

\section{Conclusion}\label{sec:conclusion}

The rough volatility observed in many markets \cite{BLP22,GJR18,LMPR18,Tak20} has so far been explained by traders who split large orders or by the absence of arbitrage \cite{EFR18,JuR20}. Such explanations leave out the market makers who provide a large share of the liquidity in electronic markets \cite{HS09,Men13}. We therefore studied desks of market makers that move their orders by gradient steps at times set by new information over their own horizons. The volume of their book follows a conservative stochastic partial differential equation whose fluctuations make the order flow self-exciting with a kernel $\phi$. The shape of the book sets the norm of $\phi$, such that close to the threshold the activity and the squared volatility of the mid price become rough with $H=\alpha-\tfrac12$. For the book itself the mechanism predicts $\varsigma<1$, $\rho<0$ and $H(Q_b)=H(N_b)$ (Section~\ref{sec:binance}).

Simulations over a finite time reproduce each of these steps and give roughness only when the desks sample information over distinct horizons (Section~\ref{sec:simulation}). On six markets of Binance the order flow and the volatility are rough with $H$ between 0.07 and 0.09 and within 0.02 of each other. In the books of these markets, market makers move orders back within 12 to 21 seconds and move volume between neighbouring depths. These data agree with our desks at a finite distance from the threshold (Section~\ref{sec:bintrades}).

We leave extensions such as a limit theorem for tails $\alpha\le\tfrac12$, tests on data of single orders and markets with lower liquidity to future work.

\section*{Data and code availability}
Code and processed data are available at \url{https://github.com/rsn870/SGD_lob_rough_vol}.
Raw Binance data are public at \url{https://data.binance.vision}.

\bibliographystyle{alpha}
\bibliography{references}

\appendix
\renewcommand{\thetable}{\thesection.\arabic{table}}

\section{Estimators}\label{app:estimators}
\setcounter{table}{0}

\subsection{Hurst parameter}\label{app:hurst}
Every Hurst parameter in Sections~\ref{sec:simulation} and~\ref{sec:binance} comes from the variogram of a logarithm. For a positive series $Y_b$ with $b=1,\dots,B$, let $y_b$ denote $\ln Y_b$ after the removal of the pattern of Section~\ref{sec:bindata}. In Section~\ref{sec:simulation} the simulated series have no pattern, and $y_b=\ln Y_b$. With the noise constant $\eta$ of the series and a geometric grid $L$ of lags, we compute
\[
\begin{aligned}
m(\Delta)&=\frac1{B-\Delta}\sum_{b=1}^{B-\Delta}\bigl(y_{b+\Delta}-y_b\bigr)^2,\qquad \Delta\in L,\\
(\hat H,\hat c)&=\operatorname*{arg\,min}_{H,\,c}\ \sum_{\Delta\in L}\Bigl(\ln\bigl(m(\Delta)-\eta\bigr)-\ln c-2H\ln\Delta\Bigr)^2 .
\end{aligned}
\]
Each series has its own noise constant,
\[
\eta_N=2\,\overline{N_b^{-1}},\qquad \eta_{\mathrm{RV}}=\hat\eta\ \ \text{(Section~\ref{sec:bintrades})},\qquad \eta_Q=\tfrac12\operatorname{Var}_b\bigl[\ln Q^{\mathrm{odd}}_b-\ln Q^{\mathrm{even}}_b\bigr],
\]
and $\eta=0$ for the simulated series of Section~\ref{sec:simulation}. The constant $\eta_N$ follows from Poisson counts with intensity $\lambda_b$ in bin $b$,
\[
\begin{aligned}
\ln N_b&=\ln\lambda_b+\frac{N_b-\lambda_b}{\lambda_b}+O\bigl(\lambda_b^{-1}\bigr),\qquad \operatorname{Var}\bigl[\ln N_b\mid\lambda_b\bigr]=\frac1{\lambda_b}+O\bigl(\lambda_b^{-2}\bigr),\\
\mathbb E\bigl[(y_{b+\Delta}-y_b)^2\bigr]&=m_{\ln\lambda}(\Delta)+\mathbb E\Bigl[\frac1{\lambda_{b+\Delta}}+\frac1{\lambda_b}\Bigr]+O\bigl(\lambda^{-2}\bigr)\approx m_{\ln\lambda}(\Delta)+2\,\overline{N_b^{-1}} .
\end{aligned}
\]
For Table~\ref{tab:binance} we estimate $\hat H$ in each of the six blocks of 30 days and give the mean with the standard error over blocks.

\begin{table}[H]
\centering\small
\caption{Settings of the estimates of $H$.}\label{tab:hsettings}
\footnotesize
\begin{tabular}{@{}p{3.3cm} p{1.6cm} p{3.4cm} p{1.1cm} p{2.6cm}@{}}
\toprule
series & bins & lags $L$ & $\eta$ & pattern\\
\midrule
$A_b$, $V_b$, $S_b(d)$ (Sec.~\ref{sec:simulation}) & 50 steps & 8 geometric, $[10^3,5\cdot10^3]$ or $[300,2000]$ steps & 0 & none\\
$N_b$ (Table~\ref{tab:binance}) & 5 min & geometric, 15 min to 1 day & $\eta_N$ & six terms and weekdays\\
$N_b$, $\mathrm{RV}_b$ (Table~\ref{tab:rv}) & 15, 30, 60 min & 8 geometric, 1 h to 1 day & $\eta_N$, $\hat\eta$ & mean per bin of the day and weekdays\\
$Q_b$ (Table~\ref{tab:book}) & 5 min & geometric, 15 min to 1 day & $\eta_Q$ & six terms and weekdays\\
\bottomrule
\end{tabular}
\end{table}

\subsection{Noise of the realised variance}\label{app:rvnoise}

\begin{lemma}[Noise of the realised variance]\label{lem:rvnoise}
Let $r_{b,i}=\sigma_b\,\varepsilon_{b,i}$ for $i=1,\dots,n$, with $\varepsilon_{b,i}$ i.i.d.\ $N(0,1)$ over $b$ and $i$ and independent of $(\sigma_b)$, and let $m_Y(\Delta)=\mathbb E\bigl[(Y_{b+\Delta}-Y_b)^2\bigr]$. Then for every $\Delta\ge1$,
\[
m_{\ln\mathrm{RV}}(\Delta)=m_{\ln\sigma^2}(\Delta)+2\,\psi_1\bigl(\tfrac n2\bigr).
\]
\end{lemma}

The noise of $\ln\mathrm{RV}_b$ thus adds the same constant at every lag. The idea of the proof is that $\ln\mathrm{RV}_b$ splits into $\ln\sigma_b^2$ and the logarithm of a Gamma variable that is independent across bins. The variance of that logarithm is the constant.

\begin{proof}[\textbf{Proof of Lemma~\ref{lem:rvnoise}}]
\emph{\textbf{The noise as a Gamma variable.}} To separate the noise from $\sigma_b$, we factor $\sigma_b^2$ out of $\mathrm{RV}_b$. As the $\varepsilon_{b,i}$ are i.i.d.\ $N(0,1)$, their squares sum to a $\chi^2_n$ variable, which is twice a $\Gamma(\tfrac n2,1)$ variable $G_b$,
\[
\mathrm{RV}_b=\sigma_b^2\sum_{i=1}^{n}\varepsilon_{b,i}^2=2\,\sigma_b^2\,G_b,
\qquad
\ln\mathrm{RV}_b=\ln2+\ln\sigma_b^2+\ln G_b .
\]
Different bins share no returns, so the $G_b$ are independent over $b$. They are also independent of $(\sigma_b)$, as the $\varepsilon_{b,i}$ are.

\emph{\textbf{The variance of $\ln G_b$.}} The constant in the lemma is twice this variance, which we obtain from the cumulant generating function of $\ln G_b$. For $t>-\tfrac n2$, the Gamma density gives $\mathbb E[G_b^{\,t}]=\Gamma(\tfrac n2+t)/\Gamma(\tfrac n2)$. Taking logarithms and differentiating twice at $t=0$, with $\psi_1=(\ln\Gamma)''$, we obtain
\[
K(t)=\ln\mathbb E\bigl[e^{t\ln G_b}\bigr]=\ln\Gamma\bigl(\tfrac n2+t\bigr)-\ln\Gamma\bigl(\tfrac n2\bigr),
\qquad
\operatorname{Var}[\ln G_b]=K''(0)=\psi_1\bigl(\tfrac n2\bigr).
\]

\emph{\textbf{The variogram.}} Let $\eta_b=\ln G_b-\mathbb E[\ln G_b]$. In the difference over a lag $\Delta\ge1$, the constants $\ln2$ and $\mathbb E[\ln G_b]$ cancel, which leaves
\[
\ln\mathrm{RV}_{b+\Delta}-\ln\mathrm{RV}_b=\bigl(\ln\sigma^2_{b+\Delta}-\ln\sigma^2_b\bigr)+\bigl(\eta_{b+\Delta}-\eta_b\bigr).
\]
We square and take expectations. The cross term vanishes, as $(\eta_b)$ has mean zero and is independent of $(\sigma_b)$. The bins $b$ and $b+\Delta$ differ, so $\eta_{b+\Delta}$ and $\eta_b$ are independent, and each has variance $\psi_1(\tfrac n2)$. Hence
\[
\begin{aligned}
m_{\ln\mathrm{RV}}(\Delta)
&=m_{\ln\sigma^2}(\Delta)+2\,\mathbb E\bigl[(\ln\sigma^2_{b+\Delta}-\ln\sigma^2_b)(\eta_{b+\Delta}-\eta_b)\bigr]+\mathbb E\bigl[(\eta_{b+\Delta}-\eta_b)^2\bigr]\\
&=m_{\ln\sigma^2}(\Delta)+0+2\,\psi_1\bigl(\tfrac n2\bigr).
\end{aligned}
\]
\end{proof}

Returns with heavier tails than Gaussian raise $\operatorname{Var}[\ln(\mathrm{RV}_b/\sigma_b^2)]$ above $\psi_1(n/2)$. The odd and the even minutes of a bin share $\sigma_b$, so the variance of their difference measures this rise,
\[
\hat r=\frac{\operatorname{Var}_b\bigl[\ln\mathrm{RV}^{\mathrm{odd}}_b-\ln\mathrm{RV}^{\mathrm{even}}_b\bigr]}{\psi_1\bigl(\lceil n/2\rceil/2\bigr)+\psi_1\bigl(\lfloor n/2\rfloor/2\bigr)},\qquad \hat\eta=2\,\psi_1\bigl(\tfrac n2\bigr)\,\hat r .
\]
This step assumes that the tails raise the noise of $n$ returns and of $n/2$ returns by the same factor. A log-volatility with $H=0.105$ drives both series in a synthetic test (Table~\ref{tab:synthetic}). With Gaussian returns $\hat r=1.02$, and the corrected $H$ is 0.105. With Student $t_3$ returns $\hat r=2.61$, and the corrected $H$ is 0.065. On the six markets $\hat r$ lies between 1.3 and 2.2.

\subsection{Effect of the estimators on \texorpdfstring{$H$}{H}}\label{app:effects}
Three steps of our estimators change the fitted $H$: the removal of the pattern, the averaging within bins and the correction for noise. Removing the pattern from $\ln N_b$ lowers $H(N_b)$ by 0.03 to 0.05 (Table~\ref{tab:pattern}). Terms of the daily pattern with periods shorter than 4 hours change it by at most 0.005, and the counting noise $\eta_N$ by at most 0.001.

\begin{table}[H]
\centering\small
\caption{$H(N_b)$ for the 180 days of Table~\protect\ref{tab:binance}, at bins of 5 minutes and lags of 15 minutes to one day, over the whole period.}\label{tab:pattern}
\begin{tabular}{@{}l cccccc@{}}
\toprule
pattern removed & BTC perp & ETH perp & SOL perp & BTC spot & ETH spot & SOL spot\\
\midrule
none & 0.137 & 0.131 & 0.099 & 0.128 & 0.126 & 0.113\\
$w_\delta$ only & 0.121 & 0.119 & 0.096 & 0.116 & 0.114 & 0.109\\
$k\le6$ and $w_\delta$ (Section~\ref{sec:bindata}) & 0.086 & 0.087 & 0.072 & 0.084 & 0.085 & 0.083\\
$k\le24$ and $w_\delta$ & 0.089 & 0.089 & 0.077 & 0.086 & 0.087 & 0.084\\
$k\le6$ and $w_\delta$, $\eta_N=0$ & 0.086 & 0.087 & 0.071 & 0.084 & 0.085 & 0.082\\
\bottomrule
\end{tabular}
\end{table}

To measure the effect of the bins, we give a log-intensity with known $H$ one value per minute over 180 days and sum it in bins of 15, 30 and 60 minutes. Its fitted $H$ rises with the bin length for every true $H$ (Table~\ref{tab:binning}).

\begin{table}[H]
\centering\small
\caption{Fitted $H$ of a log-intensity $2X_t$ with $X$ a fractional Brownian motion of known $H$, at lags of 1 hour to 1 day.}\label{tab:binning}
\begin{tabular}{@{}c cccc@{}}
\toprule
true $H$ & 1 min & 15 min & 30 min & 60 min\\
\midrule
0.05 & 0.052 & 0.135 & 0.165 & 0.215\\
0.08 & 0.082 & 0.159 & 0.186 & 0.233\\
0.10 & 0.096 & 0.166 & 0.191 & 0.235\\
\bottomrule
\end{tabular}
\end{table}

For the correction $\hat\eta$ of Section~\ref{sec:bintrades}, one log-volatility drives both the counts and the returns of a synthetic market (Table~\ref{tab:synthetic}). With Gaussian returns $\hat\eta$ restores the $H$ of the counts. With Student $t_3$ returns it restores 0.065 of 0.105.

\begin{table}[H]
\centering\small
\caption{Synthetic market over 60 days of minutes. The log-variance $\ln(\sigma_t^2/10^{-6})$ is a sum of 20 Ornstein--Uhlenbeck processes with time scales from 10 to $10^5$ minutes, scaled to a standard deviation of 0.6. Counts are Poisson with mean $200\,\sigma_t^2/10^{-6}$, and returns are $\sigma_t\varepsilon_t$. Bins of 15 minutes, lags of 1 hour to 1 day.}\label{tab:synthetic}
\begin{tabular}{@{}l ccccc@{}}
\toprule
returns $\varepsilon_t$ & $H(N_b)$ & $H(\mathrm{RV}_b)$, raw & with $2\psi_1(7.5)$ & with $\hat\eta$ & $\hat r$\\
\midrule
Gaussian & 0.105 & 0.059 & 0.103 & 0.105 & 1.02\\
Student $t_3$ & 0.105 & 0.031 & 0.039 & 0.065 & 2.61\\
\bottomrule
\end{tabular}
\end{table}

\subsection{Kernel from the spectrum}\label{app:spectrum}
Table~\ref{tab:binance} needs the norm $a$ of the kernel, its lower bound and the tail $\alpha$. As the spectrum of a self-exciting process with mean rate $\bar\lambda$ and kernel $a\,g$ of norm $\lVert g\rVert=1$ is known in closed form \cite{Haw71}, we fit the kernel to the spectrum of the counts per second,
\[
\begin{aligned}
S(\omega)&=\frac{\bar\lambda}{\bigl\lvert1-a\,\hat g(\omega)\bigr\rvert^{2}},\\
\hat g_{\mathrm{desk}}(\omega)&=\sum_{k=1}^{80}\frac{w_k}{1+i\,\omega T_k},\qquad
\hat g_{\mathrm{pow}}(\omega)=\int_0^\infty \alpha\,c^{\alpha}(s+c)^{-1-\alpha}e^{-i\omega s}\,ds .
\end{aligned}
\]
Counts in bins of one second cannot tell the frequency $\nu$ from $\nu+k$ for integer $k$, and they weight each frequency by $\operatorname{sinc}^2$. Relative to a Poisson process of the same rate, their spectrum is
\[
f(\nu)=w+A\Bigl[\sum_{k=-3}^{3}\operatorname{sinc}^2(\nu+k)\Bigl(\frac{S(2\pi\lvert\nu+k\rvert)}{\bar\lambda}-1\Bigr)+1\Bigr],
\qquad \operatorname{sinc}x=\frac{\sin\pi x}{\pi x},
\]
where $w$ adds excitation within one second and $A$ scales for the merging of fills (Section~\ref{sec:bindata}). As the rate of trading changes over a day, we compare $f$ with the periodogram of segments $s$ of 8 hours, each divided by its own mean $\bar x_s$,
\[
I(\nu)=\operatorname*{mean}_{s}\ \frac{\bigl\lvert\sum_{t}(x_{s,t}-\bar x_s)\,e^{-2\pi i\nu t}\bigr\rvert^{2}}{28\,800\ \bar x_s}.
\]
For frequencies $\nu$ between $1/(4\,\mathrm{h})$ and $1/(1\,\mathrm{min})$, the Whittle approximation of the negative log-likelihood is $\sum_\nu\bigl(\ln f(\nu)+I(\nu)/f(\nu)\bigr)$. To bound the norm from below, we minimise it over the other parameters at each fixed $a$,
\[
\begin{aligned}
\ell(a)&=\min_{\alpha,\,c,\,A,\,w}\ \sum_{\nu}\Bigl(\ln f(\nu)+\frac{I(\nu)}{f(\nu)}\Bigr),\\
\hat a&=\operatorname*{arg\,min}_a\ell(a),
\qquad a_{95}=\min\bigl\{a:\ \ell(a)-\ell(\hat a)\le1.92\bigr\},
\end{aligned}
\]
with $1.92$ half the 95\% quantile of $\chi^2_1$. The gain in Table~\ref{tab:binance} is $\ell_{\mathrm{pow}}(\hat a)-\ell_{\mathrm{desk}}(\hat a)$. For the standard error of $\alpha$ we repeat the fit on each third of the period.

\begin{table}[H]
\centering\small
\caption{Settings of the fit of the kernel.}\label{tab:whittle}
\begin{tabular}{@{}ll@{}}
\toprule
setting & value\\
\midrule
segment length & 8\,h; segments with mean below 0.2 per second left out\\
frequencies & $1/(4\,\mathrm{h})$ to $1/(1\,\mathrm{min})$, summed in 40 bins evenly spaced in $\ln\nu$\\
aliases $k$ & $-3,\dots,3$\\
norm $a$ & 50 values with $1-a$ evenly spaced in $\ln(1-a)$ on $[10^{-7},\,0.5]$\\
tail $\alpha$ & $[0.05,\,1.2]$\\
$\phi_{\mathrm{desk}}$ & 80 horizons from $c$ to $10^7$\,s\\
$\hat g_{\mathrm{pow}}$ & quadrature on 150 points in $\ln s$\\
minimiser & Nelder--Mead; three starts at the first $a$, then the previous optimum\\
\bottomrule
\end{tabular}
\end{table}

\section{Experiments}\label{app:experiments}
\setcounter{table}{0}
\renewcommand{\thealgorithm}{\thesection.\arabic{algorithm}}

\subsection{Settings of the simulations}\label{app:settings}

\begin{table}[H]
\centering\small
\caption{Settings shared by all simulations.}\label{tab:hyper}
\begin{tabular}{@{}ll@{}}
\toprule
setting & value\\
\midrule
book and mid price & $c=0.3$, $\ell=1.5$\\
desks & $\gamma=0.1$, $\Pi=1$, $K=24$, $T_k=2\cdot1.49^{k-1}$, $w_k\propto T_k^{-\alpha}$\\
desks per cluster and side, orders per desk & $N=4$, $L=6$\\
target density & $x\,e^{-x/1.5}/1.5^{2}$\\
\bottomrule
\end{tabular}
\end{table}

\begin{table}[H]
\centering\footnotesize
\caption{One experiment per column.}\label{tab:experiments}
\begin{tabular}{@{}l p{2.2cm} p{2.1cm} p{1.6cm} p{1.4cm} p{2.2cm} p{2.2cm}@{}}
\toprule
 & Dean--Kawasaki (\ref{sec:dkcheck}) & Response (\ref{sec:simkernel}) & Roughness (\ref{sec:simrough}) & Single cluster (\ref{sec:simrough}) & Mid price (\ref{sec:simrough}), book (\ref{sec:simbook}) & Shape of the book (\ref{sec:simbook})\\
\midrule
horizons & $T_k$ & $T_k$ & $T_k$ & $T=200$ & both & $T_k$\\
$\alpha$ & 0.6 & 0.6 & 0.6, 0.7 & --- & 0.6, 0.7, --- & 0.6\\
$\beta\bar\kappa$ & 0.5 & 0.1 & $1-5000^{-\alpha}$ & 0.987 & $1-5000^{-\alpha}$, 0.987 & 0.5 at $s=1.5$, then $\beta$ fixed\\
$p_0$ & 0.02 & 0.02 & $0.05(1-\beta\bar\kappa)$ & same & same & 0.02\\
steps & $10^5$ & $600\times3000$ & $2\cdot10^7$ & $2\cdot10^7$ & $10^7$ & $3\cdot10^5+2\cdot10^6$ per $s$\\
measured & $V_n(d)$, $d\in\{1,3,6\}$ & $\mathcal A_n$ after $I_{n_0}\mapsto I_{n_0}+4\cdot10^{-4}$, against the same run without & $\mathcal A_n$, bins of 20 & bins of 20 & $\operatorname{Var}[\theta_{n+1}\mid\mathcal F_n]$ and $S_b(d)$, bins of 50 & $\bar\kappa$, $Z^\pm$ and $\bar g$ for $s\in\{0.5,\dots,4\}$\\
lags & --- & $1\le n-n_0\le3000$ & $[10^3,5\cdot10^3]$ & $[300,2000]$ & both ranges & ---\\
\bottomrule
\end{tabular}
\end{table}

In every experiment we obtain $\beta$ by dividing $\beta\bar\kappa$ in Table~\ref{tab:experiments} by the measured $\bar\kappa$.

\subsection{Assumption~\protect\ref{ass:interior} and the constant \texorpdfstring{$\bar\kappa$}{kappa} along the runs}\label{app:checks}
The linear response of Section~\ref{sec:simkernel} and the limits of Section~\ref{sec:rough} rest on Assumption~\ref{ass:interior} and on a constant $\bar\kappa$ (Section~\ref{sec:response}). In the notation of Section~\ref{sec:setting}, these three conditions read
\[
\max_{k}p_k(n)<1,\qquad \min_{i,j}\bigl(x_{i,j}(n)-\sigma_i\theta_n\bigr)>0,\qquad \kappa_l(n)=\kappa_l .
\]
We measure each condition along every run of Sections~\ref{sec:dkcheck} to~\ref{sec:simbook}.

\begin{table}[H]
\centering\footnotesize
\caption{The conditions of Assumption~\protect\ref{ass:interior} and the constancy of $\bar\kappa$ ($2\cdot10^6$ steps per run; the last column gives the range over the eight values of $s$).}\label{tab:checks}
\begin{tabular}{@{}l ccccccc@{}}
\toprule
 & condition & \ref{sec:dkcheck} & \ref{sec:simkernel} & \ref{sec:simrough}, $\alpha=0.6$ & \ref{sec:simrough}, $\alpha=0.7$ & \ref{sec:simrough}, single & \ref{sec:simbook}\\
\midrule
$\beta\bar\kappa$ & --- & 0.500 & 0.100 & 0.993 & 0.997 & 0.985 & 0.22--0.55\\
$\max_{n,k}p_k(n)$ & $<1$ & 0.109 & 0.033 & 0.319 & 0.443 & 0.163 & 0.056--0.112\\
$\#\{n:\max_kp_k(n)=1\}$ & 0 & 0 & 0 & 0 & 0 & 0 & 0\\
$\#\{(n,i,j):x_{i,j}(n)=0\}$ & 0 & 0 & 0 & 0 & 0 & 0 & ---\\
s.d./mean of $\bar\kappa$ over tenths & 0 & 0.001 & 0.002 & 0.001 & 0.001 & 0.004 & 0.0016--0.0023\\
\bottomrule
\end{tabular}
\end{table}

\subsection{The limit of a single cluster}\label{app:cir}

\begin{lemma}[Roughness in the limit of a single cluster]\label{lem:cir}
Let $X$ solve $dX_t=a(\mu-X_t)\,dt+\sigma\sqrt{X_t}\,dB_t$ with $a,\mu,\sigma>0$, started from its stationary law, and let $k=2a\mu/\sigma^2>2$. Then
\[
m(\Delta)=\mathbb E\bigl[(\ln X_{t+\Delta}-\ln X_t)^2\bigr]=\frac{2a}{k-1}\,\Delta+O(\Delta^{3/2}),\qquad\Delta\to0,
\]
and hence $H=\tfrac12$.
\end{lemma}

The shape of the stationary law is $k=(\mathbb E X)^2/\operatorname{Var}X$. In the run with a single cluster of Table~\ref{tab:experiments} we measure $k=2.19$ (s.e.\ 0.09).

\begin{proof}[\textbf{Proof of Lemma~\ref{lem:cir}}]
\[
\begin{aligned}
X_t&\sim\Gamma\Bigl(k,\tfrac{2a}{\sigma^2}\Bigr),\qquad \mathbb E[X_t^{-1}]=\frac{2a}{\sigma^2(k-1)},\qquad \mathbb E[X_t^{-2}]<\infty,\\
d\ln X_t&=g(X_t)\,dt+\sigma X_t^{-1/2}\,dB_t,\qquad g(x)=\frac{a\mu-\sigma^2/2}{x}-a,\\
\ln X_{t+\Delta}-\ln X_t&=\underbrace{\int_t^{t+\Delta}g(X_s)\,ds}_{D_\Delta}+\underbrace{\sigma\int_t^{t+\Delta}X_s^{-1/2}\,dB_s}_{M_\Delta},\\
\mathbb E[M_\Delta^2]&=\sigma^2\int_t^{t+\Delta}\mathbb E[X_s^{-1}]\,ds=\frac{2a}{k-1}\,\Delta,\\
\mathbb E[D_\Delta^2]&\le\Delta\int_t^{t+\Delta}\mathbb E[g(X_s)^2]\,ds=\mathbb E[g(X_t)^2]\,\Delta^2,\\
\bigl\lvert\mathbb E[D_\Delta M_\Delta]\bigr\rvert&\le\mathbb E[D_\Delta^2]^{1/2}\,\mathbb E[M_\Delta^2]^{1/2}=O(\Delta^{3/2}),\\
m(\Delta)&=\mathbb E[M_\Delta^2]+2\,\mathbb E[D_\Delta M_\Delta]+\mathbb E[D_\Delta^2]=\frac{2a}{k-1}\,\Delta+O(\Delta^{3/2}).
\end{aligned}
\]
\end{proof}

\subsection{Data of Binance}\label{app:bindata}

\begin{table}[H]
\centering\small
\caption{Trades against resting volume on the perpetual futures, 1--3 October 2026.}\label{tab:conservation}
\begin{tabular}{@{}l ccc@{}}
\toprule
 & BTC & ETH & SOL\\
\midrule
volume traded per 30\,s / volume at 1--5\%, median & 0.0021 & 0.0023 & 0.0022\\
volume traded per 30\,s / volume at 1--5\%, 99th percentile & 0.039 & 0.035 & 0.022\\
minutes with a price range above 1\% & 0 & 0 & 0\\
tick / price & $1.2\cdot10^{-6}$ & $3.7\cdot10^{-6}$ & $8.3\cdot10^{-5}$\\
\bottomrule
\end{tabular}
\end{table}

\begin{algorithm}[H]
\caption{Change of the book caused by the move of the mid price}\label{alg:mid}
\begin{algorithmic}[1]
\Require snapshots $n$ at times $t_n$ with cumulative volumes $C^{\pm}_n(x_k)$ on the ask ($+$) and bid ($-$) sides, $x_k\in\{0,0.2,1,2,3,4,5\}\%$, $C^{\pm}_n(0)=0$; last trade prices $S_t$
\Ensure $\Delta V_n(d)$ and $\Delta^{\mathrm{mid}}_n(d)$ for $d=1,\dots,4$ on both sides; slope $\varsigma$
\For{each pair of snapshots $(n,n+1)$}
\State \textbf{skip unless} $0<t_{n+1}-t_n\le120$\,s, $C^{\pm}_n$ and $C^{\pm}_{n+1}$ increase in $x$ with $C^{\pm}(0.2)>0$, $C_{n+1}\ne C_n$, and $S_{t_n},S_{t_{n+1}}$ exist
\State $\mu_n\gets100\,\bigl(\ln S_{t_{n+1}}-\ln S_{t_n}\bigr)$ \Comment{move of the mid price in \%}
\State $\widehat C^{\pm}_n(x)\gets C^{\pm}_n(x_k)+\dfrac{x-x_k}{x_{k+1}-x_k}\bigl(C^{\pm}_n(x_{k+1})-C^{\pm}_n(x_k)\bigr)$, $\ k=\min\bigl\{\max\{k:x_k\le x\},\,5\bigr\}$
\State $C^{+,\mu}_n(x)\gets\widehat C^{+}_n\bigl((x+\mu_n)^+\bigr)-\widehat C^{+}_n\bigl(\mu_n^+\bigr)$ \Comment{orders keep their price, crossed orders leave}
\State $C^{-,\mu}_n(x)\gets\widehat C^{-}_n\bigl((x-\mu_n)^+\bigr)-\widehat C^{-}_n\bigl((-\mu_n)^+\bigr)$
\State $\Delta^{\mathrm{mid}}_n(d)\gets\bigl[C^{\mu}_n(d+1)-C^{\mu}_n(d)\bigr]-\bigl[C_n(d+1)-C_n(d)\bigr]$ \Comment{each side}
\State $\Delta V_n(d)\gets\bigl[C_{n+1}(d+1)-C_{n+1}(d)\bigr]-\bigl[C_n(d+1)-C_n(d)\bigr]$
\EndFor
\State $\varsigma\gets\dfrac{\sum_{n,d,\pm}\Delta V_n(d)\,\Delta^{\mathrm{mid}}_n(d)}{\sum_{n,d,\pm}\Delta^{\mathrm{mid}}_n(d)^2}$ \Comment{kept pairs, $d=1,\dots,4$, both sides}
\end{algorithmic}
\end{algorithm}

\subsection{Desks compared with Binance}\label{app:desksbinance}
To compare the six markets with our desks, we simulate a self-exciting process with their kernel for Section~\ref{sec:bintrades} and their books for Section~\ref{sec:binbook}, with the settings of Table~\ref{tab:desksettings}.

\begin{table}[H]
\centering\footnotesize
\caption{Settings of the desks compared with Binance.}\label{tab:desksettings}
\begin{tabular}{@{}>{\raggedright\arraybackslash}p{1.6cm} >{\raggedright\arraybackslash}p{5.2cm} >{\raggedright\arraybackslash}p{5.2cm}@{}}
\toprule
 & self-exciting process (Section~\protect\ref{sec:bintrades}) & books of the desks (Section~\protect\ref{sec:binbook})\\
\midrule
desks & kernel $a\sum_{k=1}^{50}w_k\,T_k^{-1}e^{-s/T_k}$ with $w_k\propto T_k^{-\alpha}$ & Table~\protect\ref{tab:hyper} with $\alpha=0.6$\\
horizons & 50, geometric from 1\,s to $2\cdot10^6$\,s & $T_k$ of Table~\protect\ref{tab:hyper}\\
norm & $1-a=T^{-\alpha}$, $T\in\{10^4,10^5,10^6\}$\,s & $\beta\bar\kappa=0.5$, $p_0=0.02$\\
time & mean rate 6\,s$^{-1}$, 60 days, seeds 21 and 22, simulated event by event & snapshots every 50, 100 and 200 steps, 20\,000 each\\
distance & --- & 1.484 model units per 1\%, fitted to the measured cumulative volume at 0.2, 1, \dots, 5\% (largest gap 0.061)\\
measured & $H(N_b)$ as in Table~\protect\ref{tab:binance} & $\rho$ and $\varsigma$ as in Section~\protect\ref{sec:binbook}\\
\bottomrule
\end{tabular}
\end{table}

\begin{table}[H]
\centering\small
\caption{$H(N_b)$ of the self-exciting process with the kernel of the desks.}\label{tab:reduced}
\begin{tabular}{@{}l ccccc@{}}
\toprule
 & $\alpha=0.40$ & $0.45$ & $0.50$ & $0.55$ & $0.60$\\
\midrule
$T=10^4$\,s & 0.032 & 0.043 & 0.058 & 0.075 & 0.094\\
$T=10^5$\,s & 0.051 & 0.066 & 0.085 & 0.105 & 0.125\\
$T=10^6$\,s & 0.057 & 0.078 & 0.095 & 0.112 & 0.129\\
\bottomrule
\end{tabular}
\end{table}

\begin{table}[H]
\centering\small
\caption{$\rho$ and $\varsigma$ of the books of the desks.}\label{tab:modelbook}
\begin{tabular}{@{}l ccc@{}}
\toprule
model steps per snapshot & 50 & 100 & 200\\
\midrule
$\rho$ & $-0.172$ & $-0.180$ & $-0.184$\\
$\varsigma$ & 0.031 & 0.020 & 0.013\\
\bottomrule
\end{tabular}
\end{table}

\end{document}